\documentclass{article}
\usepackage[utf8]{inputenc}
\usepackage{main}
\usepackage{microtype}
\usepackage{graphicx}
\usepackage{subfig}
\usepackage{times}
\usepackage{latexsym}
\usepackage{amsmath}
\usepackage{float}
\usepackage{footnote}
\usepackage{enumitem}
\usepackage{bm}
\usepackage{arydshln}
\usepackage{booktabs}
\usepackage{multicol}
\usepackage{multirow}
\usepackage{color}
\usepackage{xcolor}     
\usepackage{colortbl}
\usepackage{bbding}
\usepackage{makecell}
\usepackage{mathtools}
\usepackage{imakeidx}
\makeindex
\usepackage{arydshln}
\usepackage{lipsum}
\usepackage{natbib}
\usepackage[toc]{multitoc}
\usepackage[edges]{forest}
\usepackage[normalem]{ulem}
\definecolor{mydarkblue}{rgb}{0,0.08,0.45}
\usepackage[colorlinks=true,linkcolor=mydarkblue,citecolor=mydarkblue,filecolor=mydarkblue,urlcolor=mydarkblue]{hyperref}
\usepackage{caption}
\usepackage{CJKutf8}
\usepackage{awesomebox} 
\usepackage{bbding}
\usepackage[most]{tcolorbox}

\usepackage[tikz]{bclogo}
\usepackage[framemethod=tikz]{mdframed}
\definecolor{bgblue}{RGB}{245,243,253}
\definecolor{ttblue}{RGB}{91,194,224}

\mdfdefinestyle{mystyle}{%
  rightline=true,
  innerleftmargin=10,
  innerrightmargin=10,
  outerlinewidth=3pt,
  topline=false,
  rightline=true,
  bottomline=false,
  skipabove=\topsep,
  skipbelow=\topsep
}

\newtcolorbox{myboxi}[1][]{
  breakable,
  title=#1,
  colback=red!5,
  colbacktitle=red!5,
  coltitle=black,
  fonttitle=\bfseries,
  bottomrule=0pt,
  toprule=0pt,
  leftrule=2pt,
  rightrule=2pt,
  titlerule=0pt,
  arc=0pt,
  outer arc=0pt,
  colframe=red,
}

\newtcolorbox{myboxnote}[1][]{
  breakable,
  title=#1,
  colback=orange!0,
  colbacktitle=orange!0,
  coltitle=black,
  fonttitle=\bfseries,
  bottomrule=0pt,
  toprule=0pt,
  leftrule=2pt,
  rightrule=2pt,
  titlerule=0pt,
  arc=0pt,
  outer arc=0pt,
  colframe=orange,
}

\newtcolorbox{myboxii}[1][]{
  breakable,
  freelance,
  title=#1,
  colback=white,
  colbacktitle=white,
  coltitle=black,
  fonttitle=\bfseries,
  bottomrule=0pt,
  boxrule=0pt,
  colframe=white,
  overlay unbroken and first={
  \draw[red!75!black,line width=3pt]
    ([xshift=5pt]frame.north west) -- 
    (frame.north west) -- 
    (frame.south west);
  \draw[red!75!black,line width=3pt]
    ([xshift=-5pt]frame.north east) -- 
    (frame.north east) -- 
    (frame.south east);
  },
  overlay unbroken app={
  \draw[red!75!black,line width=3pt,line cap=rect]
    (frame.south west) -- 
    ([xshift=5pt]frame.south west);
  \draw[red!75!black,line width=3pt,line cap=rect]
    (frame.south east) -- 
    ([xshift=-5pt]frame.south east);
  },
  overlay middle and last={
  \draw[red!75!black,line width=3pt]
    (frame.north west) -- 
    (frame.south west);
  \draw[red!75!black,line width=3pt]
    (frame.north east) -- 
    (frame.south east);
  },
  overlay last app={
  \draw[red!75!black,line width=3pt,line cap=rect]
    (frame.south west) --
    ([xshift=5pt]frame.south west);
  \draw[red!75!black,line width=3pt,line cap=rect]
    (frame.south east) --
    ([xshift=-5pt]frame.south east);
  },
}

\usepackage{fancyhdr} 
\usepackage{blindtext} 

\DeclareCaptionFont{black}{\color{black}}
\newcommand{\foo}{\color{cyan}\makebox[0pt]{\textbullet}\hskip-0.5pt\vrule width 1pt\hspace{\labelsep}}
\newcommand{\term}[1]{\index{\lowercase{#1}}\emph{#1}}

\definecolor{myblue}{rgb}{0.9, 0.1, 0.94}
\definecolor{mygreen}{rgb}{0.64, 0.56, 0.88}
\definecolor{myyellow}{rgb}{0.68, 0.6, 0.1}
\definecolor{fancygreen}{rgb}{0.33, 0.68, 0.20}
\definecolor{salmon}{rgb}{0.94, 0.52, 0.49}
\definecolor{tablegreen}{rgb}{0.82, 0.94, 0.75}
\definecolor{tableblue}{rgb}{0.81, 0.90, 0.94}
\definecolor{tablered}{rgb}{0.97, 0.85, 0.85}
\definecolor{tableorange}{rgb}{0.96, 0.85, 0.81}

\definecolor{myorange}{rgb}{1.0, 0.49, 0.0}	
\definecolor{tlgreen}{rgb}{0.33, 0.68, 0.20}

\newenvironment{itemize*}%
 {\leftmargini=10pt\begin{itemize}%
  \setlength{\itemsep}{0pt}%
  \setlength{\parskip}{0pt}%
  }%
 {\end{itemize}}
\newenvironment{enumerate*}%
 {\begin{enumerate}%
  \setlength{\itemsep}{0pt}%
  \setlength{\parskip}{0pt}}%
 {\end{enumerate}}

\tikzset{%
    parent/.style =          {align=center,text width=2cm,rounded corners=3pt, line width=0.3mm, fill=gray!10,draw=gray!80},
    child/.style =           {align=center,text width=2.3cm,rounded corners=3pt, fill=blue!10,draw=blue!80,line width=0.3mm},
    grandchild/.style =      {align=center,text width=2cm,rounded corners=3pt},
    greatgrandchild/.style = {align=center,text width=1.5cm,rounded corners=3pt},
    greatgrandchild2/.style = {align=center,text width=1.5cm,rounded corners=3pt},    
    referenceblock/.style =  {align=center,text width=1.5cm,rounded corners=2pt},
    pretrain/.style =           {align=center,text width=1.8cm,rounded corners=3pt, fill=blue!10,draw=blue!80,line width=0.3mm},   
    pretrain_work/.style =           {align=center, text width=5cm,rounded corners=3pt, fill=blue!10,draw=blue!0,line width=0.3mm},  
    template/.style =           {align=center,text width=1.8cm,rounded corners=3pt, fill=red!10,draw=red!80,line width=0.3mm},   
    template_work/.style =           {align=center,text width=5cm,rounded corners=3pt, fill=red!10,draw=red!0,line width=0.3mm},    
    answer/.style =           {align=center,text width=1.8cm,rounded corners=3pt, fill= cyan!10,draw= cyan!80,line width=0.3mm},   
    answer_work/.style =           {align=center,text width=5cm,rounded corners=3pt, fill= cyan!10,draw= cyan!0,line width=0.3mm},      
    multiple/.style =           {align=center,text width=1.8cm,rounded corners=3pt, fill= orange!10,draw= orange!80,line width=0.3mm},   
    multiple_work/.style =           {align=center,text width=5cm,rounded corners=3pt, fill= orange!10,draw= orange!0,line width=0.3mm},        
    tuning/.style =           {align=center,text width=1.8cm,rounded corners=3pt, fill= magenta!10,draw= magenta!80,line width=0.3mm},   
    tuning_work/.style =           {align=center,text width=5cm,rounded corners=3pt, fill= magenta!10,draw= magenta!0,line width=0.3mm},          
}

\usepackage{etoolbox}
\usepackage{natbib}
\usepackage{url}
\newcounter{bibcount}
\makeatletter
\patchcmd{\@lbibitem}{\item[}{\item[\hfil\stepcounter{bibcount}{[\thebibcount]}}{}{}
\renewcommand\NAT@bibsetup%
  [1]{\setlength{\leftmargin}{\bibhang}\setlength{\itemindent}{-\parindent}%
      \setlength{\itemsep}{\bibsep}\setlength{\parsep}{\z@}}
\makeatother

\begin{document}

\title{Pre-train, Prompt, and Predict: A Systematic Survey of Prompting Methods in Natural Language Processing}

\author{
  Pengfei Liu \\
  Carnegie Mellon University \\
\texttt{pliu3@cs.cmu.edu}
\And
Weizhe Yuan \\
  Carnegie Mellon University \\
  \texttt{weizhey@cs.cmu.edu} \\
  \And
Jinlan Fu \\
  National University of Singapore \\
\texttt{jinlanjonna@gmail.com} \\
  \And  
  Zhengbao Jiang \\
  Carnegie Mellon University \\
\texttt{zhengbaj@cs.cmu.edu}   
  \And
  Hiroaki Hayashi \\
  Carnegie Mellon University \\
\texttt{hiroakih@cs.cmu.edu}   
  \And  
  Graham Neubig \\
  Carnegie Mellon University \\
\texttt{gneubig@cs.cmu.edu} 
  }
  
\maketitle

\begin{abstract}
    This paper surveys and organizes research works in a new paradigm in natural language processing, which we dub ``prompt-based learning''.
    Unlike traditional supervised learning, which trains a model to take in an input $\bm{x}$ and predict an output $\bm{y}$ as $P(\bm{y}|\bm{x})$, prompt-based learning is based on language models that model the probability of text directly.
    To use these models to perform prediction tasks, the original input $\bm{x}$ is modified using a \term{template} into a textual string \term{prompt} $\bm{x}'$ that has some unfilled slots, and then the language model is used to probabilistically fill the unfilled information to obtain a final string $\hat{\bm{x}}$, from which the final output $\bm{y}$ can be derived.
    This framework is powerful and attractive for a number of reasons: it allows the language model to be \term{pre-trained} on massive amounts of raw text, and by defining a new prompting function the model is able to perform \term{few-shot} or even \term{zero-shot} learning, adapting to new scenarios with few or no labeled data.
    In this paper we introduce the basics of this promising paradigm, describe a unified set of mathematical notations that can cover a wide variety of existing work, and organize existing work along several dimensions, e.g.~the choice of pre-trained models, prompts, and tuning strategies.
    To make the field more accessible to interested beginners, we not only make a systematic review of existing works and a highly structured typology of prompt-based concepts, but also release other resources, e.g., a website 
    \includegraphics[scale=0.01]{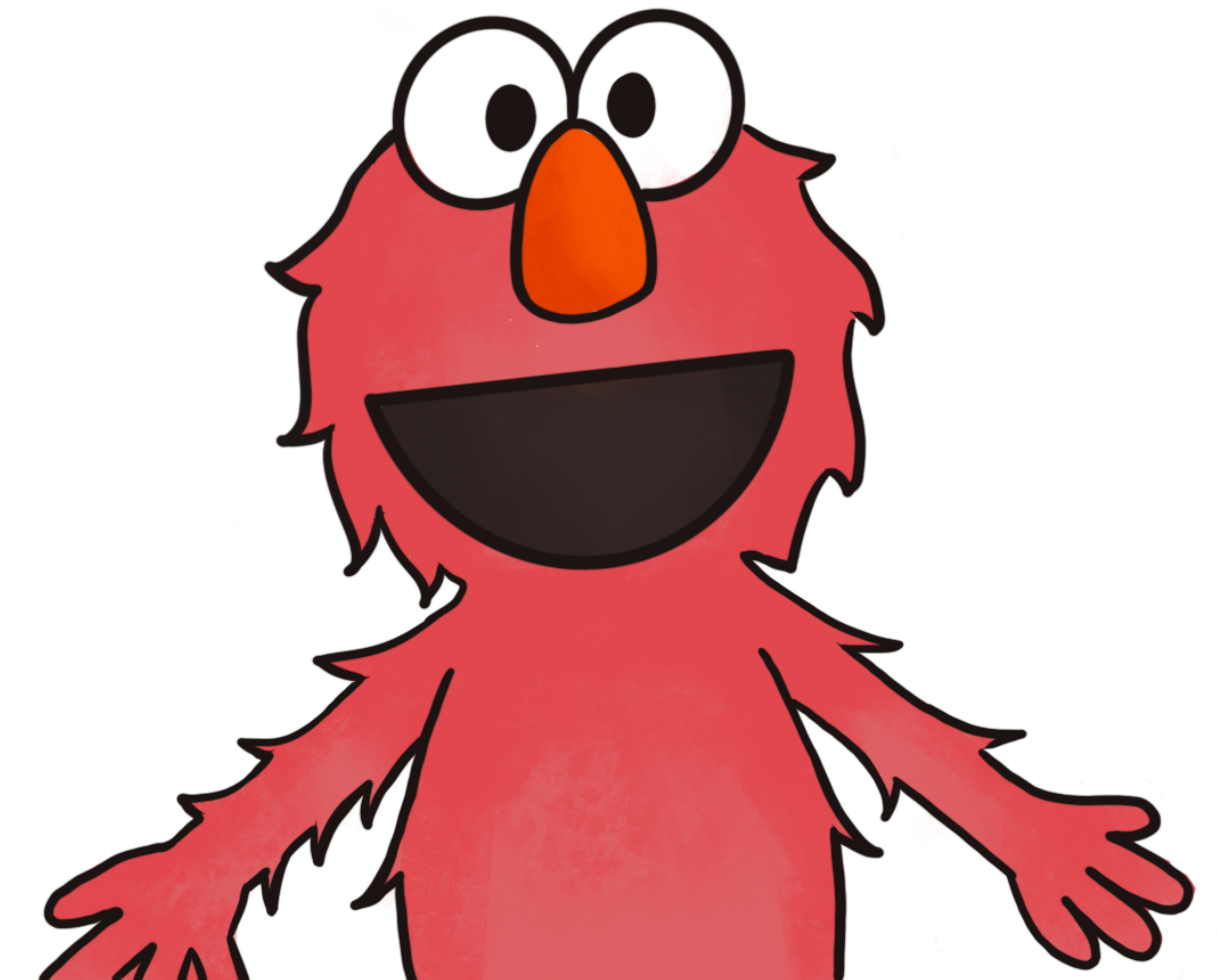}\includegraphics[scale=0.01]{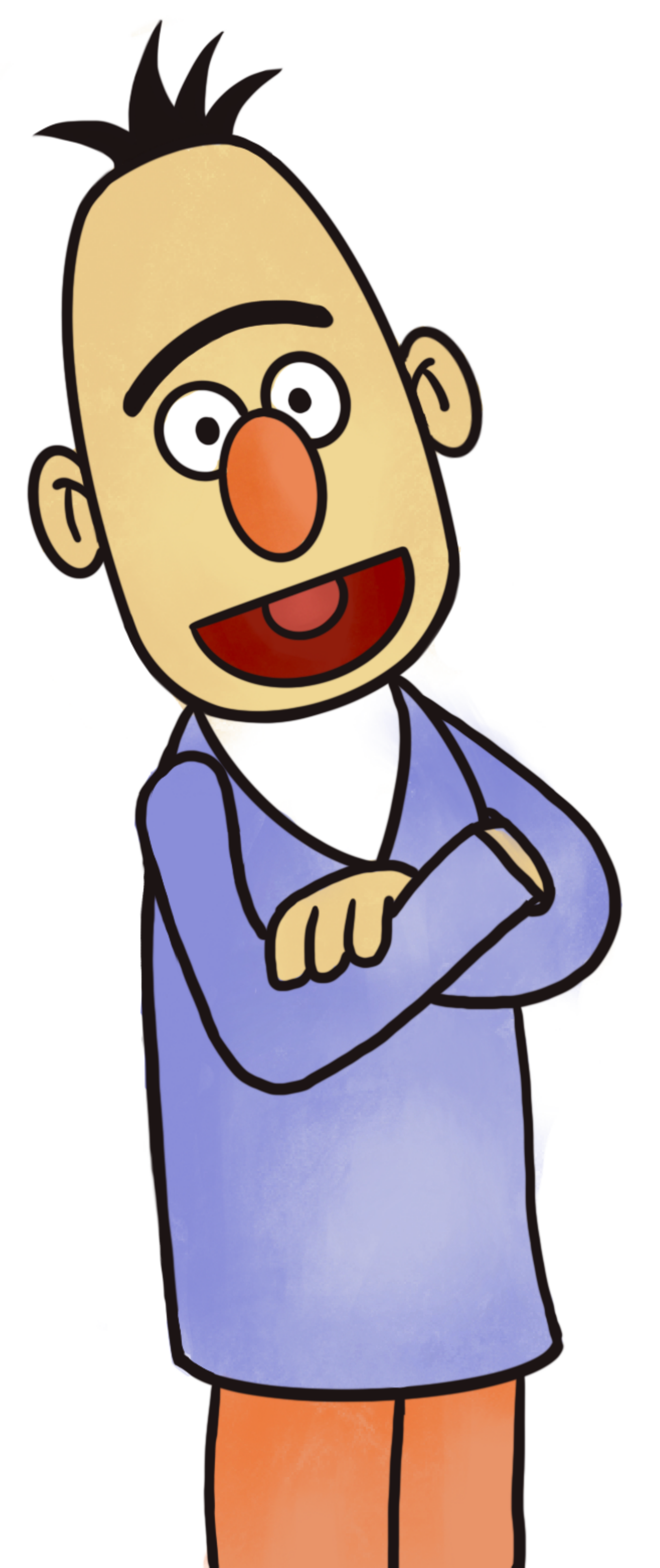}
    \href{http://pretrain.nlpedia.ai/}{NLPedia--Pretrain} \includegraphics[scale=0.01]{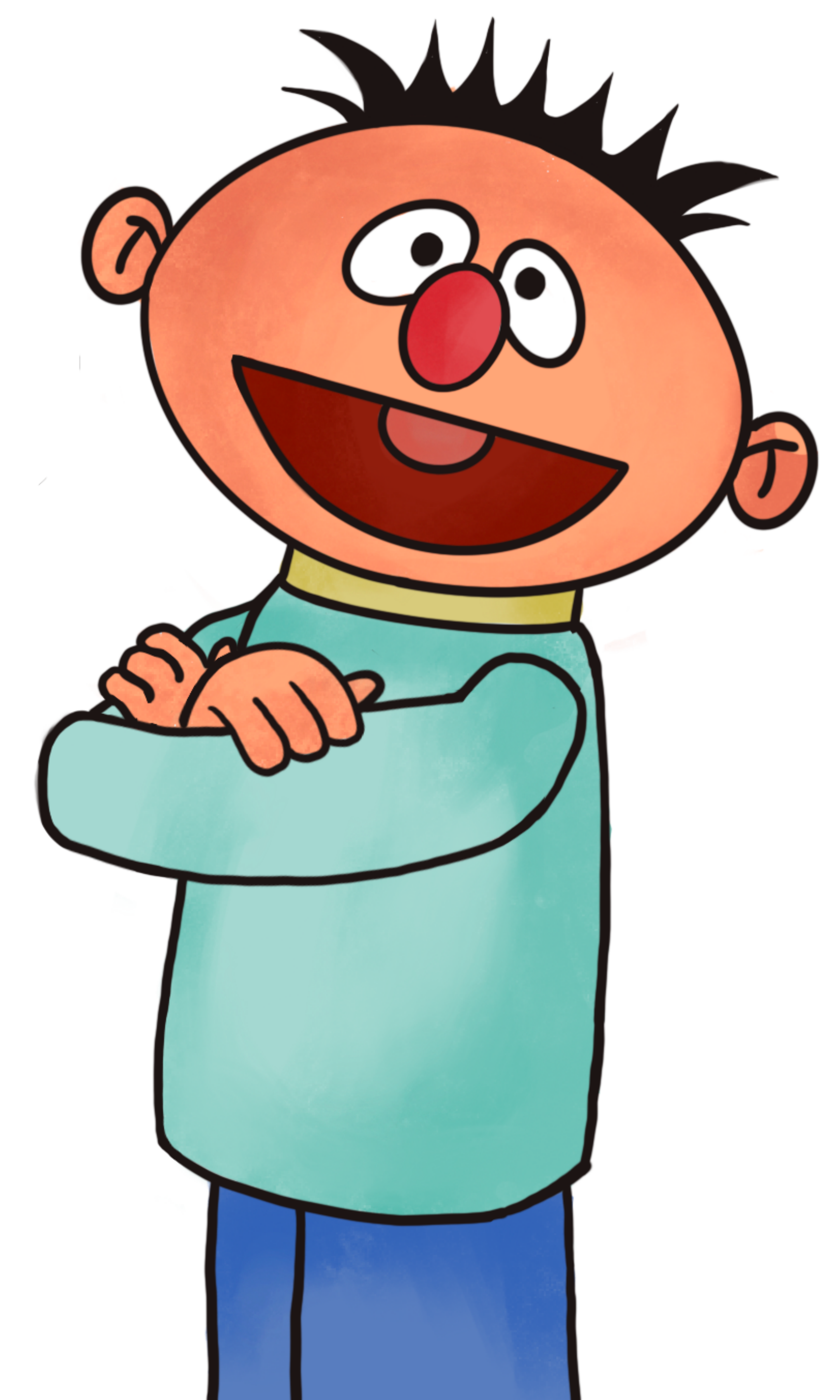}\includegraphics[scale=0.01]{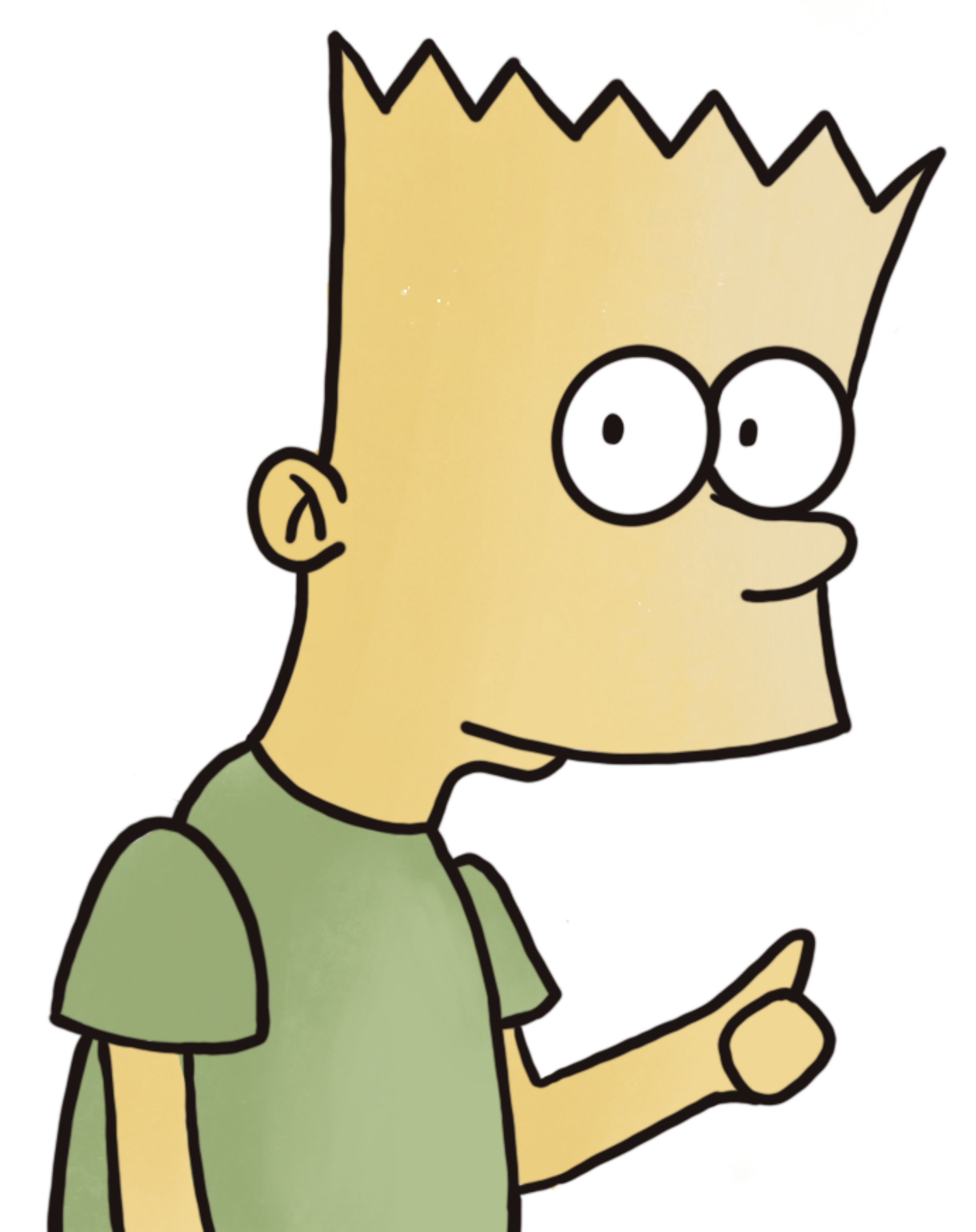}    
     including constantly-updated survey, and paperlist.    
\end{abstract}

\begin{figure}[h]
\centering
\includegraphics[width=14cm]{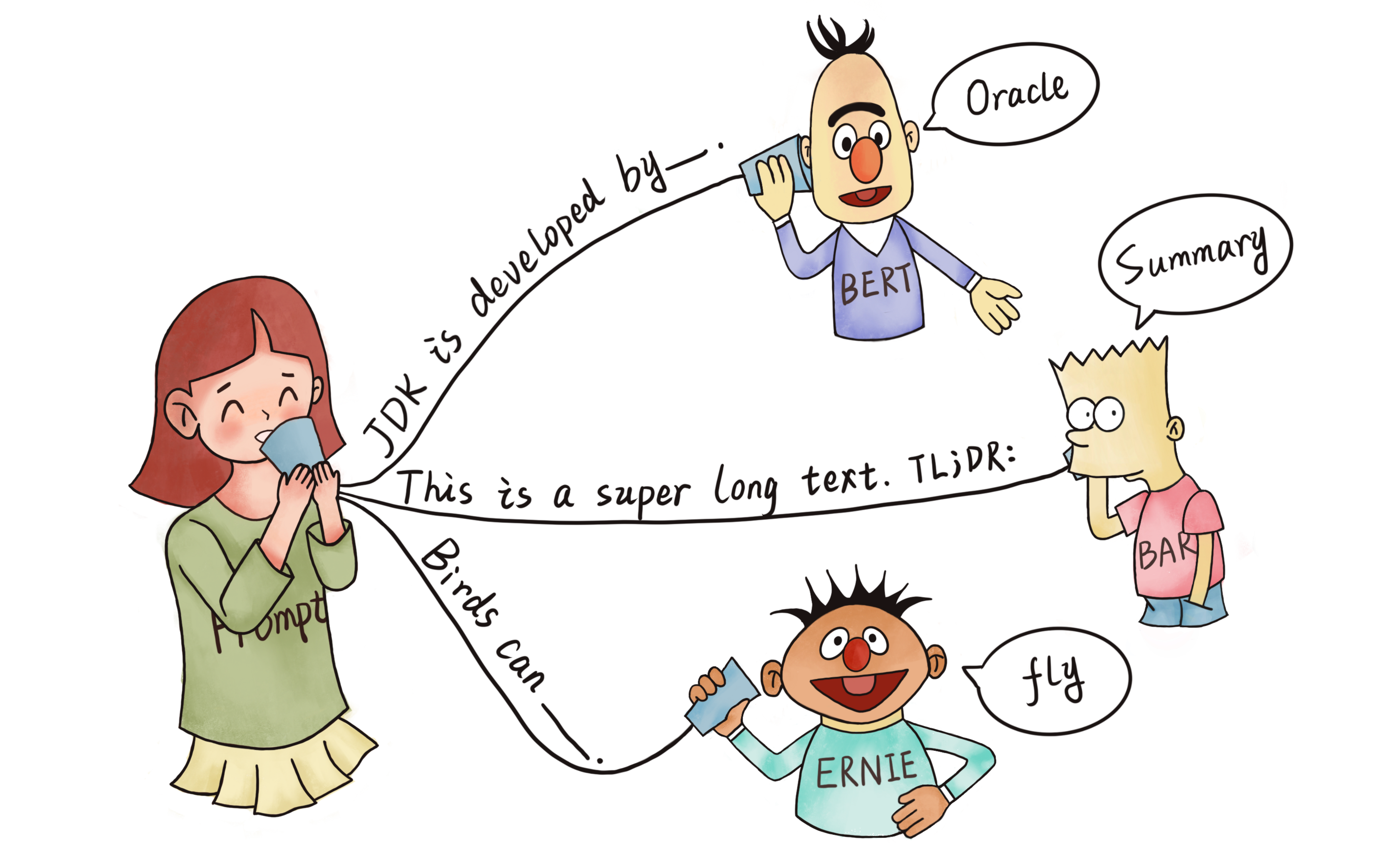}
\end{figure}
\newpage

\tableofcontents

\clearpage

\section{Two Sea Changes in NLP} \label{sec:four-paradigms}
 
\term{Fully supervised learning}, where a task-specific model is trained solely on a dataset of input-output examples for the target task, has long played a central role in many machine learning tasks \cite{kotsiantis2007supervised}, and natural language processing (NLP) was no exception.
Because such fully supervised datasets are ever-insufficient for learning high-quality models, early NLP models relied heavily on \term{feature engineering} (Tab.~\ref{table:intro} a.; e.g.~\citet{Lafferty2001ConditionalRF,guyon2002gene,och-etal-2004-smorgasbord,zhang-nivre-2011-transition}), where NLP researchers or engineers used their domain knowledge to define and extract salient features from raw data and provide models with the appropriate inductive bias to learn from this limited data.
With the advent of neural network models for NLP, salient features were learned jointly with the training of the model itself \cite{Collobert2011NaturalLP,bengio2013representation}, and hence focus shifted to \term{architecture engineering}, where inductive bias was rather provided through the design of a suitable network architecture conducive to learning such features (Tab.~\ref{table:intro} b.; e.g.~\citet{hochreiter1997long,kalchbrenner-etal-2014-convolutional,Chung2014EmpiricalEO,Kim2014ConvolutionalNN,bahdanau2014neural,vaswani2017attention}).%
\footnote{
Even during this stage, there was some use of pre-trained models exemplified by word2vec~\cite{Mikolov2013DistributedRO,Mikolov2013EfficientEO} and GloVe~\cite{Pennington2014GloveGV}, but they were used for only a limited portion of the final model parameters.
}

However, from 2017-2019 there was a sea change in the learning of NLP models, and this fully supervised paradigm is now playing an ever-shrinking role.
Specifically, the standard shifted to the \term{pre-train and fine-tune} paradigm (Tab.~\ref{table:intro} c.; e.g.~\citet{Radford2018ImprovingLU,peters-etal-2018-deep,DBLP:conf/nips/00040WWLWGZH19,DBLP:conf/nips/YangDYCSL19,lewis-etal-2020-bart}).
In this paradigm, a model with a fixed\footnote{This paradigm is less conducive to architectural exploration because (i) unsupervised pre-training allows models to learn with fewer structural priors, and
(ii) as pre-training of models is time-consuming, experimenting with structural variants is costly.} architecture is \term{pre-trained} as a language model (LM), predicting the probability of observed textual data. Because the raw textual data necessary to train LMs is available in abundance, these LMs can be trained on large datasets, in the process learning robust general-purpose features of the language it is modeling.
The above pre-trained LM will be then adapted to different downstream tasks by introducing additional parameters and \term{fine-tuning} them using task-specific objective functions.
Within this paradigm, the focus turned mainly to \term{objective engineering}, designing the training objectives used at both the pre-training and fine-tuning stages. For example, \citet{DBLP:conf/icml/ZhangZSL20} show that introducing a loss function of predicting salient sentences from a document will lead to a better pre-trained model for text summarization.
Notably, the main body of the pre-trained LM is generally (but not always; \citet{peters-etal-2019-tune}) fine-tuned as well to make it more suitable for solving the downstream task.

Now, as of this writing in 2021, we are in the middle of a second sea change, in which the ``pre-train, fine-tune'' procedure is replaced by one in which we dub ``\term{pre-train, prompt, and predict}''.
In this paradigm, instead of adapting pre-trained LMs to downstream tasks via objective engineering, downstream tasks are reformulated to look more like those solved during the original LM training with the help of a textual \term{prompt}.
For example, when recognizing the emotion of a social media post, ``I missed the bus today.'', we may continue with a prompt ``I felt so \underline{\hspace*{0.5cm}}'', and ask the LM to fill the blank with an emotion-bearing word.
Or if we choose the prompt ``English: I missed the bus today. French: \underline{\hspace*{0.5cm}}''), an LM may be able to fill in the blank with a French translation.
In this way, by selecting the appropriate prompts we can manipulate the model behavior so that the pre-trained LM itself can be used to \term{predict} the desired output, sometimes even without any additional task-specific training (Tab.~\ref{table:intro} d.; e.g.~\citet{Radford2019LanguageMA,petroni-etal-2019-language,brown2020language,JMLR:v21:20-074,schick2021its,gao2021making}).
The advantage of this method is that, given a suite of appropriate prompts, a single LM trained in an entirely unsupervised fashion can be used to solve a great number of tasks \cite{brown2020language,DBLP:journals/corr/abs-2107-02137}. 
However, as with most conceptually enticing prospects, there is a catch -- this method introduces the necessity for \term{prompt engineering}, finding the most appropriate prompt to allow a LM to solve the task at hand.

This survey attempts to organize the current state of knowledge in this rapidly developing field by providing an overview and formal definition of prompting methods (\S\ref{sec:2-formal-description}), and an overview of the pre-trained language models that use these prompts (\S\ref{sec:lm}).
This is followed by in-depth discussion of prompting methods, from basics such as prompt engineering (\S\ref{sec:4-prompt-template-engineering}) and answer engineering (\S\ref{sec:5-prompt-answer-engineering}) to more advanced concepts such as multi-prompt learning methods (\S\ref{sec:6-multi-prompt-learning}) and prompt-aware training methods (\S\ref{sec:tuning}).
We then organize the various applications to which prompt-based learning methods have been applied, and discuss how they interact with the choice of prompting method (\S\ref{sec:applications}).
Finally, we attempt to situate the current state of prompting methods in the research ecosystem, making connections to other research fields (\S\ref{sec:related}), suggesting some current challenging problems that may be ripe for further research (\S\ref{sec:challenges}), and performing a meta-analysis of current research trends (\S\ref{sec:meta-analysis}).

Finally, in order to help beginners who are interested in this field  learn more effectively, we highlight some systematic  resources about prompt learning (as well as pre-training) provided both within this survey and on companion websites:
\begin{itemize*}
    \item \includegraphics[scale=0.01]{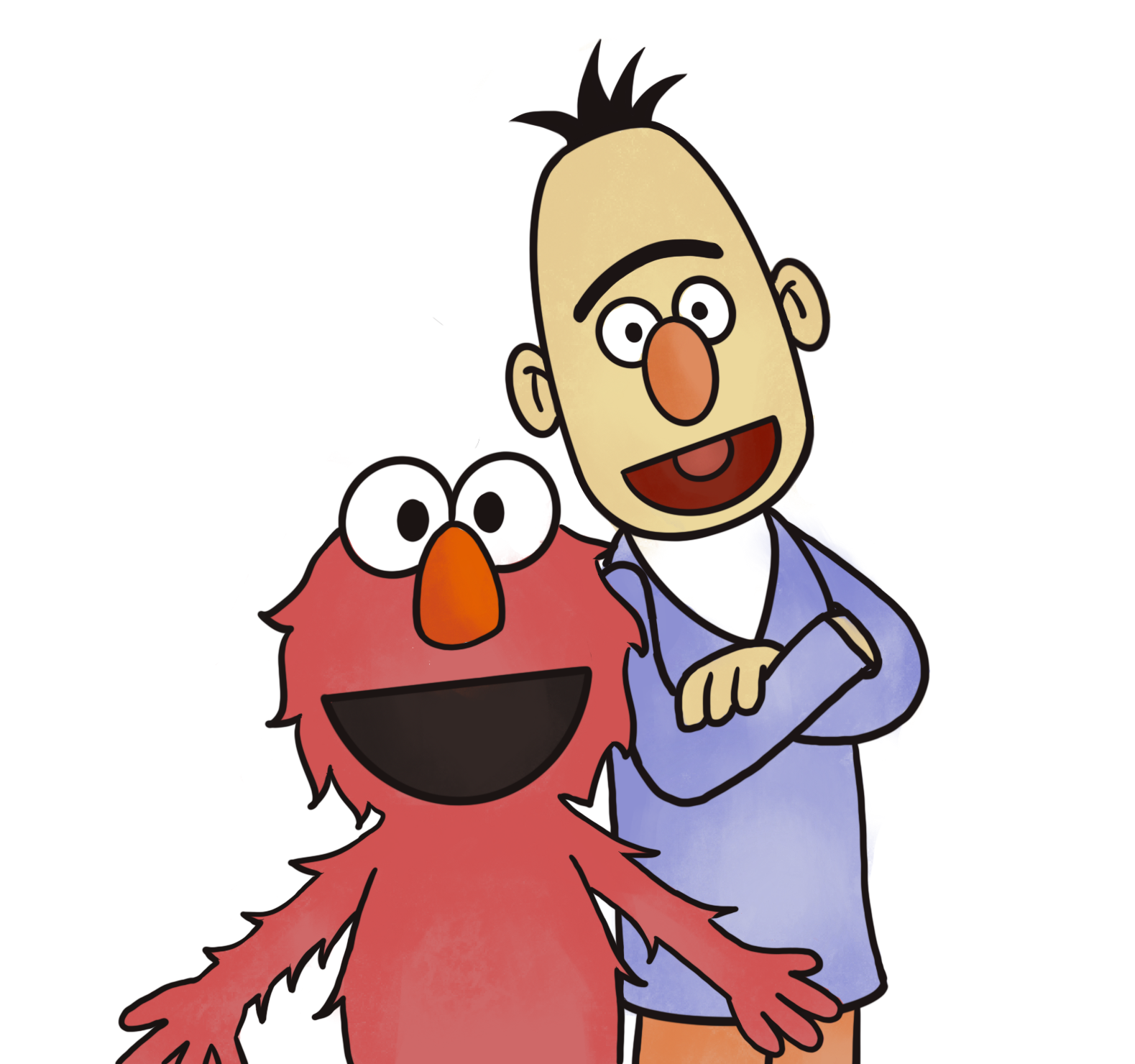}: A 
    \href{http://pretrain.nlpedia.ai/}{website } of prompt-based learning that contains: frequent updates to this survey, related slides, etc.
    \item Fig.\ref{fig:typo-prompt}: A typology of important concepts for prompt-based learning.
    \item Tab.\ref{tab:papers-part1}: A systematic and comprehensive comparison among different prompting methods.  
    \item Tab.\ref{tab:task-prompt}: An organization of commonly-used prompts.
    \item Tab.\ref{tab:timeline}: A timeline of prompt-based research works.
    \item Tab.\ref{tab:pretrained-aspect}: A systematic and comprehensive comparison among different pre-trained LMs.    
\end{itemize*}

\begin{table*}[!t]
\centering
\footnotesize
\setlength\tabcolsep{15pt}
\renewcommand{\arraystretch}{1.3}
\begin{tabular}{llc}
\toprule
\textbf{Paradigm}                                                    & \textbf{Engineering}                   & \textbf{Task Relation}           \\
\midrule
\multirow{4}{4cm}{a. Fully Supervised Learning (Non-Neural Network) }                                   & \multirow{4}{4.5cm}{Features\hspace{3cm}  (e.g.~word identity, part-of-speech, sentence length)}      & \hspace{-1.8cm}\multirow{4}{*}{\parbox[c]{0em}{\includegraphics[width=0.7in]{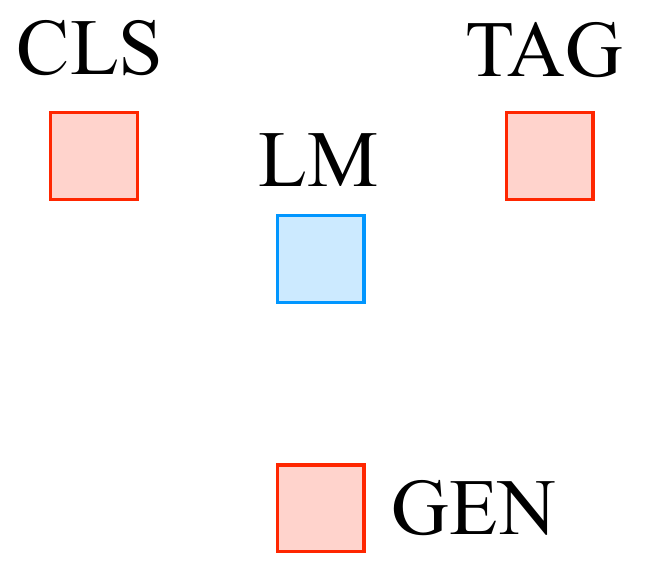}}}  \\

\\
\\
                                       &                               &                                     \\
\midrule
\multirow{4}{4cm}{b. Fully Supervised Learning (Neural Network)} & \multirow{4}{4.5cm}{Architecture \hspace{2cm} (e.g.~convolutional, recurrent, self-attentional)} & \hspace{-1.8cm}\multirow{4}{*}{\parbox[c]{0em}{\includegraphics[width=0.7in]{fig/IntroFig/stage2-taskrel.pdf}}} \\
\\
                                                            &                               &                    \\
                                                            &                               &                               \\
\midrule
\multirow{4}{4cm}{c. Pre-train, Fine-tune}              & \multirow{4}{4.5cm}{Objective \hspace{2cm} (e.g.~masked language modeling, next sentence prediction)}    & \hspace{-1.8cm}\multirow{4}{*}{\parbox[c]{0em}{\includegraphics[width=0.7in]{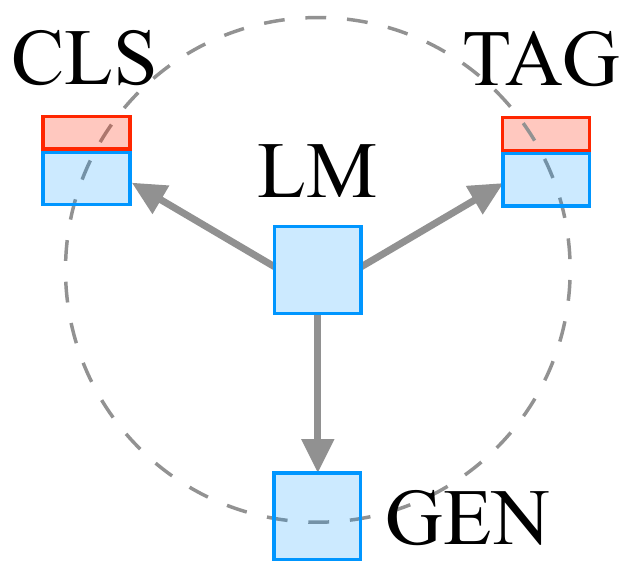}}}  \\
\\
                                                            &                               &                               \\
                                                            &                               &                                     \\
\midrule
\multirow{4}{*}{d. Pre-train, Prompt, Predict}                 & \multirow{4}{4.5cm}{Prompt (e.g.~cloze, prefix)}       & \hspace{-1.8cm}\multirow{4}{*}{\parbox[c]{0em}{\includegraphics[width=0.7in]{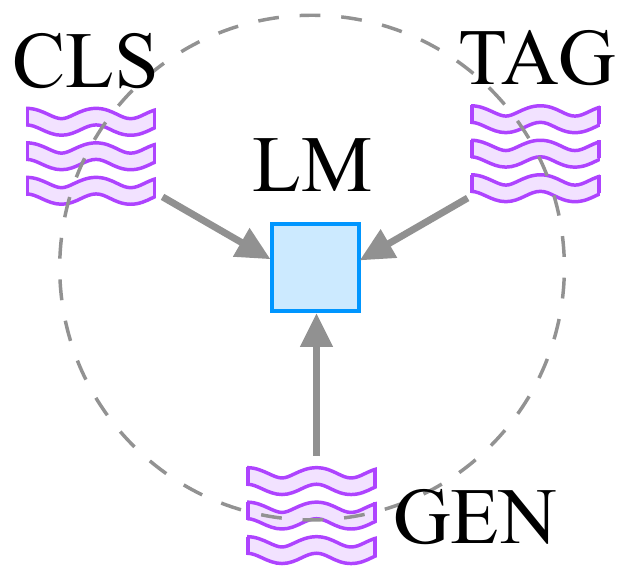}}}  \\
\\
                                                            &                               &                                      \\
                                                            &                               &                                     \\
                    
\bottomrule
\end{tabular}
\caption{Four paradigms in NLP. The ``\textbf{engineering}'' column represents the type of engineering to be done to build strong systems. The ``\textbf{task relation}'' column, shows the relationship between language models (LM) and other NLP tasks (CLS: classification, TAG: sequence tagging, GEN: text generation). \includegraphics[scale=0.3]{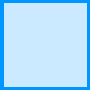}: fully unsupervised training.
\includegraphics[scale=0.3]{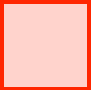}: fully supervised training.
\includegraphics[scale=0.3]{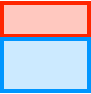}: Supervised training combined with unsupervised training.
\includegraphics[scale=0.3]{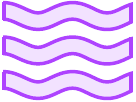} indicates a textual prompt. 
Dashed lines suggest that different tasks can be connected by sharing parameters of pre-trained models. ``LM$\rightarrow$Task'' represents \textit{adapting LMs (objectives) to downstream tasks} while ``Task$\rightarrow$LM'' denotes \textit{adapting downstream tasks (formulations) to LMs}.} 
\label{table:intro}
\end{table*}

\section{A Formal Description of Prompting}
\label{sec:2-formal-description}

\subsection{Supervised Learning in NLP}

In a traditional supervised learning system for NLP, we take an \textbf{input} $\bm{x}$, usually text, and predict an \textbf{output} $\bm{y}$ based on a model $P(\bm{y}|\bm{x};\theta)$.
$\bm{y}$ could be a label, text, or other variety of output.
In order to learn the parameters $\theta$ of this model, we use a dataset containing pairs of inputs and outputs, and train a model to predict this conditional probability.
We will illustrate this with two stereotypical examples.

First, \term{text classification} takes an input text $\bm{x}$ and predicts a label $y$ from a fixed label set $\mathcal{Y}$.
To give an example, sentiment analysis \citep{pang-etal-2002-thumbs,socher2013recursive} may take an input $\bm{x}=$``I love this movie.'' and predict a label $y=\text{++}$, out of a label set $\mathcal{Y}=\{\text{++}, \text{+}, \text{\texttildelow}, \text{-}, \text{-}\text{-}\}$.

Second, \term{conditional text generation} takes an input $\bm{x}$ and generates another text $\bm{y}$.
One example is machine translation \citep{koehn2009statistical}, where the input is text in one language such as the Finnish $\bm{x}=\text{``Hyvää huomenta.''}$ and the output is the English $\bm{y}=\text{``Good morning''.}$.

\subsection{Prompting Basics} \label{sec:2-basics}

The main issue with supervised learning is that in order to train a model $P(\bm{y}|\bm{x};\theta)$, it is necessary to have supervised data for the task, which for many tasks cannot be found in large amounts.
Prompt-based learning methods for NLP attempt to circumvent this issue by instead learning an LM that models the probability $P(\bm{x};\theta)$ of text $\bm{x}$ itself (details in \S\ref{sec:lm}) and using this probability to predict $\bm{y}$, reducing or obviating the need for large supervised datasets.
In this section we lay out a mathematical description of the most fundamental form of prompting, which encompasses many works on prompting and can be expanded to cover others as well.
Specifically, basic prompting predicts the highest-scoring $\hat{\bm{y}}$ in three steps.

\begin{table*}[!t]
\centering
\footnotesize
\renewcommand{\arraystretch}{1.2}
\begin{tabular}{m{1.8cm}cm{5.8cm}m{5.2cm}}

\toprule
\textbf{Name}         & \textbf{Notation}           & \textbf{Example}                                                         & \textbf{Description}                                                 \\ 
         \midrule
\term{Input}     & $\bm{x}$                  & I love this movie.                                               & One or multiple texts                                    \\

\term{Output}   & $\bm{y}$                  & \texttt{++} (very positive)                                                        &  Output label or text   \\
\midrule
\term{Prompting Function}  & $f_{\text{prompt}}(\bm{x})$ & \texttt{[X]} Overall, it was a \texttt{[Z]} movie.                      & A function that converts the input into a specific form by inserting the input $\bm{x}$ and adding a slot \texttt{[Z]} where answer $\bm{z}$ may be filled later.
\\
\midrule
\term{Prompt}     & $\bm{x}'$       & I  love  this  movie.    Overall,  it  was  a \texttt{[Z]} movie. & A text where \texttt{[X]} is instantiated by input $\bm{x}$ but answer slot \texttt{[Z]} is not. \\
\term{Filled Prompt}     &  $f_{\text{fill}}(\bm{x'}, \bm{z})$      & I  love  this  movie.    Overall,  it  was  a bad movie. & A prompt where slot \texttt{[Z]} is filled with any answer. \\
\term{Answered Prompt}     & $f_{\text{fill}}(\bm{x'}, \bm{z}^{*})$       & I  love  this  movie. Overall, it was a good movie. & A prompt where slot \texttt{[Z]} is filled with a true answer. 
\\
\midrule
\term{Answer}    & $\bm{z}$                  & ``good'', ``fantastic'', ``boring''                                             & A token,  phrase, or sentence that fills \texttt{[Z]}                                 \\
\bottomrule
\end{tabular}
\caption{\label{tab:example} Terminology and notation of prompting methods. $\bm{z}^*$ represents answers that correspond to true output $\bm{y}^*$. 
}
\end{table*}

\subsubsection{Prompt Addition}
In this step a \term{prompting function} $f_{\text{prompt}}(\cdot)$ is applied to modify the input text $\bm{x}$ into a \term{prompt} $\bm{x}' = f_{\text{prompt}}(\bm{x})$.
In the majority of previous work~\cite{kumar2016ask,mccann2018natural,Radford2019LanguageMA,schick2021exploiting}, this function consists of a two step process:
\begin{enumerate}
    \item Apply a \term{template}, which is a textual string that has two slots: an \term{input slot} \texttt{[X]} for input $\bm{x}$ and an \term{answer slot} \texttt{[Z]} for an intermediate generated \term{answer} text $\bm{z}$ that will later be mapped into $\bm{y}$.
    \item Fill slot \texttt{[X]} with the input text $\bm{x}$.
\end{enumerate}

In the case of sentiment analysis where $\bm{x}=$``I love this movie.'', the template may take a form such as ``\texttt{[X]} Overall, it was a \texttt{[Z]} movie.''.
Then, $\bm{x}'$ would become ``I love this movie. Overall it was a \texttt{[Z]} movie.'' given the previous example.
In the case of machine translation, the template may take a form such as ``Finnish: \texttt{[X]} English: \texttt{[Z]}'', where the text of the input and answer are connected together with headers indicating the language.
We show more examples in Tab.~\ref{tab:def}

\begin{table*}[!t]
\centering
\footnotesize
\renewcommand{\arraystretch}{1}
\setlength\tabcolsep{6pt}
\begin{tabular}{lllll}
\toprule
\bf{Type}                                      & \bf{Task}                            & \bf{Input (\texttt{[X]})}                                 & \bf{Template}                                                        & \bf{Answer (\texttt{[Z]})}                     \\
\midrule
\multirow{9}{*}{Text CLS}      & \multirow{3}{*}{Sentiment}        & \multirow{3}{*}{I love this movie.}       & \multirow{3}{*}{\texttt{[X]} The movie is \texttt{[Z]}.}                 & great                     \\
                                          &                                   &                                           &                                                                  & fantastic                 \\
                                          &                                   &                                           &                                                                  & ...                       \\
                                           \cmidrule{2-5}
                                          & \multirow{3}{*}{Topics}           & \multirow{3}{*}{He prompted the LM.}  & \multirow{3}{*}{\texttt{[X]} The text is about \texttt{[Z]}.}        & sports                    \\
                                          &                                   &                                           &                                                                  & science                   \\
                                          &                                   &                                           &                                                                  & ...                       \\
                                          \cmidrule{2-5}
                                          & \multirow{3}{*}{Intention}        & \multirow{3}{*}{What is taxi fare to Denver?} & \multirow{3}{*}{\texttt{[X]} The question is about \texttt{[Z]}.}        & quantity                  \\
                                          &                                   &                                           &                                                                  & city                      \\
                                          &                                   &                                           &                                                                  & ...                       \\
                                          \midrule
\multirow{3}{*}{Text-span CLS} & \multirow{3}{2cm}{Aspect Sentiment} & \multirow{3}{*}{Poor service but good food.}         &                                                                  & Bad                       \\
                                          &                                   &                                           & \texttt{[X]} What about service? \texttt{[Z]}.               & Terrible                  \\
                                          &                                   &                                           &                                                           & ...                       \\
                                          \midrule
\multirow{3}{*}{Text-pair CLS} & \multirow{3}{*}{NLI}              & \texttt{[X1]}: An old man with ...                   & \multirow{3}{*}{ \texttt{[X1]}? \texttt{[Z]}, \texttt{[X2]}} & Yes                       \\
                                          &                                   & \texttt{[X2]}: A man walks ...                       &                                                                  & No                     \\
                                          &                                   &                                           &                                                                  & ...                       \\
                                          \midrule
\multirow{3}{*}{Tagging}        &   & \texttt{[X1]}: Mike went to Paris.   &    & organization  \\
                                          & NER  & \texttt{[X2]}: Paris &  \texttt{[X1]}\texttt{[X2]} is a \texttt{[Z]} entity.                                                                 &           location                \\
                                          &                           &                                           &                                                                  &      ...                     \\
                                          \midrule
\multirow{6}{*}{Text Generation}          & \multirow{3}{*}{Summarization}    & \multirow{3}{*}{Las Vegas police ...}     & \multirow{3}{*}{\texttt{[X]} TL;DR: \texttt{[Z]}}                       & The victim ... \\
                                          &                                   &                                           &                                                                  & A woman ...    \\
                                          &                                   &                                           &                                                                  & ...                       \\
                                          \cmidrule{2-5}
                                          & \multirow{3}{*}{Translation}               & \multirow{3}{*}{Je vous aime.}            & \multirow{3}{*}{French: \texttt{[X]} English: \texttt{[Z]}}             & I love you.               \\
                                          &                                   &                                           &                                                                  & I fancy you.     \\
                                          &                                   &                                           &                                                                  & ...  \\
                                          \bottomrule
\end{tabular}
\caption{\label{tab:def}Examples of \term{input}, \term{template}, and \term{answer} for different tasks. In the \textbf{Type} column, ``CLS" is an abbreviation for ``classification". In the \textbf{Task} column, ``NLI" and ``NER'' are abbreviations for ``natural language inference"~\cite{bowman-etal-2015-large} and ``named entity recognition"~\cite{tjong-kim-sang-de-meulder-2003-introduction} respectively.}
\end{table*}

Notably, 
(1) the prompts above will have an empty slot to fill in for $\bm{z}$, either in the middle of the prompt or at the end. 
In the following text, we will refer to the first variety of prompt with a slot to fill in the middle of the text as a \term{cloze prompt}, and the second variety of prompt where the input text comes entirely before $\bm{z}$ as a \term{prefix prompt}.
(2) In many cases these template words are not necessarily composed of natural language tokens; they could be virtual words (e.g.~represented by numeric ids) which would be embedded in a continuous space later, and some prompting methods even generate continuous vectors directly (more in \S\ref{sec:4-continuous}). (3) The number of \texttt{[X]} slots and the number of \texttt{[Z]} slots can be flexibly changed for the need of tasks at hand.

\subsubsection{Answer Search}
Next, we search for the highest-scoring text $\hat{\bm{z}}$ 
that maximizes the score of the LM.
We first define $\mathcal{Z}$ as a set of permissible values for $\bm{z}$.
$\mathcal{Z}$ could range from the entirety of the language in the case of generative tasks, or could be a small subset of the words in the language in the case of classification, such as defining $\mathcal{Z} = \{\text{``excellent''}, \text{``good''}, \text{``OK''}, \text{``bad''}, \text{``horrible''}\}$ to represent each of the classes in $\mathcal{Y}=\{\text{++}, \text{+}, \text{\texttildelow}, \text{-}, \text{-}\text{-}\}$.

We then define a function $f_{\text{fill}}(\bm{x'}, \bm{z})$ that fills in the location $\texttt{[Z]}$ in prompt $\bm{x}'$ with the potential answer $\bm{z}$.
We will call any prompt that has gone through this process as a \term{filled prompt}. Particularly, if the prompt is filled with a true answer, we will refer to it as an \term{answered prompt} (Tab.~\ref{tab:example} shows an example).
Finally, we search over the set of potential answers $\bm{z}$ by calculating the probability of their corresponding filled prompts using a pre-trained LM $P(\cdot;\theta)$
\begin{equation}
    \hat{\bm{z}} = \underset{\bm{z} \in \mathcal{Z}}{\text{search}} \, P(f_{\text{fill}}(\bm{x'}, \bm{z}); \theta).
\end{equation}
This search function could be an \term{argmax} search that searches for the highest-scoring output, or \term{sampling} that randomly generates outputs following the probability distribution of the LM.

\subsubsection{Answer Mapping}

Finally, we would like to go from the highest-scoring \emph{answer}  $\hat{\bm{z}}$ to the highest-scoring \emph{output} $\hat{\bm{y}}$.
This is trivial in some cases, where the answer itself is the output (as in language generation tasks such as translation), but there are also other cases where multiple answers could result in the same output.
For example, one may use multiple different sentiment-bearing words (e.g.~``excellent'', ``fabulous'', ``wonderful'') to represent a single class (e.g.~``++''), in which case it is necessary to have a mapping between the searched answer and the output value.

\subsection{Design Considerations for Prompting} \label{sec:design-consider}

Now that we have our basic mathematical formulation, we elaborate a few of the basic design considerations that go into a prompting method, which we will elaborate in the following sections:
\begin{itemize}
    \item \textbf{Pre-trained Model Choice:} There are a wide variety of pre-trained LMs 
    that could be used to calculate $P(\bm{x};\theta)$. In \S\ref{sec:lm} we give a primer on pre-trained LMs, specifically from the dimensions that are important for interpreting their utility in prompting methods.
    \item \textbf{Prompt Engineering:} Given that the prompt specifies the task, choosing a proper prompt has a large effect not only on the accuracy, but also on which task the model performs in the first place. 
    In \S\ref{sec:4-prompt-template-engineering} we discuss methods to choose which prompt we should use as $f_{\text{prompt}}(\bm{x})$.
    \item \textbf{Answer Engineering:} Depending on the task, we may want to design $\mathcal{Z}$ differently, possibly along with the mapping function.
    In \S\ref{sec:5-prompt-answer-engineering} we discuss different ways to do so.
    \item \textbf{Expanding the Paradigm:} As stated above, the above equations represent only the simplest of the various underlying frameworks that have been proposed to do this variety of prompting.
    In \S\ref{sec:6-multi-prompt-learning} we discuss ways to expand this underlying paradigm to further improve results or applicability.
    \item \textbf{Prompt-based Training Strategies:} There are also methods to train parameters, either of the prompt, the LM, or both.
    In \S\ref{sec:tuning}, we summarize different strategies and detail their relative advantages.
\end{itemize}

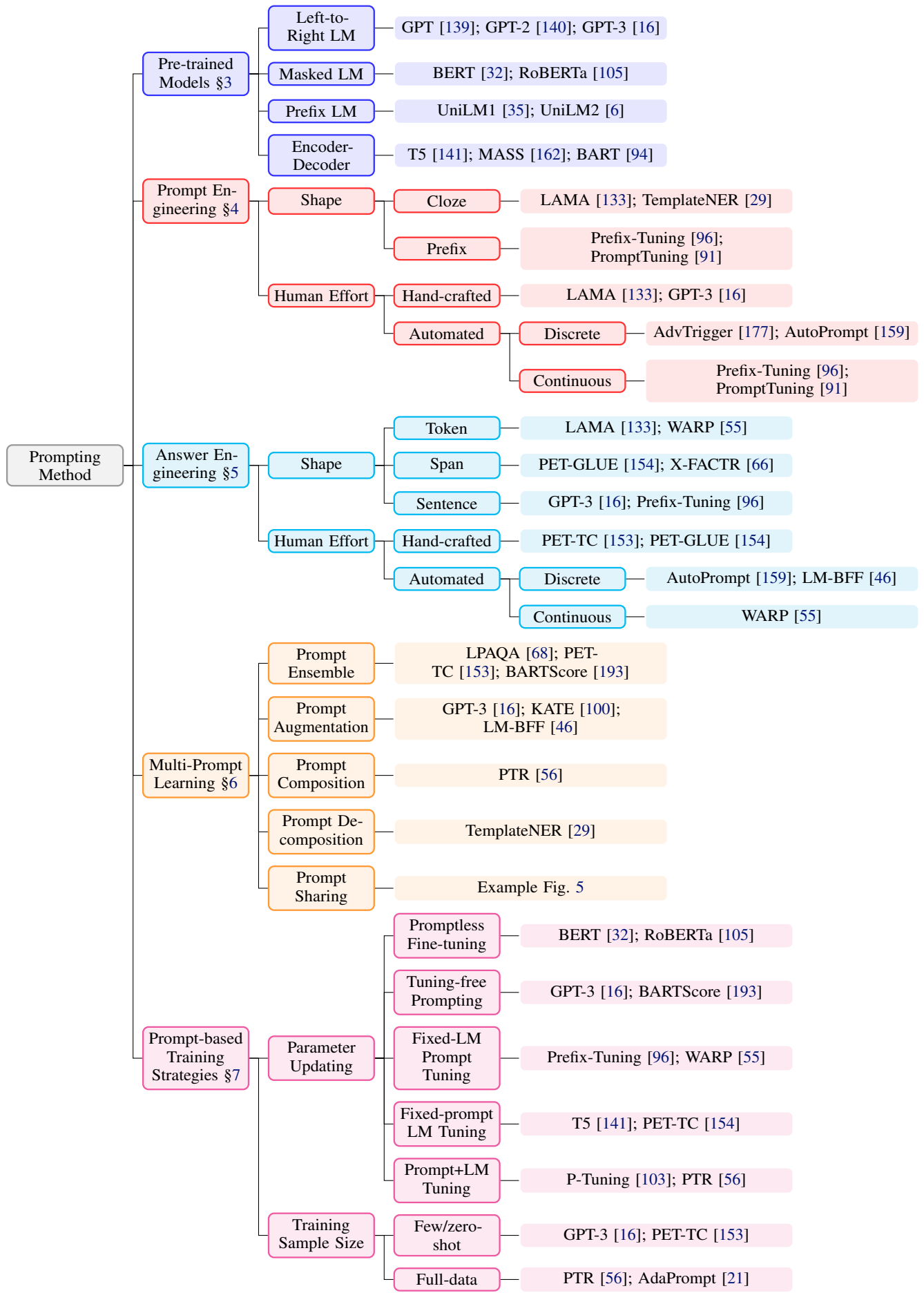
\begin{figure*}
\footnotesize
        \begin{forest}
            for tree={
                forked edges,
                grow'=0,
                draw,
                rounded corners,
                node options={align=center,},
                text width=2.7cm,
                s sep=6pt,
                calign=child edge, calign child=(n_children()+1)/2,
            },
            [Prompting Method, fill=gray!45, parent
                [Pre-trained Models \S\ref{sec:lm}, for tree={ pretrain}
                    [Left-to-Right LM,  pretrain
                        [GPT~\citenumber{Radford2018ImprovingLU}; GPT-2~\citenumber{Radford2019LanguageMA};   GPT-3~\citenumber{brown2020language}, 
                        style = pretrain_work]
                    ]
                    [Masked LM, pretrain
                        [BERT~\citenumber{devlin-etal-2019-bert};  RoBERTa~\citenumber{Liu2019RoBERTaAR} 
                        , pretrain_work]
                    ]
                    [Prefix LM,  pretrain
                        [UniLM1~\citenumber{DBLP:conf/nips/00040WWLWGZH19};  UniLM2~\citenumber{DBLP:conf/icml/Bao0WW0L0GP0H20}  , pretrain_work]                        
                    ]
                    [Encoder-Decoder, pretrain
                        [T5~\citenumber{JMLR:v21:20-074}; MASS~\citenumber{DBLP:conf/icml/SongTQLL19}; BART~\citenumber{lewis-etal-2020-bart}  , pretrain_work]                        
                    ]
                ]
                [Prompt Engineering \S\ref{sec:4-prompt-template-engineering}, for tree={fill=red!45,template}
                    [Shape,  template
                        [Cloze, template
                        [LAMA~\citenumber{petroni-etal-2019-language}; TemplateNER~\citenumber{cui2021templatebased}, template_work]                            
                        ]
                        [Prefix, template
                            [Prefix-Tuning~\citenumber{li2021prefix}; PromptTuning~\citenumber{lester2021power}, template_work]                            
                        ]
                    ]
                    [Human Effort,  template
                        [Hand-crafted, template
                            [LAMA~\citenumber{petroni-etal-2019-language};  GPT-3~\citenumber{brown2020language}, template_work]                            
                        ]
                        [Automated, template
                            [Discrete, template
                                [AdvTrigger~\citenumber{DBLP:conf/emnlp/WallaceFKGS19}; AutoPrompt~\citenumber{autoprompt:emnlp20}, template_work]                                
                            ]
                            [Continuous, template
                                [Prefix-Tuning~\citenumber{li2021prefix};  PromptTuning~\citenumber{lester2021power}, template_work]                                
                            ]
                        ]
                    ]
                ]
                [Answer Engineering \S\ref{sec:5-prompt-answer-engineering}, for tree={fill=blue!45, answer}
                    [Shape, answer
                        [Token, answer
                            [LAMA~\citenumber{petroni-etal-2019-language}; WARP~\citenumber{Hambardzumyan2021WARPWA}, answer_work]                            
                        ]
                        [Span, answer
                            [PET-GLUE~\citenumber{schick2021its}; X-FACTR~\citenumber{jiang-etal-2020-x}, answer_work]                             
                        ]
                        [Sentence, answer
                           [GPT-3~\citenumber{brown2020language}; Prefix-Tuning~\citenumber{li2021prefix}, answer_work]                             
                        ]
                    ]
                    [Human Effort,  answer
                        [Hand-crafted, answer
                            [PET-TC~\citenumber{schick2021exploiting};  PET-GLUE~\citenumber{schick2021its}, answer_work]                             
                        ]
                        [Automated, answer
                            [Discrete, answer
                                [AutoPrompt~\citenumber{autoprompt:emnlp20};  LM-BFF~\citenumber{gao2021making}, answer_work]                                 
                            ]
                            [Continuous, answer
                                [WARP~\citenumber{Hambardzumyan2021WARPWA}, answer_work]                                 
                            ]
                        ]
                    ]
                ]
                [Multi-Prompt Learning \S\ref{sec:6-multi-prompt-learning}, for tree={multiple}
                    [Prompt Ensemble, multiple
                                [LPAQA~\citenumber{jiang-etal-2020-know}; PET-TC~\citenumber{schick2021exploiting}; BARTScore~\citenumber{yuan2021bartscore}, multiple_work]                          
                    ]
                     [Prompt Augmentation, multiple
                                [GPT-3~\citenumber{brown2020language};  KATE~\citenumber{liu2021makes};  LM-BFF~\citenumber{gao2021making}, multiple_work]                         
                    ]
                    [Prompt Composition,  multiple
                                [PTR~\citenumber{han2021ptr}, multiple_work]                         
                    ]
                    [Prompt Decomposition,  multiple
                                [TemplateNER~\citenumber{cui2021templatebased}, multiple_work]                         
                    ]
                    [Prompt Sharing,  multiple 
                    [Example Fig.~\ref{fig:prompt-sharing}, multiple_work]
                    ]                    
                ]   
                [Prompt-based Training Strategies \S\ref{sec:tuning}, for tree={tuning}
                    [Parameter Updating, tuning
                            [Promptless Fine-tuning, tuning
                                [BERT~\citenumber{devlin-etal-2019-bert};  RoBERTa~\citenumber{Liu2019RoBERTaAR}, tuning_work]
                            ]                    
                            [Tuning-free Prompting, tuning
                                [GPT-3~\citenumber{brown2020language}; BARTScore~\citenumber{yuan2021bartscore}, tuning_work]  
                            ]
                            [Fixed-LM Prompt Tuning, tuning
                                [Prefix-Tuning~\citenumber{li2021prefix}; WARP~\citenumber{Hambardzumyan2021WARPWA}, tuning_work]
                            ]
                            [Fixed-prompt LM Tuning, tuning
                                [T5~\citenumber{JMLR:v21:20-074}; PET-TC~\citenumber{schick2021its}, tuning_work]
                            ]
                            [Prompt+LM Tuning, tuning
                                [P-Tuning~\citenumber{liu2021ptuning};  PTR~\citenumber{han2021ptr}, tuning_work]
                            ]
                    ]
                    [Training Sample Size,  tuning
                            [Few/zero-shot, tuning
                                [GPT-3~\citenumber{brown2020language};  PET-TC~\citenumber{schick2021exploiting}, tuning_work]
                            ]
                            [Full-data, tuning
                                [PTR~\citenumber{han2021ptr};  AdaPrompt~\citenumber{DBLP:journals/corr/abs-2104-07650}, tuning_work]
                            ]  
                    ]
                ]                 
            ]
        \end{forest}
            \caption{Typology of prompting methods.}
            \label{fig:typo-prompt}
\end{figure*}

\clearpage

\section{Pre-trained Language Models}
\label{sec:lm}

Given the large impact that pre-trained LMs have had on NLP in the pre-train and fine-tune paradigm, there are already a number of high-quality surveys that interested readers where interested readers can learn more \cite{JMLR:v21:20-074,qiu2020pre,xu2021pre,doddapaneni2021primer}.
Nonetheless, in this chapter we present a systematic view of various pre-trained LMs which (i) organizes them along various axes in a more systematic way, (ii) particularly focuses on aspects salient to prompting methods.
Below, we will detail them through the lens of \textit{main training objective}, \textit{type of text noising}, \textit{auxiliary training objective}, \textit{attention mask}, \textit{typical architecture}, and \textit{preferred application scenarios}.
We describe each of these objectives below, and also summarize a number of pre-trained LMs along each of these axes in Tab.~\ref{tab:pretrained-aspect} in the appendix.

\subsection{Training Objectives}

The main training objective of a pre-trained LM almost invariably consists of some sort of objective predicting the probability of text $\bm{x}$.

\paragraph{Standard Language Model (SLM)} objectives do precisely this, training the model to optimize the probability $P(\bm{x})$ of text from a training corpus~\citep{Radford2019LanguageMA}.
In these cases, the text is generally predicted in an \term{autoregressive} fashion, predicting the tokens in the sequence one at a time.
This is usually done from left to right (as detailed below), but can be done in other orders as well.

\paragraph{}
A popular alternative to standard LM objectives are \term{denoising} objectives, which apply some noising function $\tilde{\bm{x}} = f_{\text{noise}}(\bm{x})$ to the input sentence (details in the following subsection), then try to predict the original input sentence given this noised text $P(\bm{x}|\tilde{\bm{x}})$.
There are two common flavors of these objectives:

\paragraph{Corrupted Text Reconstruction (CTR)}
These objectives restore the processed text to its uncorrupted state by calculating loss over \emph{only} the noised parts of the input sentence.

\paragraph{Full Text Reconstruction (FTR)}
These objectives reconstruct the text by calculating the loss over the \emph{entirety} of the input texts whether it has been noised or not~\citep{lewis-etal-2020-bart}. 

\paragraph{}
The main training objective of the pre-trained LMs plays an important role in determining its applicability to particular prompting tasks.
For example, left-to-right autoregressive LMs may be particularly suitable for prefix prompts, whereas reconstruction objectives may be more suitable for cloze prompts.
In addition, models trained with standard LM and FTR objectives may be more suitable for tasks regarding text generation, whereas other tasks such as classification can be formulated using models trained with any of these objectives. 

In addition to the main training objectives above, a number of \term{auxiliary objectives} have been engineered to further improve models' ability to perform certain varieties of downstream tasks.
We list some commonly-used auxiliary objectives in Appendix~\ref{appendix:sec:auxiliary}.

\subsection{Noising Functions}
In training objectives based on reconstruction, the specific type of corruption applied to obtain the noised text $\tilde{\bm{x}}$ has an effect on the efficacy of the learning algorithm.
In addition, prior knowledge can be incorporated by controlling the type of noise, e.g.~the noise could focus on entities of a sentence, which allows us to learn a pre-trained model with particularly high predictive performance for entities.
In the following, we introduce several types of noising functions, and give detailed examples in Tab.~\ref{tab:corruption-exp}.

\begin{CJK}{UTF8}{gbsn}

\begin{table*}[!htb]
  \centering \footnotesize
    \begin{tabular}{llll}
    \toprule
    \textbf{Operation} & \textbf{Element} & \textbf{Original Text} & \textbf{Corrupted Text} \\
    \midrule
    \multirow{3}[1]{*}{Mask} & one token & Jane will \textcolor{red}{move} to New York . & Jane will \texttt{[Z]} to New York . \\
          & two tokens & Jane will \textcolor{red}{move to} New York . & Jane will \texttt{[Z]} \texttt{[Z]} New York . \\
          & one entity & Jane will move to \textcolor{red}{New York} . & Jane will move to \texttt{[Z]} . \\
          \midrule
    \multirow{3}[0]{*}{Replace} & one token & Jane will move \textcolor{red}{to} New York . & Jane will move \texttt{[X]} New York . \\
          & two tokens & Jane will move \textcolor{red}{to New} York . & Jane will move \texttt{[X]} \texttt{[Y]} York . \\
          & one entity & Jane will move to \textcolor{red}{New York} . & Jane will move to \texttt{[X]} . \\
          \midrule
    \multirow{2}[0]{*}{Delete} & one token & Jane \sout{\textcolor{red}{will}} move to New York . & Jane move to New York . \\
          & two token & Jane \sout{\textcolor{red}{will move}} to New York . & Jane to New York . \\
          \midrule
    Permute & token  & Jane will move to New York . & New York . Jane will move to  \\
    Rotate & none  & Jane will move to New York . &  to New York . Jane will move \\
    Concatenation & two languages & Jane will move to New York . & Jane will move to New York . \texttt{[/s]} 简将搬到纽约。 \\
    \bottomrule
    \end{tabular}%
    \caption{Detailed examples for different noising operations.}
  \label{tab:corruption-exp}%
\end{table*}%

\end{CJK}


\paragraph{Masking} (e.g.~\citet{devlin-etal-2019-bert})  The text will be masked in different levels, replacing a token or multi-token span with a special token such as \texttt{[MASK]}.
Notably, masking can either be random from some distribution or specifically designed to introduce prior knowledge, such as the above-mentioned example of masking entities to encourage the model to be good at predicting entities.

\paragraph{Replacement} (e.g.~\citet{JMLR:v21:20-074}) 
Replacement is similar to masking, except that the token or multi-token span is not replaced with a \texttt{[MASK]} but rather another token or piece of information (e.g., an image region~\cite{DBLP:conf/iclr/SuZCLLWD20}).


\paragraph{Deletion}
(e.g.~\citet{lewis-etal-2020-bart}) 
Tokens or multi-token spans will be deleted from a text without the addition of \texttt{[MASK]} or any other token. This operation is usually used together with the FTR loss.

\paragraph{Permutation}
(e.g.~\citet{DBLP:journals/tacl/LiuGGLEGLZ20})
The text is first divided into different spans (tokens, sub-sentential spans, or sentences), and then these spans are be permuted into a new text.

\subsection{Directionality of Representations}

A final important factor that should be considered in understanding pre-trained LMs and the difference between them is the directionality of the calculation of representations.
In general, there are two widely used ways to calculate such representations:

\paragraph{Left-to-Right}
The representation of each word is calculated based on the word itself and all previous words in the sentence.
For example, if we have a sentence ``This is a good movie'', the representation of the word ``good'' would be calculated based on previous words.
This variety of factorization is particularly widely used when calculating standard LM objectives or when calculating the output side of an FTR objective, as we discuss in more detail below.

\paragraph{Bidirectional}
The representation of each word is calculated based on all words in the sentence, including words to the left of the current word.
In the example above, ``good'' would be influenced by all words in the sentence, even the following ``movie''.

\begin{figure}[h]
    \centering
    \subfloat[Full.]{
    \includegraphics[height=0.17\linewidth]{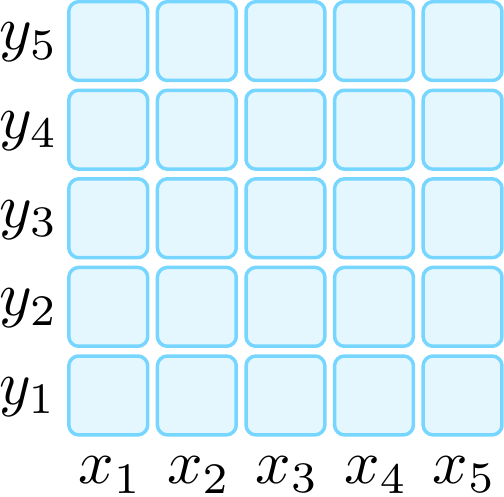}
    }         \hspace{3.2em}
    \subfloat[Diagonal.]{  
    \includegraphics[height=0.17\linewidth]{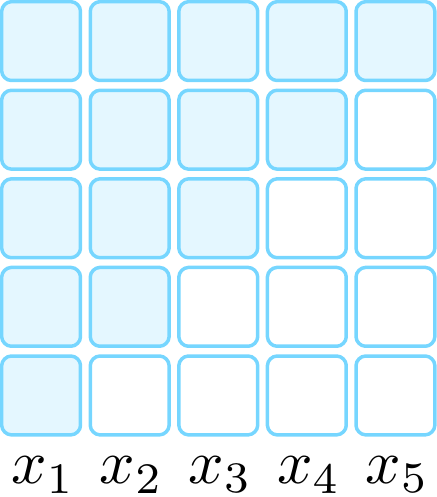}
    }        \hspace{3.2em}
    \subfloat[Mixture.]{ 
    \includegraphics[height=0.17\linewidth]{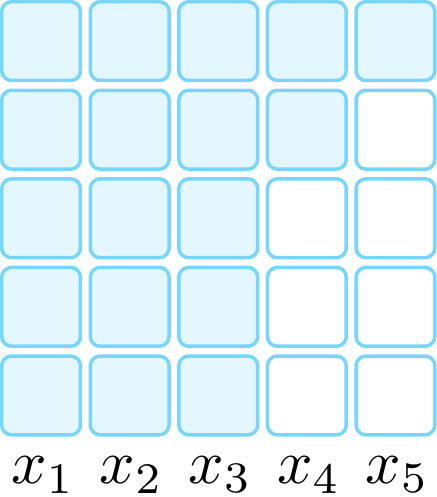}
        } 
    \caption{Three popular attention mask patterns, where the subscript $t$ indicates the $t$-th timestep. 
    A shaded box at $(i, j)$ indicates that the attention mechanism is allowed to attend to the input element $i$ at output time step $j$. A white box indicates that the attention mechanism is not allowed to attend to the corresponding $i$ and $j$ combination.}
    \label{fig:attn_mask}
\end{figure}

\paragraph{}
In addition to the two most common directionalities above, it is also possible to mix the two strategies together in a single model \citep{DBLP:conf/nips/00040WWLWGZH19,DBLP:conf/icml/Bao0WW0L0GP0H20}, or perform conditioning of the representations in a randomly permuted order \citep{DBLP:conf/nips/YangDYCSL19}, although these strategies are less widely used.
Notably, when implementing these strategies within a neural model, this conditioning is generally implemented through \term{attention masking}, which masks out the values in an attentional model \citep{bahdanau2014neural}, such as the popular Transformer architecture~\cite{vaswani2017attention}.
Some examples of such attention masks are shown in Figure~\ref{fig:attn_mask}.

\subsection{Typical Pre-training Methods} \label{sec:typical-architecture}

\begin{figure*}[!t]
    \centering
    
    \subfloat[Left-to-right LM.]{
    \includegraphics[height=0.18\linewidth]{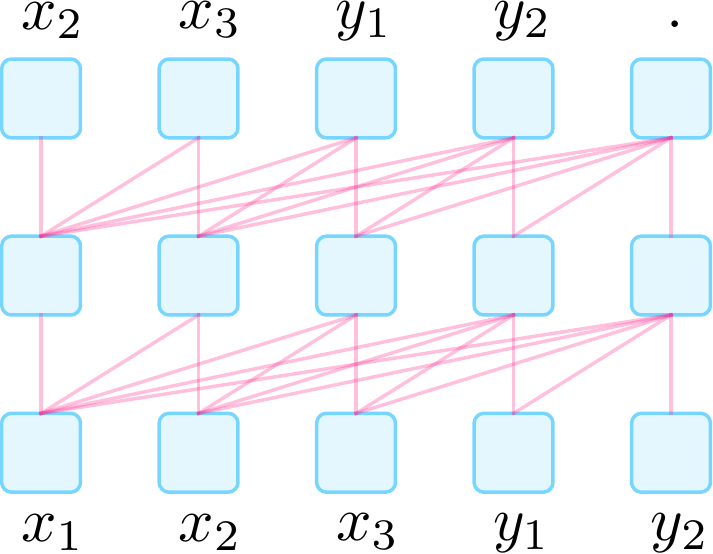}
    }         \hspace{1.3em}
    \subfloat[Masked LM.]{  
    \includegraphics[height=0.18\linewidth]{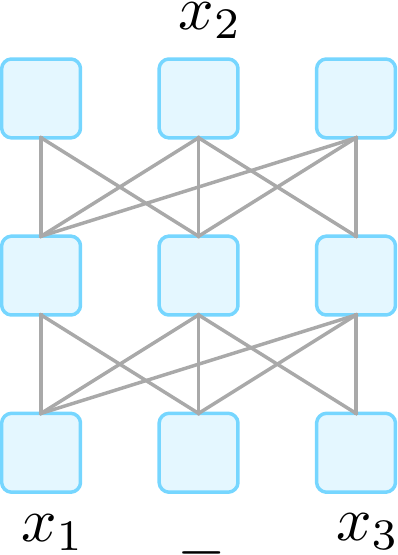}
    }        \hspace{1.3em}
    \subfloat[Prefix LM.]{ 
    \includegraphics[height=0.18\linewidth]{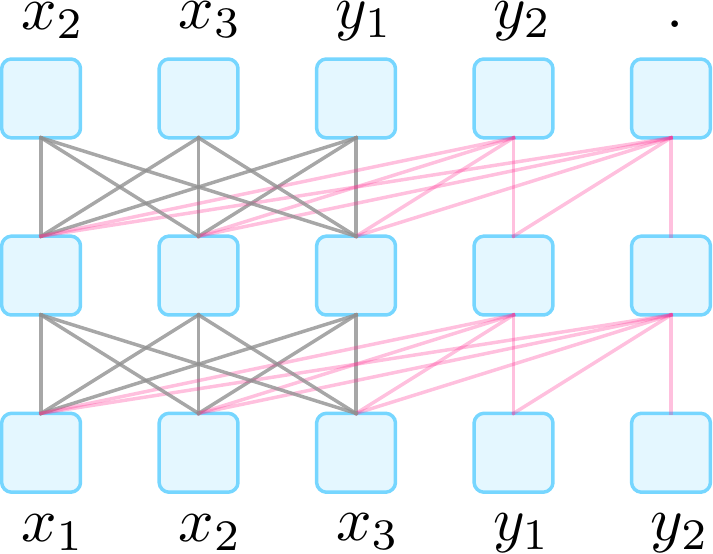}
        }  \hspace{1.3em}
    \subfloat[Encoder-Decoder.]{
    \includegraphics[height=0.18\linewidth]{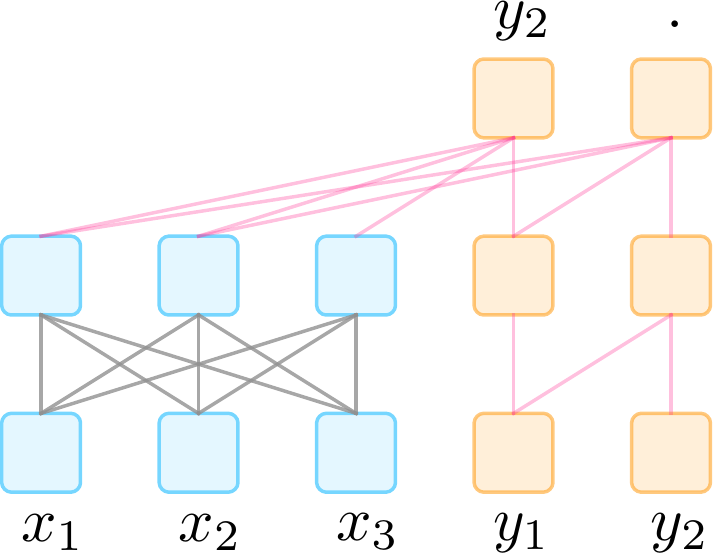}
    }      
    \caption{Typical paradigms of pre-trained LMs.}
    \label{fig:four-architecture}
\end{figure*}

With the above concepts in mind, we introduce four popular pre-training methods, resulting from diverse combinations of objective, noising function, and directionality.
These are described below, and summarized in Fig.~\ref{fig:four-architecture} and Tab.~\ref{tab:four-architecture}.

\subsubsection{Left-to-Right Language Model}

Left-to-right LMs (L2R LMs), a variety of \term{auto-regressive LM}, predict the upcoming words or assign a probability $P(\bm{x})$ to a sequence of words $\bm{x} = x_1, \cdots, x_n$ \cite{jurafskyspeech}.
The probability is commonly broken down using the chain rule in a left-to-right fashion: $P(\bm{x}) = P(x_1)  \times \cdots P(x_n|x_1\cdots x_{n-1})$.\footnote{Similarly, a right-to-left LM can predict preceding words based on the future context, such as $P(x_i|x_{i+1},\cdots,x_n)$.}

\begin{myboxi}[Example \& Applicable Scenario]
Left-to-right LMs have been standard since their proposal by Markov in 1913 \cite{markov2006example}, and have been used continuously since then in both count-based \citep{goodman2001bit} and neural forms \citep{bengio2003neural,mikolov2010recurrent,Radford2018ImprovingLU}.
Representative examples of modern pre-trained left-to-right LMs include GPT-3 \citep{brown2020language}, 
and GPT-Neo \citep{gpt-neo}.

L2R pre-trained LMs are also the popular backbone that many prompting methods adopt~\cite{Radford2019LanguageMA,brown2020language}
. 
One practical reason for this is that many such models are large (PanGu-$\alpha$~\citep{zeng2021pangualpha}, Ernie-3~\citep{DBLP:journals/corr/abs-2107-02137}) and ponderous to train, or not even available publicly.
Thus using these models in the pre-train and fine-tune regimen is often not possible.
\end{myboxi}


\begin{table*}[!htbp]
  \centering \footnotesize
    \begin{tabular}{lcccccll}
    \toprule
    \multirow{2}[2]{*}{\textbf{LMs}} & \multicolumn{3}{c}{$\bm{x}$} & \multicolumn{3}{c}{$\bm{y}$}  & \multirow{2}[2]{*}{\textbf{Application}}\\
    \cmidrule(lr){2-4} \cmidrule(lr){5-7}
          & \textbf{Mask} & \textbf{Noise} & \textbf{Main Obj.} & \textbf{Mask} & \textbf{Noise} & \textbf{Main Obj.} \\
    \midrule
    L2R   &  Diagonal  & None  & SLM   & -  & -  & - & NLU \& NLG\\
    Mask  & Full & Mask  & CTR   & -     & -     & -  & NLU\\
    Prefix& Full & Any   & CTR   & Diagonal  & None  & SLM & NLU \& NLG \\
    En-De & Full & Any   & None$\dag$  & Diagonal  & None  & FTR/CRT & NLU \& NLG \\
    \bottomrule
    \end{tabular}%
    \caption{Typical architectures for pre-trained LMs. $\bm{x}$ and $\bm{y}$ represent text to be encoded and decoded, respectively. 
    \textbf{SLM}: Standard language model.
    \textbf{CTR}: Corrupted text reconstruction.
    \textbf{FTR}: Full text reconstruction.
    $\dag$: Encoder-decoder architectures usually apply objective functions to the decoder only.
    }
      \label{tab:four-architecture}%
\end{table*}%

\subsubsection{Masked Language Models}

While autoregressive language models provide a powerful tool for modeling the probability of text, they also have disadvantages such as requiring representations be calculated from left-to-right.
When the focus is shifted to generating the optimal representations for down-stream tasks such as classification, many other options become possible, and often preferable.
One popular bidirectional objective function used widely in representation learning is the \term{masked language model}~(MLM; \citet{devlin-etal-2019-bert}), which aims to
predict masked text pieces based on surrounded context. For example, $P(x_i|x_1,\ldots,x_{i-1},x_{i+1},\ldots,x_n)$ represents the probability of the word $x_i$ given the surrounding context.

\begin{myboxi}[Example \& Applicable Scenario]
Representative pre-trained models using MLMs include: BERT~\cite{devlin-etal-2019-bert}, ERNIE~\cite{DBLP:conf/acl/ZhangHLJSL19,sun2019ernie} and many variants.
In prompting methods, MLMs are generally most suitable for natural language understanding or analysis tasks (e.g., text classification, natural language inference 
, and extractive question answering).
These tasks are often relatively easy to be reformulated into cloze problems, which are consistent with the training objectives of the MLM.
Additionally, MLMs have been a pre-trained model of choice when exploring methods that combine prompting with fine-tuning, elaborated further in \S\ref{sec:tuning}.
\end{myboxi}

\subsubsection{Prefix and Encoder-Decoder}

For conditional text generation tasks such as translation and summarization where an input text $\bm{x} = x_1, \cdots, x_n$ is given and the goal is to generate target text $\bm{y}$, we need a pre-trained model that is both capable of encoding the input text and generating the output text.
There are two popular architectures for this purpose that share a common thread of (1) using an encoder with fully-connected mask to encode the source $\bm{x}$ first and then (2) decode the target $\bm{y}$ auto-regressively (from the left to right).

\paragraph{Prefix Language Model}
The prefix LM is a left-to-right LM that decodes $\bm{y}$ conditioned on a prefixed sequence $\bm{x}$, which is encoded by the \textit{same} model parameters but with a fully-connected mask.
Notably, to encourage the prefix LM to learn better representations of the input, a corrupted text reconstruction objective is usually applied over $\bm{x}$, in addition to a standard conditional language modeling objective over $\bm{y}$.

\paragraph{Encoder-decoder}
The encoder-decoder model is a model that uses a left-to-right LM to decode $\bm{y}$ conditioned on a \textit{separate} encoder for text $\bm{x}$ with a fully-connected mask; the parameters of the encoder and decoder are not shared.
Similarly to the prefix LM, diverse types of noising can be applied to the input $\bm{x}$.

\begin{myboxi}[Example \& Applicable Scenario]

Prefix LMs have been explored in UniLM 1-2 \cite{DBLP:conf/nips/00040WWLWGZH19,DBLP:conf/icml/Bao0WW0L0GP0H20} and ERNIE-M~\cite{DBLP:journals/corr/abs-2012-15674} while encoder-decoder models are widely used in pre-trained models such as T5~\cite{JMLR:v21:20-074}, BART~\cite{lewis-etal-2020-bart}, MASS~\cite{DBLP:conf/icml/SongTQLL19} and their variants.

Pre-trained models with prefix LMs and encoder-decoder paradigms can be naturally used to text generation tasks with \cite{dou-etal-2021-gsum} or without \cite{yuan2021can,liu2021simcls} prompting using input texts.
However, recent studies reveal that other non-generation tasks, such as information extraction \cite{cui2021templatebased}, question answering  \cite{khashabi-etal-2020-unifiedqa}
, and text generation evaluation  \cite{yuan2021bartscore} can be reformulated a generation problems by providing appropriate prompts.
Therefore, prompting methods (i) broaden the applicability of these generation-oriented pre-trained models. For example, pre-trained models like BART are less used in NER while prompting methods make BART applicable,
and (ii) breaks the difficulty of unified modelling among different tasks~\cite{khashabi-etal-2020-unifiedqa}. 

\end{myboxi}

\section{Prompt Engineering}
\label{sec:4-prompt-template-engineering}

\term{Prompt engineering} is the process of creating a prompting function $f_{\text{prompt}}(\bm{x})$ that results in the most effective performance on the downstream task.
In many previous works, this has involved \term{prompt template engineering}, where a human engineer or algorithm searches for the best template for each task the model is expected to perform.
As shown in the ``Prompt Engineering'' section of Fig.\ref{fig:typo-prompt}, one must first consider the \term{prompt shape}, and then decide whether to take a \term{manual} or \term{automated} approach to create prompts of the desired shape, as detailed below.

\subsection{Prompt Shape}
As noted above, there are two main varieties of prompts: \term{cloze prompts} \cite{petroni-etal-2019-language,cui2021templatebased}, which fill in the blanks of a textual string, and \term{prefix prompts} \cite{li2021prefix,lester2021power}, which continue a string prefix.
Which one is chosen will depend both on the task and the model that is being used to solve the task.
In general, for tasks regarding generation, or tasks being solved using a standard auto-regressive LM, prefix prompts tend to be more conducive, as they mesh well with the left-to-right nature of the model. For tasks that are solved using masked LMs, cloze prompts are a good fit, as they very closely match the form of the pre-training task.
Full text reconstruction models are more versatile, and can be used with either cloze or prefix prompts.
Finally, for some tasks regarding multiple inputs such as \term{text pair classification}, prompt templates must contain space for two inputs, \texttt{[X1]} and \texttt{[X2]}, or more.



\subsection{Manual Template Engineering}
Perhaps the most natural way to create prompts is to manually create intuitive templates based on human introspection.
For example, the seminal LAMA dataset \citep{petroni-etal-2019-language} provides manually created cloze templates to probe knowledge in LMs.
\citet{brown2020language} create manually crafted prefix prompts to handle a wide variety of tasks, including question answering, translation, and probing tasks for common sense reasoning.
\citet{schick2020fewshot, schick2021exploiting, schick2021its} use pre-defined templates in a few-shot learning setting on text classification and conditional text generation tasks.

\subsection{Automated Template Learning}
While the strategy of manually crafting templates is intuitive and does allow solving various tasks with some degree of accuracy, there are also several issues with this approach: (1) creating and experimenting with these prompts is an art that takes time and experience, particularly for some complicated tasks such as semantic parsing \citep{Shin2021ConstrainedLM};
(2) even experienced prompt designers may fail to manually discover optimal prompts \cite{jiang-etal-2020-know}.

To address these problems, a number of methods have been proposed to automate the template design process.
In particular, the automatically induced prompts can be further separated into \term{discrete prompts}, where the prompt is an actual text string, and \term{continuous prompts}, where the prompt is instead described directly in the embedding space of the underlying LM.

One other orthogonal design consideration is whether the prompting function $f_{\text{prompt}}(\bm{x})$ is \term{static}, using essentially the same prompt template for each input, or \term{dynamic}, generating a custom template for each input.
Both static and dynamic strategies have been used for different varieties of discrete and continuous prompts, as we will mention below.

\subsubsection{Discrete Prompts}

Works on discovering \term{discrete prompts} (a.k.a~\term{hard prompts}) automatically search for templates described in a discrete space, usually corresponding to natural language phrases.
We detail several methods that have been proposed for this below:

\paragraph{D1: Prompt Mining}
\citet{jiang-etal-2020-know}'s \textsc{Mine} approach is a mining-based method to automatically find templates given a set of training inputs $\bm{x}$ and outputs $\bm{y}$.
This method scrapes a large text corpus (e.g. Wikipedia) for strings containing $\bm{x}$ and $\bm{y}$, and finds either the \term{middle words} or \term{dependency paths} between the inputs and outputs.
Frequent middle words or dependency paths can serve as a template as in ``\texttt{[X]} middle words \texttt{[Z]}''.

\paragraph{D2: Prompt Paraphrasing}
Paraphrasing-based approaches take in an existing seed prompt (e.g.~manually constructed or mined), and paraphrases it into a set of other candidate prompts, then selects the one that achieves the highest training accuracy on the target task.
This paraphrasing can be done in a number of ways, including using round-trip translation of the prompt into another language then back \citep{jiang-etal-2020-know}, using replacement of phrases from a thesaurus \citep{yuan2021bartscore}, or using a neural prompt rewriter specifically optimized to improve accuracy of systems using the prompt \citep{haviv-etal-2021-bertese}.
Notably, \citet{haviv-etal-2021-bertese} perform paraphrasing \emph{after} the input $\bm{x}$ is input into the prompt template, allowing a different paraphrase to be generated for each individual input.

\paragraph{D3: Gradient-based Search} \citet{DBLP:conf/emnlp/WallaceFKGS19} applied a gradient-based search over actual tokens to find short sequences that can trigger the underlying pre-trained LM to generate the desired target prediction. This search is done in an iterative fashion, stepping through tokens in the prompt 
.
Built upon this method, \citet{autoprompt:emnlp20} automatically search for template tokens using downstream application training samples and demonstrates strong performance in prompting scenarios.

\paragraph{D4: Prompt Generation}
Other works treat the generation of prompts as a text generation task and use standard natural language generation models to perform this task.
For example, \citet{gao2021making} introduce the seq2seq pre-trained model T5 into the template search process. Since T5 has been pre-trained on a task of filling in missing spans, they use T5 to generate template tokens by (1) specifying the position to insert template tokens within a template\footnote{The number of template tokens do not need to be pre-specified since T5 can decode multiple tokens at a masked position.} (2) provide training samples for T5 to decode template tokens.
\citet{bendavid2021pada} propose a domain adaptation algorithm that trains T5 to generate unique domain relevant features (DRFs; a set of keywords that characterize domain information) for each input. Then those DRFs can be concatenated with the input to form a template and be further used by downstream tasks.



\paragraph{D5: Prompt Scoring}
\citet{DBLP:conf/emnlp/DavisonFR19} investigate the task of knowledge base completion and design a template for an input (head-relation-tail triple) using LMs.
They first hand-craft a set of templates as potential candidates, and fill the input and answer slots to form a filled prompt.
They then use a unidirectional LM to score those filled prompts, selecting the one with the highest LM probability.
This will result in custom template for each individual input.

\subsubsection{Continuous Prompts}
\label{sec:4-continuous}
Because
the purpose of prompt construction is to find a method that allows an LM to effectively perform a task, rather than being for human consumption, it is not necessary to limit the prompt to human-interpretable natural language. 
Because of this, there are also methods that examine \term{continuous prompts} (a.k.a.~\term{soft prompts}) that perform prompting directly in the embedding space of the model.
Specifically, continuous prompts remove two constraints:
(1) relax the constraint that the embeddings of template words be the embeddings of natural language (e.g., English) words.
(2) Remove the restriction that the template is parameterized by the pre-trained LM's parameters. Instead, templates have their own parameters that can be tuned based on training data from the downstream task.
We highlight several representative methods below.

\paragraph{C1: {Prefix Tuning}} Prefix Tuning \citep{li2021prefix} is a method that prepends a sequence of continuous task-specific vectors to the input, while keeping the LM parameters frozen. Mathematically, this consists of optimizing over the following log-likelihood objective given a trainable prefix matrix $M_{\phi}$ and a fixed pre-trained LM parameterized by $\theta$.
\begin{equation}
\label{eq:prefix-tuning}
    \max_{\phi} \log P(\bm{y}|\bm{x}; \theta; \phi) = \max_{\phi} \sum_{y_i} \log P(y_i | h_{< i}; \theta; \phi) 
\end{equation}
In Eq.~\ref{eq:prefix-tuning}, $h_{<i} = [h_{<i}^{(1)}; \cdots; h_{<i}^{(n)}]$ is the concatenation of all neural network layers at time step $i$. It is copied from $M_{\phi}$ directly if the corresponding time step is within the prefix ($h_i$ is $M_{\phi}[i]$), otherwise it is computed using the pre-trained LM.

Experimentally, \citet{li2021prefix} observe that such continuous prefix-based learning is more sensitive to different initialization in low-data settings than the use of discrete prompts with real words. 
Similarly, \citet{lester2021power} prepend the input sequence with special tokens to form a template and tune the embeddings of these tokens directly. Compared to \citet{li2021prefix}'s method, this adds fewer parameters as it doesn't introduce additional tunable parameters within each network layer.
\citet{DBLP:journals/corr/abs-2106-13884} train a vision encoder that encodes an image into a sequence of embeddings that can be used to prompt a frozen auto-regressive LM to generate the appropriate caption. They show that the resulting model can perform few-shot learning for vision-language tasks such as visual question answering etc. Different from the above two works, the prefix used in \cite{DBLP:journals/corr/abs-2106-13884} is sample-dependent, namely a representation of input images, instead of a task embedding.

\paragraph{C2: Tuning Initialized with Discrete Prompts}
There are also methods that initialize the search for a continuous prompt using a prompt that has already been created or discovered using discrete prompt search methods.
For example, \citet{zhong2021optiprompt} first define a template using a discrete search method such as \textsc{AutoPrompt}~\cite{autoprompt:emnlp20}'s, initialize virtual tokens based on this discovered prompt, then fine-tune the embeddings to increase task accuracy.
This work found that initializing with manual templates can provide a better starting point for the search process.
\citet{qin-eisner-2021-learning} propose to learn a mixture of soft templates for each input where the weights and parameters for each template are jointly learned using training samples.
The initial set of templates they use are either manually crafted ones or those obtained using the  ``prompt mining'' method.
Similarly, \citet{Hambardzumyan2021WARPWA} introduce the use of a continuous template whose shape follows a manual prompt template. 

\paragraph{C3: Hard-Soft Prompt Hybrid Tuning}
Instead of using a purely learnable prompt template, these methods insert some tunable embeddings into a hard prompt template.
\citet{liu2021ptuning} propose ``P-tuning", where continuous prompts are learned by inserting trainable variables into the embedded input.
To account for interaction between prompt tokens, they represent prompt embeddings as the output of a BiLSTM \citep{6638947}.
P-tuning also introduces the use of task-related anchor tokens (such as ``capital" in relation extraction) 
within the template for further improvement. These anchor tokens are not tuned during training.
\citet{han2021ptr} propose prompt tuning with rules (PTR), which uses manually crafted sub-templates to compose a complete template using logic rules. To enhance the representation ability of the resulting template, they also insert several virtual tokens whose embeddings can be tuned together with the pre-trained LMs parameters using training samples. The template tokens in \textsc{PTR} contain both actual tokens and virtual tokens. Experiment results demonstrate the effectiveness of this prompt design method in relation classification tasks.

\section{Answer Engineering} \label{sec:5-prompt-answer-engineering}

In contrast to prompt engineering, which designs appropriate inputs for prompting methods,
\term{answer engineering} aims to search for an answer space $\mathcal{Z}$ and a map to the original output $\mathcal{Y}$ that results in an effective predictive model.
Fig.\ref{fig:typo-prompt}'s ``Answer Engineering'' section illustrates two dimensions that must be considered when performing answer engineering: deciding the \term{answer shape} and choosing an \term{answer design method}.





\subsection{Answer Shape}

The shape of an answer characterizes its granularity.
Some common choices include:
\begin{itemize}
    \item \textbf{Tokens:} One of the tokens in the pre-trained LM's vocabulary, or a subset of the vocabulary.
    \item \textbf{Span:} A short multi-token span. These are usually used together with cloze prompts.
    \item \textbf{Sentence:} A sentence or document. These are commonly used with prefix prompts.
\end{itemize}
In practice, how to choose the shape of acceptable answers depends on the task we want to perform. 
Token or text-span  answer spaces are widely used in classification tasks (e.g.~sentiment classification; \citet{DBLP:conf/emnlp/YinHR19}), but also other tasks such as relation extraction \citep{petroni-etal-2019-language} or named entity recognition \citep{cui2021templatebased}.
Longer phrasal or sentential answers are often used in language generation tasks \citep{Radford2019LanguageMA}, but also used in other tasks such as multiple-choice question answering (where the scores of multiple phrases are compared against each-other; \citet{khashabi-etal-2020-unifiedqa}).


\subsection{Answer Space Design Methods}

The next question to answer is how to design the appropriate answer space $\mathcal{Z}$, as well as the mapping to the output space $\mathcal{Y}$ if the answers are not used as the final outputs.

\subsubsection{Manual Design}
In manual design, the space of potential answers $\mathcal{Z}$ and its mapping to $\mathcal{Y}$ are crafted manually by an interested system or benchmark designer.
There are a number of strategies that can be taken to perform this design.

\paragraph{Unconstrained Spaces}
In many cases, the answer space $\mathcal{Z}$ is the space of all tokens \citep{petroni-etal-2019-language}, fixed-length spans \citep{jiang-etal-2020-x}, or token sequences \citep{Radford2019LanguageMA}.
In these cases, it is most common to directly map answer $\bm{z}$ to the final output $\bm{y}$ using the identity mapping.

\paragraph{Constrained Spaces}
However, there are also cases where the space of possible outputs is constrained.
This is often performed for tasks with a limited label space such as text classification or entity recognition, or multiple-choice question answering.
To give some examples, \citet{DBLP:conf/emnlp/YinHR19} manually design lists of words relating to relevant topics (``health'', ``finance'', ``politics'',
``sports'', etc.), emotions (``anger'', ``joy'', ``sadness'',
``fear'', etc.), or other aspects of the input text to be classified.
\citet{cui2021templatebased} manually design lists such as ``person'', ``location'', etc. for NER tasks.
In these cases, it is necessary to have a mapping between the answer $\mathcal{Z}$ and the underlying class $\mathcal{Y}$.

With regards to multiple-choice question answering, it is common to use an LM to calculate the probability of an output among multiple choices, with \citet{zweig-etal-2012-computational} being an early example.




\subsubsection{Discrete Answer Search}
\label{sec:discrete-answer-search}
As with manually created prompts, it is possible that manually created answers are sub-optimal for getting the LM to achieve ideal prediction performance.
Because of this, there is some work on automatic answer search, albeit less than that on searching for ideal prompts.
These work on both discrete answer spaces (this section) and continuous answer spaces (the following).

\paragraph{Answer Paraphrasing}
These methods start with an initial answer space $\mathcal{Z}'$, and then use paraphrasing to expand this answer space to broaden its coverage \citep{jiang2020when}.
Given a pair of answer and output $\langle \bm{z}', \bm{y} \rangle$, we define a function that generates a paraphrased set of answers $\text{para}(\bm{z}')$.
The probability of the final output is then defined as the marginal probability \emph{all} of the answers in this paraphrase set $P(\bm{y}|\bm{x})=\sum_{\bm{z} \in \text{para}(\bm{z'})} P(\bm{z}|\bm{x})$.
This paraphrasing can be performed using any method, but \citet{jiang2020when} specifically use a back-translation method, first translating into another language then back to generate a list of multiple paraphrased answers.

\paragraph{Prune-then-Search}
In these methods, first, an initial pruned answer space of several plausible answers $\mathcal{Z}'$ is generated, and then an algorithm further searches over this pruned space to select a final set of answers.
Note that in some of the papers introduced below, they define a function from label $\bm{y}$ to a single answer token $\bm{z}$, which is often called a \term{verbalizer} \citep{schick2021exploiting}.
\citet{schick2021exploiting,DBLP:conf/coling/SchickSS20} find tokens containing at least two alphabetic characters that are frequent in a large unlabeled dataset.
In the search step, they iteratively compute a word's suitability as a representative answer $\bm{z}$ for a label $\bm{y}$ by maximizing the likelihood of the label over training data.
 \citet{autoprompt:emnlp20} learn a logistic classifier using the contextualized representation of the \texttt{[Z]} token as input. In the search step, they select the top-$k$ tokens that achieve the highest probability score using the learned logistic classifier in the first step. Those selected tokens will form the answer.
\citet{gao2021making} first construct a pruned search space $\mathcal{Z}'$ by selecting top-$k$ vocabulary words based on their generation probability at the \texttt{[Z]} position determined by training samples. Then the search space is further pruned down by only selecting a subset of words within $\mathcal{Z}'$ based on their zero-shot accuracy on the training samples. (2) In the search step, they fine-tune the LM with fixed templates together with every answer mapping using training data and select the best label word as the answer based on the accuracy on the development set. 



\paragraph{Label Decomposition}
When performing relation extraction, \citet{DBLP:journals/corr/abs-2104-07650} automatically decompose each relation label into its constituent words and use them as an answer. For example, for the relation \texttt{per:city\_of\_death}, the decomposed label words would be \{\texttt{person}, \texttt{city}, \texttt{death}\}. The probability of the answer span will be calculated as the sum of each token's probability. 

\subsubsection{Continuous Answer Search}
Very few works explore the possibility of using soft answer tokens which can be optimized through gradient descent.
\citet{Hambardzumyan2021WARPWA} assign a virtual token for each class label and optimize the token embedding for each class together with prompt token embeddings. Since the answer tokens are optimized directly in the embedding space, they do not make use of the embeddings learned by the LM and instead learn an embedding from scratch for each label.

\section{Multi-Prompt Learning} \label{sec:6-multi-prompt-learning}

\begin{figure*}[t]
    \centering
    
    \subfloat[Prompt Ensembling. 
    ]{
    \includegraphics[height=0.19\linewidth]{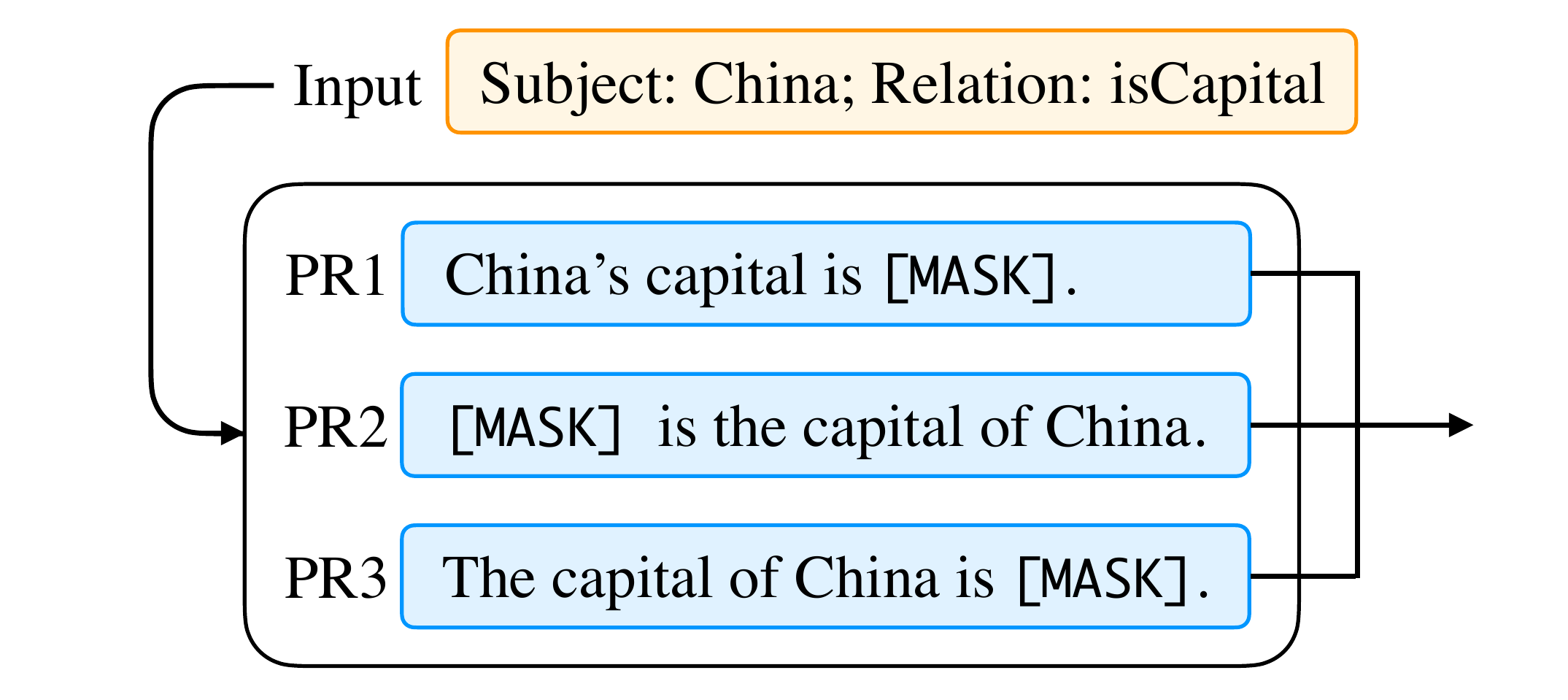}
    }      
    \subfloat[Prompt Augmentation.]{  
    \includegraphics[height=0.19\linewidth]{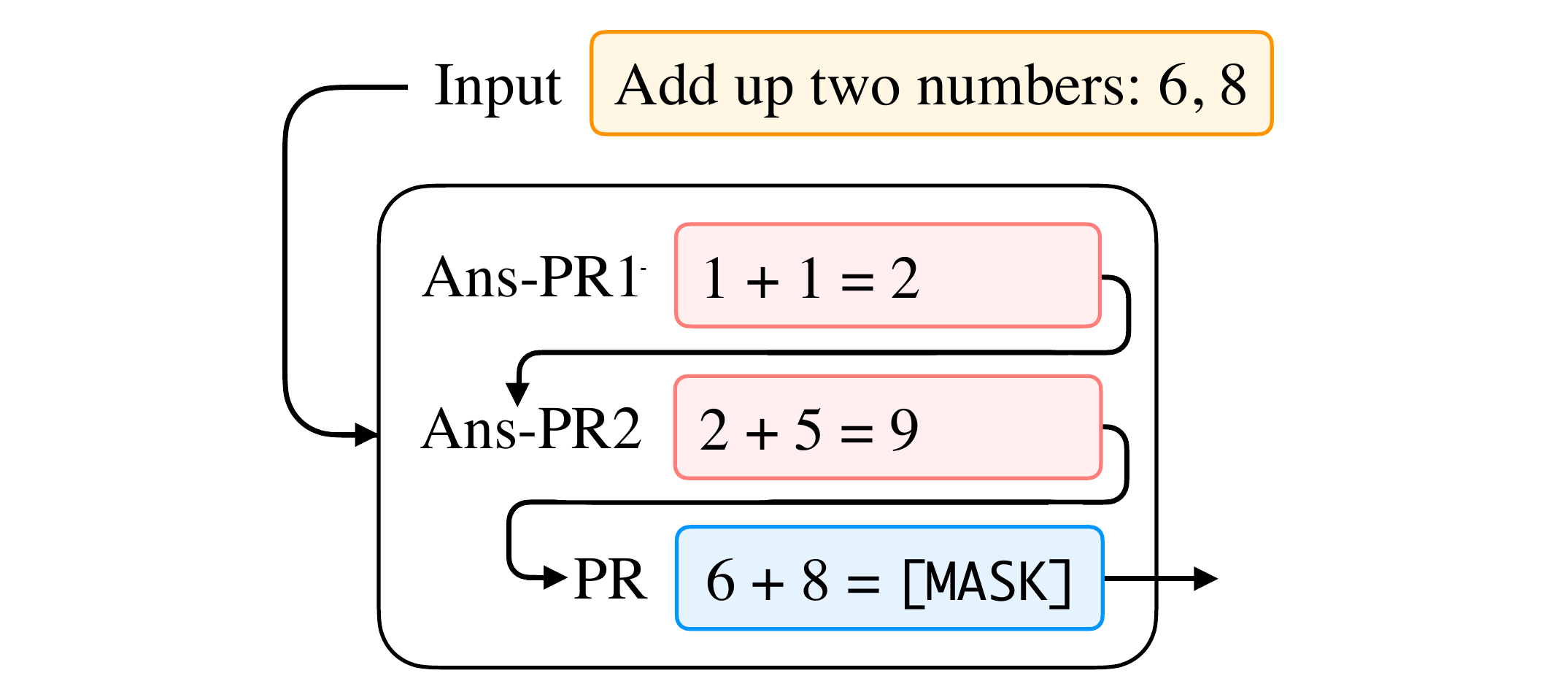}
    }      
    \\
    \subfloat[Prompt Composition.]{
    \includegraphics[height=0.23\linewidth]{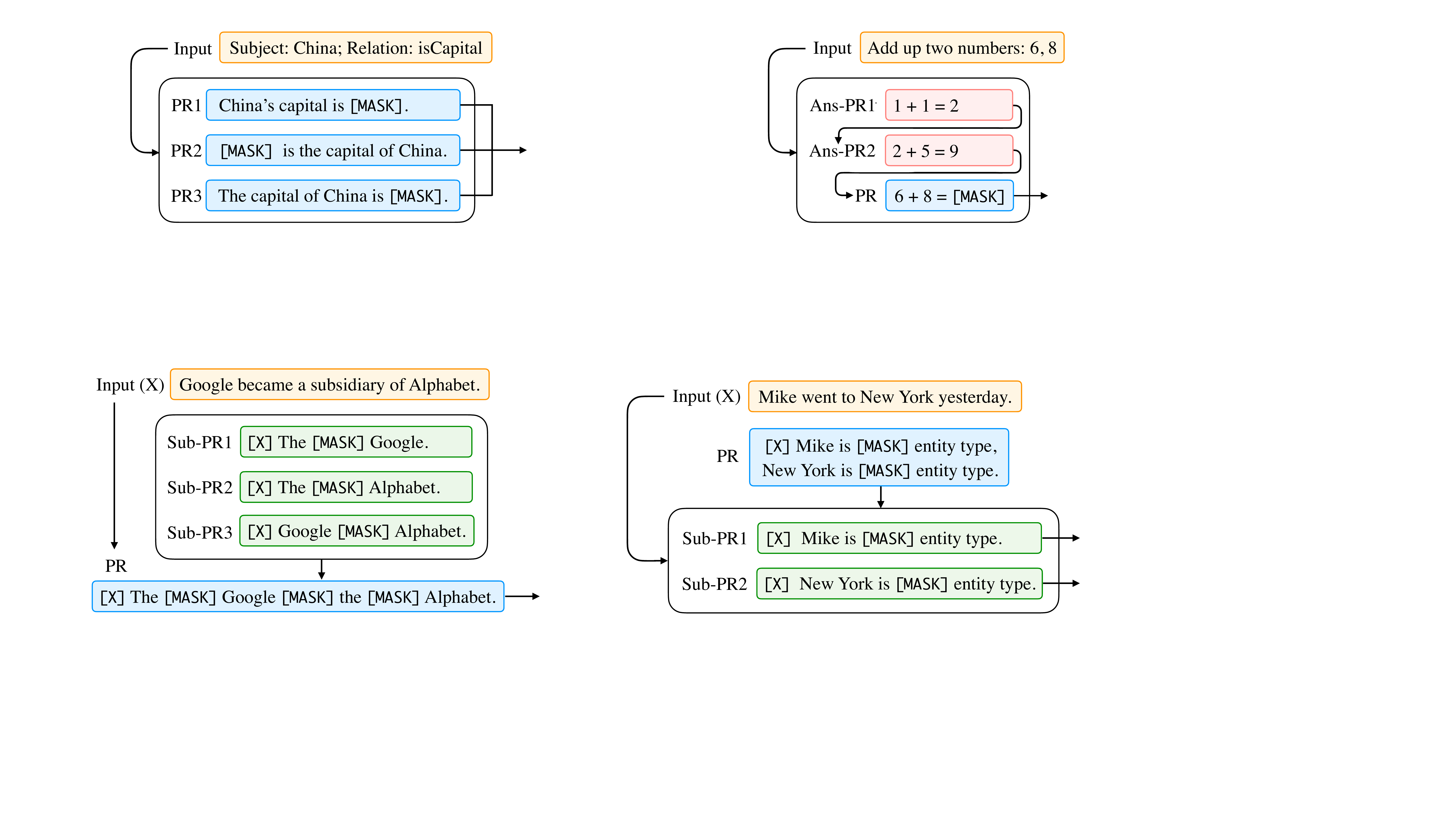}
    }     
    \subfloat[Prompt Decomposition.]{
    \includegraphics[height=0.23\linewidth]{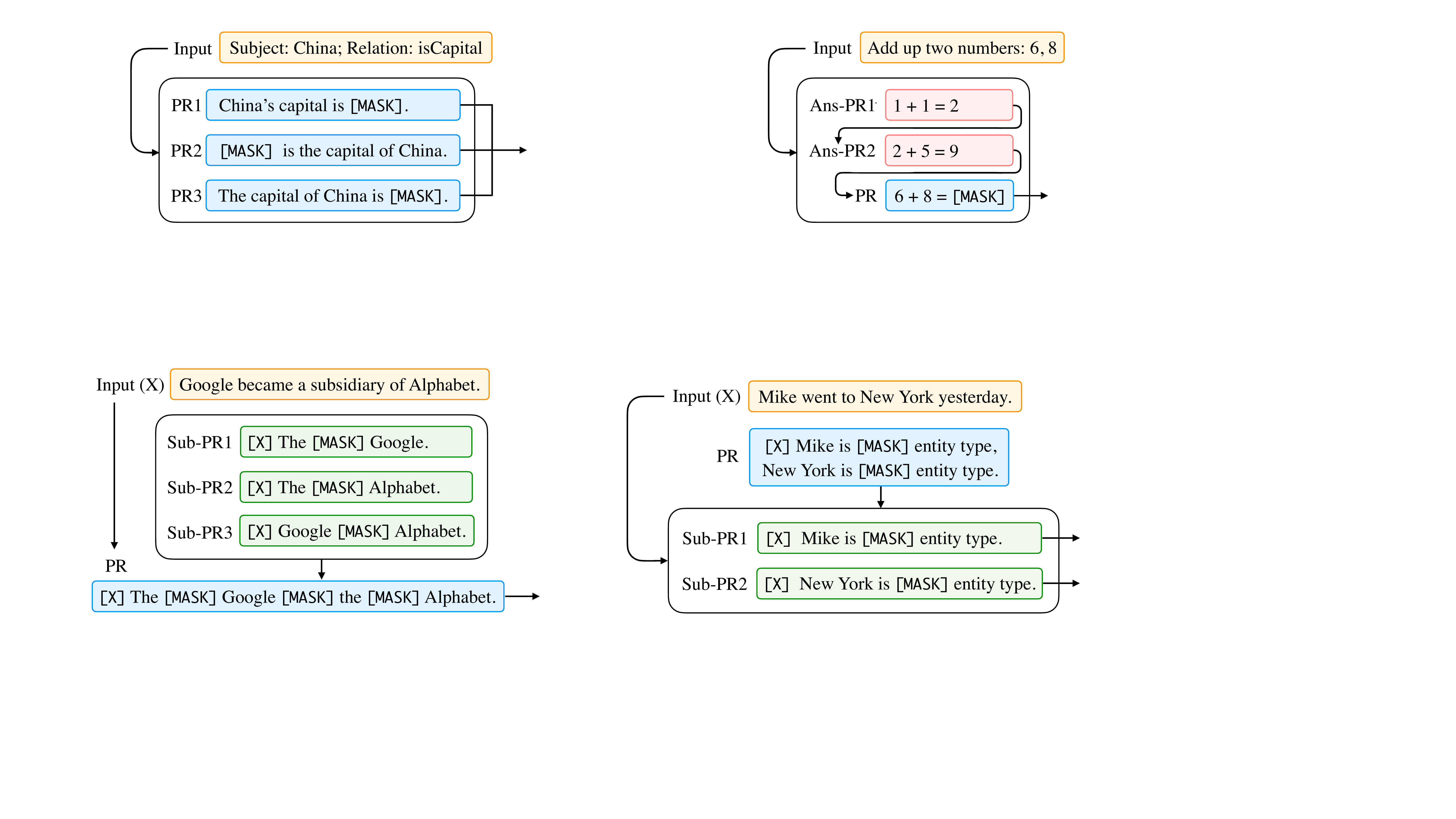}
    }    
    \caption{Different multi-prompt learning strategies. We use different colors to differentiate different components as follows. `` \includegraphics[scale=0.25]{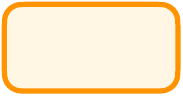} " for input text, 
    `` \includegraphics[scale=0.25]{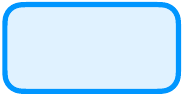} " for prompt, `` \includegraphics[scale=0.25]{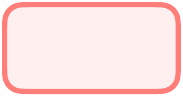} " for answered prompt. `` \includegraphics[scale=0.25]{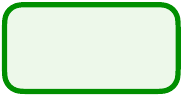} " for sub-prompt. We use the following abbreviations. ``PR" for prompt, ``Ans-PR" for answered prompt, ``Sub-PR" for sub-prompt.}
    \label{fig:multi-prompt-example}
\end{figure*}

The prompt engineering methods we discussed so far focused mainly on constructing a \emph{single} prompt for an input. 
However, a significant body of research has demonstrated that the use of multiple prompts can further improve the efficacy of prompting methods, and we will call these methods \term{multi-prompt learning} methods. In practice, there are several ways to extend the single prompt learning to the use multiple prompts, which have a variety of motivations. 
We summarize representative methods in the ``Multi-prompt Learning'' section of Fig.\ref{fig:typo-prompt} as well as Fig.\ref{fig:multi-prompt-example}.

\subsection{Prompt Ensembling}
\label{sec:prompt-ensembling}

\term{Prompt ensembling} is the process of using multiple \emph{unanswered} prompts for an input at inference time to make predictions.
An example is shown in Fig.~\ref{fig:multi-prompt-example}-(a). The multiple prompts can either be discrete prompts or continuous prompts.\footnote{Multiple continuous prompts are typically learned by using different initializations or different random seeds.} This sort of prompt ensembling can (1) leverage the complementary advantages of different prompts, (2) alleviate the cost of prompt engineering, since choosing one best-performing prompt is challenging,
(3) stabilize performance on downstream tasks.

Prompt ensembling is connected to ensembling methods that are used to combine together multiple systems, which have a long history in machine learning \cite{DBLP:conf/ijcai/TingG97,zhou2002ensembling,duh2011generalized}.
Current research also borrows ideas from these works to derive effective ways for prompt ensembling, as described below.


\paragraph{Uniform averaging}
The most intuitive way to combine the predictions when using multiple prompts is to take the average of probabilities from different prompts. 
Concretely, this indicates that $P(\bm{z}|\bm{x}) \coloneqq \frac{1}{K} \sum_{i}^K P(\bm{z}|f_{\text{prompt},i}(\bm{x}))$ where $f_{\text{prompt},i}(\cdot)$ is the $i$th prompt in the prompt ensemble.
\citet{jiang-etal-2020-know} first filter their prompts by selecting $K$ prompts that achieve the highest accuracy on the training set, and then use the average log probabilities obtained from the top $K$ prompts to calculate the probability for a single token at \texttt{[Z]} position when performing factual probing tasks.
\citet{schick2021exploiting} also try a simple average when using an ensemble model to annotate an unlabeled dataset. 
When performing text generation evaluation, \citet{yuan2021bartscore} formulates this task as a text generation problem and take the average of the final generation scores obtained using different prompts. 


\paragraph{Weighted averaging}
Simple uniform averaging of results from multiple prompts is easy to implement, but can also be suboptimal given that some prompts are more performant than others.
To account for this, some works also explore to use of weighted averages for prompt ensembling where each prompt is associated with a weight.
The weights are typically pre-specified based on prompt performance or optimized using a training set.
For example, \citet{jiang-etal-2020-know} learn the weight for each prompt by maximizing the probability of the target output over training data.
\citet{qin-eisner-2021-learning} use the same approach except that the weight for each prompt is optimized together with soft prompt parameters.
Besides, \citet{qin-eisner-2021-learning} also introduce a data-dependent weighting strategy where the probability of the input appearing in that prompt is considered in weighting different prompts as well. \citet{schick2021exploiting, schick2021its} set the weight for each prompt proportional to the accuracy on the training set before training.

\paragraph{Majority voting}
For classification tasks, majority voting can also be used to combine the results from different prompts \citep{lester2021power, Hambardzumyan2021WARPWA}.

\paragraph{Knowledge distillation}
An ensemble of deep learning models can typically improve the performance, and this superior performance can be distilled into a single model using knowledge distillation \citep{DBLP:journals/corr/abs-2012-09816}. To incorporate this idea, \citet{schick2021exploiting, schick2021its, schick2020fewshot} train a separate model for each manually-created template-answer pair, and use the ensemble of them to annotate an unlabeled dataset. Then the final model is trained to distill the knowledge from the annotated dataset. \citet{gao2021making} use a similar ensemble method on their automatically generated templates.

\paragraph{Prompt ensembling for text generation}
There is relatively little work on prompt ensembling for generation tasks (i.e. tasks where the answers is a string of tokens instead of a single one).
A simple way to perform ensembling in this case is to use standard methods that generate the output based on the ensembled probability of the next word in the answer sequence $P(z_t|\bm{x},z_{<t}) \coloneqq \frac{1}{K} \sum_{i}^K P(z_t|f_{\text{prompt},i}(\bm{x}),z_{<t})$.
In contrast, \citet{schick2020fewshot} train a separate model for each prompt $f_{\text{prompt},i}(\bm{x})$, and thus storing each of these fine-tuned LMs in memory is infeasible.
Instead, they first decode generations using each model and then score each generation by averaging their generation probability across all models.

\subsection{Prompt Augmentation}
\label{sec:prompt-augmentation}
\term{Prompt augmentation}, also sometimes called \term{demonstration learning}~\cite{gao2021making}, provides a few additional \emph{answered prompts} that can be used to demonstrate how the LM should provide the answer to the actual prompt instantiated with the input $\bm{x}$. 
For example, instead of just providing a prompt of ``China's capital is \texttt{[Z]} .'', the prompt can be prefaced by a few examples such as ``Great Britain's capital is London . Japan's capital is Tokyo . China's capital is \texttt{[Z]} .'' Another example of performing addition of two numbers can be found in Fig.~\ref{fig:multi-prompt-example}-(b).
These few-shot demonstrations take advantage of the ability of strong language models to learn repetitive patterns \citep{brown2020language}. 

Although the idea of prompt augmentation is simple, there are several aspects that make it challenging:
(1) \emph{Sample Selection:} how to choose the most effective examples? (2) \emph{Sample Ordering:} How to order the chosen examples with the prompt?

\paragraph{Sample Selection}
Researchers have found that the choice of examples used in this few-shot scenario can result in very different performance, ranging from near state-of-the-art accuracy on some tasks to near random guess \citep{Lu2021FantasticallyOP}. 
To address this issue, \citet{gao2021making, liu2021makes} utilize sentence embeddings to sample examples that are close to the input in this embedding space.
To measure the generalization capability of pre-trained LMs to perform new tasks based on instructions, \citet{DBLP:journals/corr/abs-2104-08773} provide both positive samples and negative samples that highlight things to avoid.

\paragraph{Sample Ordering}
\citet{Lu2021FantasticallyOP} found that the order of answered prompts provided to the model plays an important role in model performance, and propose entropy-based methods to score different candidate permutations. 
\citet{kumar2021reordering} search for a good permutation of training examples as augmented prompts and learn a separator token between the prompts for further gains in performance.

Prompt augmentation is closely related to retrieval-based methods that provide more textual context to the model to improve performance \citep{guu-etal-2018-generating}, a method which has also been shown to be effective in prompt-based learning \citep{Petroni2020HowCA}. However, the key difference lies in the fact that prompt augmentation also leverages the template and answer, while larger context learning does not.

\subsection{Prompt Composition}
\label{sec:6-prompt-composition}
For those composable tasks, which can be composed based on more fundamental subtasks,
we can also perform \term{prompt composition}, using multiple sub-prompts, each for one subtask, and then defining a composite prompt based on those sub-prompts. This process is illustrated in Fig.~\ref{fig:multi-prompt-example}-(c). For example, in the relation extraction task, which aims to extract the relation of two entities, we can break down the task into several subtasks including identifying the characteristics of entities and classifying the relationships between entities. Based on this intuition, \citet{han2021ptr} first use multiple manually created sub-prompts for entity recognition and relation classification and then compose them into a complete prompt based on logic rules for relation extraction.

\subsection{Prompt Decomposition}
For tasks where multiple predictions should be performed for one sample (e.g., sequence labeling), directly defining a holistic prompt with regards to the entire input text $\bm{x}$ is challenging. One intuitive method to address this problem is to break down the holistic prompt into different sub-prompts, and then answer each sub-prompt separately.
Fig.\ref{fig:multi-prompt-example}-(d) illustrates this idea with an example from the named entity recognition task, which aims to identify all named entities in an input sentence.
In this case, the input will first be converted into a set of text spans, and the model can then be prompted to predict the entity type (including ``Not an Entity'') for each span.
It is not easy to predict all the span types at the same time due to the large number of spans, so different prompts for each span can be created and predicted separately. This sort of \term{prompt decomposition} for named entity recognition has been explored by \citet{cui2021templatebased} where they apply the approach we discussed here.


\section{Training Strategies for Prompting Methods}
\label{sec:tuning}

With the methods in the above sections, it is now clear how to obtain an appropriate prompt (or prompts) and corresponding answers.
Now we discuss about methods that explicitly train models in concert with prompting methods, as outlined in the ``Training Strategies'' section of Fig.\ref{fig:typo-prompt}.

\subsection{Training Settings}

In many cases, prompting methods can be used without \emph{any} explicit training of the LM for the down-stream task, simply taking an LM that has been trained to predict the probability of text $P(\bm{x})$ and applying it as-is to fill the cloze or prefix prompts defined to specify the task.
This is traditionally called the \term{zero-shot} setting, as there is zero training data for the task of interest.

However, there are also methods that use training data to train the model in concert with prompting methods.
These consist of either \term{full-data learning}, where a reasonably large number of training examples are used to train the model, or \term{few-shot learning} where a very small number of examples are used to train the model.
Prompting methods are particularly useful in the latter case, as there are generally not enough training examples to fully specify the desired behavior, and thus using a prompt to push the model in the right direction is particularly effective.

One thing to note is that for many of the prompt engineering methods described in \S\ref{sec:4-prompt-template-engineering}, although annotated training samples are not explicitly used in the training of the downstream task model, they \emph{are} often used in the construction or validation of the prompts that the downstream task will use.
As noted by \citet{perez2021true}, this is arguably not true zero-shot learning with respect to the downstream task.

\subsection{Parameter Update Methods}
In prompt-based downstream task learning, there are usually two types of parameters, namely those from (1) pre-trained models and (2) prompts.
Which part of parameters should be updated is one important design decision, which can lead to different levels of applicability in different scenarios. We summarize five tuning strategies (as shown in Tab.~\ref{tab1:tuning-strategy}) based on (i) whether the parameters of the underlying LM are tuned, (ii) whether there are additional prompt-related parameters, (iii) if there are additional prompt-related parameters, whether those parameters are tuned.


\begin{table}[!th]
  \centering
  \footnotesize
    \begin{tabular}{llccl}
    \toprule
    \multicolumn{1}{l}{\multirow{2}[2]{*}{\textbf{Strategy}}} & \multirow{2}[2]{*}{\textbf{LM Params}} & \multicolumn{2}{c}{\textbf{Prompt Params}} &
    \multicolumn{1}{l}{\multirow{2}[2]{*}{\textbf{Example}}} \\
        \cmidrule(lr){3-4}
          &       & \multicolumn{1}{l}{\textbf{Additional}} & \multicolumn{1}{l}{\textbf{Tuned}} &  \\
    \midrule
        Promptless Fine-tuning & Tuned & \multicolumn{2}{c}{-}    & ELMo \citenumber{peters-etal-2018-deep}, BERT \citenumber{devlin-etal-2019-bert}, BART \citenumber{lewis-etal-2020-bart} \\
\midrule
    Tuning-free Prompting & Frozen & \XSolidBrush    & \XSolidBrush    & GPT-3 \citenumber{brown2020language}, AutoPrompt \citenumber{autoprompt:emnlp20}, LAMA \citenumber{petroni-etal-2019-language} \\
    \midrule
    Fixed-LM Prompt Tuning & Frozen & \Checkmark   & Tuned & 
    Prefix-Tuning \citenumber{li2021prefix}, 
    Prompt-Tuning \citenumber{lester2021power} \\
    \midrule
    \multirow{1}[0]{*}{Fixed-prompt LM Tuning} & Tuned & \XSolidBrush    & \XSolidBrush    & PET-TC \citenumber{schick2021exploiting}, PET-Gen \citenumber{schick2020fewshot}, LM-BFF \citenumber{gao2021making} \\
\midrule
    Prompt+LM Fine-tuning & Tuned & \Checkmark   & Tuned & PADA \citenumber{bendavid2021pada}, P-Tuning \citenumber{liu2021ptuning}, PTR~\citenumber{han2021ptr} \\
    \bottomrule
    \end{tabular}%
  \caption{Characteristics of different tuning strategies. ``Additional'' represents if there are additional parameters beyond LM parameters while ``Tuned'' denotes if parameters are updated.}
  \label{tab1:tuning-strategy}
\end{table}%

\subsubsection{Promptless Fine-tuning} \label{sec:model-fine-tuning}
As mentioned in the introduction, the \term{pre-train and fine-tune} strategy has been widely used in NLP since before the popularization of prompting methods.
Here we refer to pre-training and fine-tuning \emph{without} prompts as \term{promptless fine-tuning}, to contrast with the prompt-based learning methods introduced in the following sections.
In this strategy, given a dataset of a task, all (or some \cite{howard2018universal,peters-etal-2019-tune}) of the parameters of the pre-trained LM will be updated via gradients induced from downstream training samples. 
Typical examples of pre-trained models tuned in this way include BERT~\citenumber{devlin-etal-2019-bert} and RoBERTa~\citenumber{Liu2019RoBERTaAR}.
This is a simple, powerful, and widely-used method, but it may overfit or not learn stably on small datasets~\cite{dodge2020fine}.
Models are also prone to \term{catastrophic forgetting}, where the LM loses its ability to do things that it was able to do before fine-tuning \citep{mccloskey1989catastrophic}.

\begin{myboxnote}[]
\begin{itemize*}
    \item \textbf{Advantages}: Simplicity, no need for prompt design. Tuning all the LM parameters allows the model to fit to larger training datasets.
    \item \textbf{Disadvantages}: LMs may overfit or not learn stably on smaller datasets.
\end{itemize*}
\end{myboxnote}

\subsubsection{Tuning-free Prompting} \label{sec:tuning-free-prompting}

\term{Tuning-free prompting} directly generates the answers without changing the parameters of the pre-trained LMs based only on a prompt, as described in the simplest incarnation of prompting in \S\ref{sec:2-formal-description}.
These can be optionally augmenting input with answered prompts as described in \S\ref{sec:prompt-augmentation}, and this combination of tuning-free prompting and prompt augmentation is also referred to as \term{in-context learning} \citep{brown2020language}.
Typical examples of tuning-free prompting include LAMA \citenumber{petroni-etal-2019-language}  and GPT-3~\citenumber{brown2020language}.

\begin{myboxnote}[]
\begin{itemize*}
    \item \textbf{Advantages}: Efficiency, there is no parameter update process. No catastrophic forgetting, as LM parameters remain fixed. Applicable in zero-shot  settings.
    \item \textbf{Disadvantages}: Because prompts are the only method that provide the task specification, heavy engineering is necessary to achieve high accuracy. In particular in the in-context learning setting, providing many answered prompts can be slow at test time, and thus cannot easily use large training datasets.
\end{itemize*}
\end{myboxnote}

\subsubsection{Fixed-LM Prompt Tuning} \label{sec:prompt-only-tuning}
In the scenario where additional prompt-relevant parameters are introduced besides parameters of the pre-trained model, 
\term{fixed-LM prompt tuning} updates only the prompts' parameters using the supervision signal obtained from the downstream training samples, while keeping the entire pre-trained LM unchanged.
Typical examples are Prefix-Tuning~\citenumber{li2021prefix} and WARP~\citenumber{Hambardzumyan2021WARPWA}.

\begin{myboxnote}[]
\begin{itemize*}
    \item \textbf{Advantages}: Similarly to tuning-free prompting, it can retain knowledge in LMs and is suitable in few-shot scenarios.
    Often superior accuracy to tuning-free prompting.
    \item \textbf{Disadvantages}: Not applicable in zero-shot scenarios.
    While effective in few-shot scenarios, representation power is limited in large-data settings.
    Prompt engineering through choice of hyperparameters or seed prompts is necessary.
    Prompts are usually not human-interpretable or manipulable.
\end{itemize*}
\end{myboxnote}

\subsubsection{Fixed-prompt LM Tuning} \label{sec:prompt-fix-fine-tuning}

\term{Fixed-prompt LM tuning} tunes the parameters of the LM, as in the standard pre-train and fine-tune paradigm, but additionally uses prompts with fixed parameters to specify the model behavior.
This potentially leads to improvements, particularly in few-shot scenarios.

The most natural way to do so is to provide a discrete textual template that is applied to every training and test example.
Typical examples include PET-TC \citenumber{schick2021exploiting}, PET-Gen \citenumber{schick2020fewshot}, LM-BFF \citenumber{gao2021making}.
\citet{logan2021cutting} more recently observe that the prompt engineering can be reduced by allowing for a combination of answer engineering and partial LM fine-tuning.
For example, they define a very simple template, \textit{null prompt}, where the input and mask are directly concatenated ``\texttt{[X][Z]}'' without any template words, and find this achieves competitive accuracy.

\begin{myboxnote}[]
\begin{itemize*}
    \item \textbf{Advantages}: Prompt or answer engineering more completely specify the task, allowing for more efficient learning, particularly in few-shot scenarios.
    \item \textbf{Disadvantages}: Prompt or answer engineering are still required, although perhaps not as much as without prompting. LMs fine-tuned on one downstream task may not be effective on another one.
\end{itemize*}
\end{myboxnote}

\subsubsection{Prompt+LM Tuning} \label{sec:prompt-fine-tuning}
In this setting, there are prompt-relevant parameters, which can be fine-tuned together with the all or some of the parameters of the pre-trained models.
Representative examples include PADA \citenumber{bendavid2021pada}, P-Tuning \citenumber{liu2021ptuning}.
Notably, this setting is very similar to the standard pre-train and fine-tune paradigm, but the addition of the prompt can provide additional bootstrapping at the start of model training.

\begin{myboxnote}[]
\begin{itemize*}
    \item \textbf{Advantages}: This is the most expressive method, likely suitable for high-data settings.
    \item \textbf{Disadvantages}: Requires training and storing all parameters of the models.
    May overfit to small datasets.
\end{itemize*}
\end{myboxnote}

\section{Applications}
\label{sec:applications}
In previous sections, we examined prompting methods from the point of view of the mechanism of the method itself.
In this section, we rather organize prompting methods from the point of view of which applications they have been applied to.
We list these applications in Tab.~\ref{tab:papers-part1}-\ref{tab:papers-part2} and summarize them in the following sections.

\subsection{Knowledge Probing}
\paragraph{Factual Probing}
 \term{Factual probing} (a.k.a.~fact retrieval) is one of the earliest scenarios with respect to which prompting methods were applied. The motivation of exploring this task is to quantify how much factual knowledge the pre-trained LM's internal representations bear. In this task, parameters of pre-trained models are usually fixed, and knowledge is retrieved by transforming the original input into a cloze prompt as defined in \S\ref{sec:2-basics}, which can be manually crafted or automatically discovered. Relevant datasets including \texttt{LAMA} \citep{petroni-etal-2019-language} and \texttt{X-FACTR} \citep{jiang-etal-2020-x}. Since the answers are pre-defined, fact retrieval only focuses on finding effective templates and analyzing the results of different models using these templates. Both discrete template search \citep{petroni-etal-2019-language, Petroni2020HowCA, jiang-etal-2020-know, jiang-etal-2020-x, haviv-etal-2021-bertese,  autoprompt:emnlp20, perez2021true} and continuous template learning \citep{qin-eisner-2021-learning, liu2021ptuning, zhong2021optiprompt} have been explored within this context, as well as prompt ensemble learning \citep{jiang-etal-2020-know, qin-eisner-2021-learning}.
 
\paragraph{Linguistic Probing}
Besides factual knowledge, large-scale pre-training also allows LMs to handle linguistic phenomena such as analogies \citep{brown2020language}, negations \citep{DBLP:journals/tacl/Ettinger20}, semantic role sensitivity \citep{DBLP:journals/tacl/Ettinger20}, semantic similarity \citep{DBLP:journals/corr/abs-2107-02137}, cant understanding \citep{DBLP:journals/corr/abs-2107-02137}, and rare word understanding \citep{DBLP:conf/aaai/SchickS20}. The above knowledge can also be elicited by presenting \term{linguistic probing} tasks in the form of natural language sentences that are to be completed by the LM.

\subsection{Classification-based Tasks}
Prompt-based learning has been widely explored in classification-based tasks where prompt templates can be constructed relatively easily, such as text classification \citep{DBLP:conf/emnlp/YinHR19} and natural language inference \citep{schick2021exploiting}.
The key to prompting for classification-based tasks is reformulating it as an appropriate prompt.
For example, \citet{DBLP:conf/emnlp/YinHR19} use a prompt such as ``the topic of this document is \texttt{[Z]}.'', which is then fed into mask pre-trained LMs for slot filling.

\paragraph{Text Classification} 
For \term{text classification} tasks, most previous work has used cloze prompts, and both prompt engineering \cite{gao2021making, Hambardzumyan2021WARPWA, lester2021power} and answer engineering \cite{schick2021exploiting, DBLP:conf/coling/SchickSS20, gao2021making} have been explored extensively. 
Most existing works explore the efficacy of prompt learning for text classification in the context of \term{few-shot} setting with ``\term{fixed-prompt LM Tuning}'' strategies (defined in \S\ref{sec:prompt-fix-fine-tuning}).

\paragraph{Natural Language Inference (NLI)}
NLI aims to predict the relationship (e.g., \texttt{entailment}) of two given sentences. Similar to text classification tasks, for \term{natural language inference} tasks, cloze prompts are commonly used~\cite{schick2021exploiting}. Regarding prompt engineering, researchers mainly focus on the template search in the few-shot learning setting and the answer space $\mathcal{Z}$ is usually manually pre-selected from the vocabulary.

\begin{table*}[!ht]
\centering
\scriptsize
\renewcommand\arraystretch{1.1}
\setlength\tabcolsep{4.5pt}
\begin{tabular}{lllllcllclll}
\toprule
                                      &                                     &                                     &                                    & \multicolumn{3}{c}{\textbf{Prompt Engineering}}                                                 & \multicolumn{3}{c}{\textbf{Answer Engineering}}                                                    &                                   &                                         \\
                                      \cmidrule(lr){5-7} \cmidrule(lr){8-10}
\multirow{-2}{*}{\textbf{Work}}       & \multirow{-2}{*}{\textbf{Task}}     & \multirow{-2}{*}{\textbf{PLM}}      & \multirow{-2}{*}{\textbf{Setting}} & \textbf{Shape}             & \textbf{Man}                             & \textbf{Auto}          & \textbf{Shape}                 & \textbf{Man}                             & \textbf{Auto}       & \multirow{-2}{*}{\textbf{Tuning}} & \multirow{-2}{*}{\textbf{Mul-Pr}} \\
\midrule
LMComm \citenumber{DBLP:journals/corr/abs-1806-02847}                                & CR                                  & L2R                                & Zero                               & Clo                        & \Checkmark                   & -                      & Sp                             & \Checkmark                   & -                   & TFP                               & -                                       \\
\midrule
                                      & CR,QA                              &                                     &                                    &                            &                                             &                        &                                &                                             &                     &                                   &                                         \\
\multirow{-2}{*}{GPT-2 \citenumber{Radford2019LanguageMA}}               & \cellcolor[HTML]{FFFFFF}SUM,MT     & \multirow{-2}{*}{GPT-2}             & \multirow{-2}{*}{Zero,Few}        & \multirow{-2}{*}{Clo,Pre} & \multirow{-2}{*}{\Checkmark} & \multirow{-2}{*}{-}    & \multirow{-2}{*}{Tok,Sp,Sen} & \multirow{-2}{*}{\Checkmark} & \multirow{-2}{*}{-} & \multirow{-2}{*}{TFP}             & \multirow{-2}{*}{PA}                    \\
\midrule
WNLaMPro \citenumber{DBLP:conf/aaai/SchickS20}                              & LCP                                 & BERT                                & Zero                               & Clo                        & \Checkmark                   & -                      & Tok                            & \Checkmark                   & -                   & TFP                               & -                                       \\
\midrule

LMDiagnose \citenumber{DBLP:journals/tacl/Ettinger20}                           & CR,LCP                             & BERT                                & Zero                               & Clo                        & \Checkmark                   & -                      & Tok                            & \Checkmark                   & -                   & TFP                               & -                                       \\
\midrule
AdvTrigger \citenumber{DBLP:conf/emnlp/WallaceFKGS19}                            & GCG                                 & GPT-2                               & Full                               & Pre                        & -                                           & Disc                   & Sen                            & \Checkmark                   & -                   & TFP                               & -                                       \\
\midrule

CohRank \citenumber{DBLP:conf/emnlp/DavisonFR19}                              & CKM                                 & BERT                                & Zero                               & Clo                        & \Checkmark                   & -                      & Tok,Sp                        & \Checkmark                   & -                   & TFP                               & -                                       \\
\midrule
                                      &                                     & Conv,Trans                         &                                    &                            &                                             &                        &                                &                                             &                     &                                   &                                         \\
\multirow{-2}{*}{LAMA \citenumber{petroni-etal-2019-language}}                & \multirow{-2}{*}{FP}                & \cellcolor[HTML]{FFFFFF}ELMo,BERT  & \multirow{-2}{*}{Zero}             & \multirow{-2}{*}{Clo}      & \multirow{-2}{*}{\Checkmark} & \multirow{-2}{*}{-}    & \multirow{-2}{*}{Tok}          & \multirow{-2}{*}{\Checkmark} & \multirow{-2}{*}{-} & \multirow{-2}{*}{TFP}             & \multirow{-2}{*}{-}                     \\
\midrule

CTRL \citenumber{DBLP:journals/corr/abs-1909-05858}                                 & GCG                                 & CTRL                                & Full                               & Pre                        & \Checkmark                   & -                      & Sen                            & \Checkmark                   & -                   & LMT                               & -                                       \\
\midrule
                                      & TC,SUM                             &                                     &                                    &                            &                                             &                        &                                &                                             &                     &                                   &                                         \\
                                      
\multirow{-2}{*}{T5 \citenumber{JMLR:v21:20-074}}                  & \cellcolor[HTML]{FFFFFF}QA,MT      & \multirow{-2}{*}{T5}                & \multirow{-2}{*}{Full}             & \multirow{-2}{*}{Pre}      & \multirow{-2}{*}{\Checkmark} & \multirow{-2}{*}{-}    & \multirow{-2}{*}{Tok,Sp,Sen} & \multirow{-2}{*}{\Checkmark} & \multirow{-2}{*}{-} & \multirow{-2}{*}{LMT}             & \multirow{-2}{*}{-}                     \\
\midrule
                                      &                                     & Trans,ELMo                         &                                    &                            &                                             &                        &                                &                                             &                     &                                   &                                         \\
\multirow{-2}{*}{Neg \& Mis \citenumber{DBLP:conf/acl/KassnerS20}}      & \multirow{-2}{*}{FP}                & \cellcolor[HTML]{FFFFFF}BERT        & \multirow{-2}{*}{Zero}             & \multirow{-2}{*}{Clo}      & \multirow{-2}{*}{\Checkmark} & \multirow{-2}{*}{-}    & \multirow{-2}{*}{Tok}          & \multirow{-2}{*}{\Checkmark} & \multirow{-2}{*}{-} & \multirow{-2}{*}{TFP}             & \multirow{-2}{*}{-}                     \\
\midrule

LPAQA \citenumber{jiang-etal-2020-know}                                & FP                                  & BERT,ERNIE                         & Full                               & Clo                        & \Checkmark                   & Disc                   & Tok                            & \Checkmark                   & -                   & TFP                               & PE                                      \\
\midrule

ZSC \citenumber{DBLP:journals/corr/abs-1912-10165}                            & TC                                  & GPT-2                               & Full                               & Pre                        & \Checkmark                   & -                      & Tok,Sp                        & \Checkmark                   & -                   & LMT                               & -                                       \\
\midrule

PET-TC \citenumber{schick2021exploiting}                               & TC                                  & RoBERTa,XLM-R                      & Few                                & Pre                        & \Checkmark                   & -                      & Tok                            & \Checkmark                   & Disc                   & LMT                               & PE                                      \\
\midrule

ContxFP \citenumber{Petroni2020HowCA}                               & FP                                  & BERT,RoBERTa                       & Zero                               & Clo                        & \Checkmark                   & Disc                   & Tok                            & \Checkmark                   & -                   & TFP                               & -                                       \\
\midrule
UnifiedQA \citenumber{khashabi-etal-2020-unifiedqa} &	QA &	T5,BART &	Full &	Prefix &	\Checkmark &	- &	Tok,Sp,Sen &	\Checkmark &	- &	LMT	& - \\
\midrule
RAG \citenumber{DBLP:conf/nips/LewisPPPKGKLYR020} & QA,GCG,TC & BART & Full & Pre & - & Disc & Tok,Sp,Sen & \Checkmark & - & LMPT & PE \\

\midrule

                                      & QA,MT,GCG                         &                                     &                                    &                            &                                             &                        &                                &                                             &                     &                                   &                                         \\
                                      & \cellcolor[HTML]{FFFFFF}CR,TC,LCP &                                     &                                    &                            &                                             &                        &                                &                                             &                     &                                   &                                         \\
\multirow{-3}{*}{GPT-3 \citenumber{brown2020language}}               & MR,SR,AR                          & \multirow{-3}{*}{GPT-3}             & \multirow{-3}{*}{Zero,Few}        & \multirow{-3}{*}{Clo,Pre} & \multirow{-3}{*}{\Checkmark} & \multirow{-3}{*}{-}    & \multirow{-3}{*}{Tok,Sp,Sen} & \multirow{-3}{*}{\Checkmark} & \multirow{-3}{*}{-} & \multirow{-3}{*}{TFP}             & \multirow{-3}{*}{PA}                    \\
\midrule

CommS2S \citenumber{yang-etal-2020-designing}                               & CR                                  & T5                                  & Full                               & Pre                        & \Checkmark                   & -                      & Tok                            & \Checkmark                   & -                   & LMT                               & -                                       \\
\midrule

PET-SGLUE \citenumber{schick2021its}                        & TC                                  & ALBERT                              & Few                                & Clo                        & \Checkmark                   & -                      & Tok,Sp                        & \Checkmark                   & -                   & LMT                               & PE                                      \\
\midrule
                                      &                                     & GPT-1,GPT-2                        &                                    &                            &                                             &                        & \multicolumn{3}{l}{}                                                                               &                                   &                                         \\
\multirow{-2}{*}{ToxicityPrompts \citenumber{gehman2020realtoxicityprompts}} & \multirow{-2}{*}{GCG}               & \cellcolor[HTML]{FFFFFF}GPT-3,CTRL & \multirow{-2}{*}{Zero}             & \multirow{-2}{*}{Pre}      & \multirow{-2}{*}{\Checkmark} & \multirow{-2}{*}{-}    & \multicolumn{3}{l}{\multirow{-2}{*}{N/A}}                                                          & \multirow{-2}{*}{TFP}             & \multirow{-2}{*}{-}                     \\
\midrule

WhyLM \citenumber{DBLP:conf/iclr/SaunshiMA21}                                & Theory                              & GPT-2                               & Full                               & Pre                        & \Checkmark                   & -                      & Tok                            & \Checkmark                   & -                   & PT                               & -                                       \\
\midrule
                                      &                                     & mBERT,BERT                         &                                    &                            &                                             &                        &                                &                                             &                     &                                   &                                         \\
\multirow{-2}{*}{X-FACTR \citenumber{jiang-etal-2020-x}}             & \multirow{-2}{*}{FP}                & \cellcolor[HTML]{FFFFFF}XLM,XLM-R  & \multirow{-2}{*}{Zero}             & \multirow{-2}{*}{Clo}      & \multirow{-2}{*}{\Checkmark} & \multirow{-2}{*}{-}    & \multirow{-2}{*}{Tok,Sp}      & \multirow{-2}{*}{\Checkmark} & \multirow{-2}{*}{-} & \multirow{-2}{*}{TFP}             & \multirow{-2}{*}{-}                     \\
\midrule

Petal \citenumber{DBLP:conf/coling/SchickSS20}                                & TC                                  & RoBERTa                             & Few                                & Clo                        & \Checkmark                   & -                      & Tok                            & -                                           & Disc                & LMT                               & PE                                      \\
\midrule

AutoPrompt \citenumber{autoprompt:emnlp20}                           & TC,FP,IE                          & BERT,RoBERTa                       & Full                               & Clo                        & -                                           & Disc                   & Tok                            & -                                           & Disc                & TFP                               & -                                       \\
\midrule

CTRLsum \citenumber{DBLP:journals/corr/abs-2012-04281}                              & SUM                                 & BART                                & Full                               & Pre                        & \Checkmark                   & -                      & Sen                            & \Checkmark                   & -                   & LMT                               & -                                       \\
\midrule

PET-Gen \citenumber{schick2020fewshot}                              & SUM                                 & PEGASUS                             & Few                                & Pre                        & \Checkmark                   & -                      & Sen                            & \Checkmark                   & -                   & LMT                               & PE                                      \\
\midrule

LM-BFF \citenumber{gao2021making}                               & TC                                  & RoBERTa                             & Few                                & Clo                        & -                                           & Disc                   & Tok                            & -                                           & Disc                & LMT                               & PE,PA                                  \\
\midrule

WARP \citenumber{Hambardzumyan2021WARPWA}                              & TC                                  & RoBERTa                             & Few,Full                               & Clo,Pre                   & \Checkmark                   & Cont                   & Tok                            & \Checkmark                   & Cont                & PT                               & PE                                      \\
\midrule

Prefix-Tuning \citenumber{li2021prefix}                        & D2T,SUM                            & GPT-2,BART                         & Full                               & Pre                        & -                                           & Cont                   & Sen                            & \Checkmark                   & -                   & PT                               & -                                       \\
\midrule

KATE \citenumber{liu2021makes}                                 & TC,D2T,QA                         & GPT-3                               & Few                                & Pre                        & \Checkmark                   & -                      & Tok,Sp,Sen                   & \Checkmark                   & -                   & TFP                               & PA                                      \\
\midrule
                                      & MT,MR                              &                                     &                                    &                            &                                             &                        &                                &                                             &                     &                                   &                                         \\
\multirow{-2}{*}{PromptProg \citenumber{10.1145/3411763.3451760}}          & \cellcolor[HTML]{FFFFFF}AR,QA      & \multirow{-2}{*}{GPT-3}             & \multirow{-2}{*}{Zero,Few}        & \multirow{-2}{*}{Pre}      & \multirow{-2}{*}{\Checkmark} & \multirow{-2}{*}{-}    & \multirow{-2}{*}{Tok,Sp,Sen} & \multirow{-2}{*}{\Checkmark} & \multirow{-2}{*}{-} & \multirow{-2}{*}{TFP}             & \multirow{-2}{*}{PA}                    \\
\midrule

ContxCalibrate \citenumber{zhao2021calibrate}                       & TC,FP,IE                          & GPT-2,GPT-3                        & Few                                & Pre                        & \Checkmark                   & -                      & Tok,Sp                        & \Checkmark                   & -                   & TFP                               & PA                                      \\
\midrule
PADA \citenumber{bendavid2021pada}                                  & TC,TAG                             & T5                                  & Full                               & Pre                        & -                                           & Disc                   & \multicolumn{3}{l}{N/A}                                                                            & LMPT                                & -                                       \\
\midrule
SD \citenumber{schick2021selfdiagnosis}                                   & GCG                                 & GPT-2                               & Zero                               & Pre                        & \Checkmark                   & -                      & \multicolumn{3}{l}{N/A}                                                                            & TFP                               & -                                       \\
\midrule
BERTese \citenumber{haviv-etal-2021-bertese}                              & FP                                  & BERT                                & Full                               & Clo                        & \Checkmark                   & Disc                   & Tok                            & \Checkmark                   & -                   & TFP                               & -                                       \\
\midrule
Prompt2Data \citenumber{scao2021many} & TC                              & RoBERTa                             & Full                               & Clo                        & \Checkmark                   & -                      & Tok,Sp                        & \Checkmark                   & -                   & LMT                               & -                                       \\
\midrule

                                      &                                     & GPT-2,BERT                         &                                    &                            &                                             &                        &                                &                                             &                     &                                   &                                         \\
\multirow{-2}{*}{P-Tuning \citenumber{liu2021ptuning}}            & \multirow{-2}{*}{FP,TC}            & \cellcolor[HTML]{FFFFFF}ALBERT      & \multirow{-2}{*}{Few,Full}        & \multirow{-2}{*}{Clo,Pre} & \multirow{-2}{*}{\Checkmark} & \multirow{-2}{*}{Cont} & \multirow{-2}{*}{Tok,Sp}      & \multirow{-2}{*}{\Checkmark} & \multirow{-2}{*}{-} & \multirow{-2}{*}{TFP,LMPT}         & \multirow{-2}{*}{-}                     \\
\midrule
GLM \citenumber{du2021nlp}                                  & TC                                  & GLM                                 & Full                               & Clo                        & \Checkmark                   & -                      & Tok,Sp                        & \Checkmark                   & -                   & LMT                               & -                                       \\

\bottomrule
\end{tabular}
\caption{\label{tab:papers-part1}An organization of works on prompting (Part 1). See the caption of Tab.~\ref{tab:papers-part2} for a detailed description for all the abbreviations used in this table.}
\end{table*}

\begin{table*}[!htbp]
\centering
\scriptsize
\renewcommand\arraystretch{1.1}
\setlength\tabcolsep{4.1pt}
\begin{tabular}{lllllcllclll}
\toprule
                                      &                                     &                                     &                                    & \multicolumn{3}{c}{\textbf{Prompt Engineering}}                                                 & \multicolumn{3}{c}{\textbf{Answer Engineering}}                                                    &                                   &                                         \\
                                                                            \cmidrule(lr){5-7} \cmidrule(lr){8-10}
\multirow{-2}{*}{\textbf{Work}}       & \multirow{-2}{*}{\textbf{Task}}     & \multirow{-2}{*}{\textbf{PLM}}      & \multirow{-2}{*}{\textbf{Setting}} & \textbf{Shape}             & \textbf{Man}                             & \textbf{Auto}          & \textbf{Shape}                 & \textbf{Man}                             & \textbf{Auto}       & \multirow{-2}{*}{\textbf{Tuning}} & \multirow{-2}{*}{\textbf{Mul-Pr}} \\
\midrule

ADAPET \citenumber{tam2021improving}                               & TC                                  & ALBERT                              & Few                                & Clo                        & \Checkmark                   & -                      & Tok,Sp                        & \Checkmark                   & -                   & LMT                               & -                                       \\
\midrule
Meta \citenumber{zhong2021meta}                                 & TC                                  & T5                                  & Full                               & Pre                        & \Checkmark                   & -                      & Tok                            & \Checkmark                   & -                   & LMT                               & -                                       \\
\midrule

OptiPrompt \citenumber{zhong2021optiprompt}                            & FP                                  & BERT                                & Full                               & Clo                        & \Checkmark                   & Cont                   & Tok                            & \Checkmark                   & -                   & PT                               & -                                       \\
\midrule
                                      &                                     & BERT,BART                          &                                    &                            &                                             &                        &                                &                                             &                     &                                   &                                         \\
\multirow{-2}{*}{Soft \citenumber{qin-eisner-2021-learning}}                & \multirow{-2}{*}{FP}                & RoBERTa                             & \multirow{-2}{*}{Full}             & \multirow{-2}{*}{Clo}      & \multirow{-2}{*}{\Checkmark} & \multirow{-2}{*}{Cont} & \multirow{-2}{*}{Tok}          & \multirow{-2}{*}{\Checkmark} & \multirow{-2}{*}{-} & \multirow{-2}{*}{PT}             & \multirow{-2}{*}{PE}                    \\
\midrule

DINO \citenumber{schick2021generating}                              & GCG                                 & GPT-2                               & Zero                               & Pre                        & \Checkmark                   & -                      & \multicolumn{3}{l}{N/A}                                                                            & TFP                               & -                                       \\
\midrule

AdaPrompt \citenumber{DBLP:journals/corr/abs-2104-07650}                            & IE                                  & BERT                                & Few,Full                          & Clo                        & \Checkmark                   & -                      & Tok                            & -                   & Disc                   & LMT                               & -                                       \\
\midrule

PMI$_\text{DC}$ \citenumber{holtzman2021surface}                           & GCG,QA,TC                         & GPT-2,GPT-3                        & Zero                               & Pre                        & \Checkmark                   & -                      & Tok,Sp,Sen                   & \Checkmark                   & -                   & TFP                               & -                                       \\
\midrule

Prompt-Tuning \citenumber{lester2021power}                          & TC                                  & T5                                  & Full                               & Pre                        & -                                           & Cont                   & Tok,Sp                        & \Checkmark                   & -                   & PT                               & PE                                      \\
\midrule

Natural-Instr \citenumber{DBLP:journals/corr/abs-2104-08773}                  & GCG                                 & GPT-3,BART                         & Few,Full                          & Pre                        & \Checkmark                   & -                      & Tok,Sp,Sen                   & \Checkmark                   & -                   & TFP,LMT                          & PA                                      \\
\midrule

OrderEntropy \citenumber{Lu2021FantasticallyOP}                         & TC                                  & GPT-2,GPT-3                        & Few                                & Pre                        & \Checkmark                   & -                      & Tok                            & \Checkmark                   & -                   & TFP                               & PA                                      \\
\midrule

FewshotSemp \citenumber{Shin2021ConstrainedLM}                           & SEMP                                & GPT-3                               & Few                                & Pre                        & \Checkmark                   & -                      & Sen                            & \Checkmark                   & -                   & TFP                               & PA                                      \\
\midrule
& QA,CR,TC & & & & & & & & & & \\
\multirow{-2}{*}{PanGu-$\alpha$ \citenumber{zeng2021pangualpha}} & SUM,GCG &	\multirow{-2}{*}{PanGu-$\alpha$} & \multirow{-2}{*}{Zero,Few}	 &	\multirow{-2}{*}{Clo,Pre} &	\multirow{-2}{*}{\Checkmark}	& \multirow{-2}{*}{-}	 & \multirow{-2}{*}{Tok,Sp,Sen} &\multirow{-2}{*}{\Checkmark} & \multirow{-2}{*}{-} &	\multirow{-2}{*}{TFP} &	\multirow{-2}{*}{PA} \\
\midrule
                                      &                                     & GPT-2,GPT-3                        &                                    &                            &                                             &                        &                                &                                             &                     &                                   &                                         \\
\multirow{-2}{*}{TrueFewshot \citenumber{perez2021true}}         & \multirow{-2}{*}{TC,FP}            & \cellcolor[HTML]{FFFFFF}ALBERT      & \multirow{-2}{*}{Few}              & \multirow{-2}{*}{Clo,Pre} & \multirow{-2}{*}{\Checkmark} & \multirow{-2}{*}{Disc} & \multirow{-2}{*}{Tok,Sp}      & \multirow{-2}{*}{\Checkmark} & \multirow{-2}{*}{-} & \multirow{-2}{*}{TFP,LMT}        & \multirow{-2}{*}{-}                     \\
\midrule

PTR \citenumber{han2021ptr}                                   & IE                                  & RoBERTa                             & Full                               & Clo                        & \Checkmark                   & Cont                   & Tok,Sp                        & \Checkmark                   & -                   & LMPT                                & PC                                      \\
\midrule

TemplateNER \citenumber{cui2021templatebased}                           & TAG                                 & BART                                & Few,Full                          & Clo,Pre                   & \Checkmark                   & -                      & Tok                            & \Checkmark                   & -                   & LMT                               & PD                                      \\
\midrule

PERO \citenumber{kumar2021reordering}                                  & TC,FP                              & BERT,RoBERTa                      & Few                                & Pre                        & \Checkmark                   & -                      & Tok                            & \Checkmark                   & -                   & TFP                               & PA                                      \\
\midrule

PromptAnalysis \citenumber{wei2021pretrained}                        & Theory                                  & BERT                                & Full                               & Clo                        & -                                           & Cont                   & \multicolumn{3}{l}{N/A}                                                                            & PT                               & -                                       \\
\midrule

& QA,MR,SUM & & & & & & & & & & \\
\multirow{-2}{*}{CPM-2 \citenumber{DBLP:journals/corr/abs-2106-10715}} &
TC,GCG,MT & 
\multirow{-2}{*}{CPM-2} &
\multirow{-2}{*}{Full} & 
\multirow{-2}{*}{Pre} &
\multirow{-2}{*}{-} &
\multirow{-2}{*}{Cont} & 
\multirow{-2}{*}{Tok,Sp,Sent} &
\multirow{-2}{*}{\Checkmark} &
\multirow{-2}{*}{-} & 
\multirow{-2}{*}{PT,LMPT} & 
\multirow{-2}{*}{-} \\
\midrule

BARTScore \citenumber{yuan2021bartscore}                             & EVALG                               & BART                                & Zero                               & Pre                        & \Checkmark                   & Disc                   & Sen                            & \Checkmark                   & -                   & TFP                               & PE                                      \\
\midrule

NullPrompt \citenumber{logan2021cutting}                           & TC                                  & RoBERTa,ALBERT                     & Few                                & Pre                        & \Checkmark                   & -                      & Tok                            & \Checkmark                   & -                   & LMPT                                & -                                       \\
\midrule

Frozen \citenumber{DBLP:journals/corr/abs-2106-13884}                               & VQA,VFP,MG                        & GPT-like                            & Full                               & Pre                        & -                                           & Cont                   & Sp (Visual)                    & \Checkmark                   & -                   & PT                              & PA       \\
\midrule
& TC,LCP,NLI & & & & & & & & & & \\
& CR,QA,SUM & & & & & & & & & & \\
\multirow{-3}{*}{ERNIE-B3 \citenumber{DBLP:journals/corr/abs-2107-02137}} & GCG & \multirow{-3}{*}{ERNIE-B3} &	\multirow{-3}{*}{Zero} &	\multirow{-3}{*}{Clo,Pre} &	\multirow{-3}{*}{\Checkmark} &	\multirow{-3}{*}{-} &	\multirow{-3}{*}{Tok,Sp,Sen} &	\multirow{-3}{*}{\Checkmark} &	\multirow{-3}{*}{-} &	\multirow{-3}{*}{TFP} & \multirow{-3}{*}{-}\\
\midrule
& & & Zero,Few & & & & & & & & \\
\multirow{-2}{*}{Codex \citenumber{chen2021evaluating}} &	\multirow{-2}{*}{CodeGen} &	\multirow{-2}{*}{GPT} &	Full &	\multirow{-2}{*}{Pre} &	\multirow{-2}{*}{\Checkmark }&	\multirow{-2}{*}{-} &	\multirow{-2}{*}{Span} &	\multirow{-2}{*}{\Checkmark} &	\multirow{-2}{*}{Disc} &	\multirow{-2}{*}{TFP,LMT}	 & \multirow{-2}{*}{PA} \\
\midrule

& & & Zero,Few & & & & & & & & \\
\multirow{-2}{*}{HTLM \citenumber{aghajanyan2021htlm}} & 
\multirow{-2}{*}{TC,SUM} &
\multirow{-2}{*}{BART} & 
Full &
\multirow{-2}{*}{Clo} &
\multirow{-2}{*}{\Checkmark} &
\multirow{-2}{*}{Disc} &
\multirow{-2}{*}{Tok,Sp,Sen} &
\multirow{-2}{*}{\Checkmark} &
\multirow{-2}{*}{-} &
\multirow{-2}{*}{LMT} &
\multirow{-2}{*}{PA} \\
\midrule
FLEX \citenumber{DBLP:journals/corr/abs-2107-07170} & TC & T5 & Zero,Few & Pre & \Checkmark & - & Tok,Sp & \Checkmark & - & LMT & - \\

\bottomrule
\end{tabular}
\caption{\label{tab:papers-part2}An organization of works on prompting (Part 2). The \textbf{Task} column lists the tasks that are performed in corresponding papers. We use the following abbreviations. \textbf{CR}: Commonsense Reasoning. \textbf{QA}: Question Answering. \textbf{SUM}: Summarization. \textbf{MT}: Machine Translation. \textbf{LCP}: Linguistic Capacity Probing. \textbf{GCG}: General Conditional Generation. \textbf{CKM}: Commonsense Knowledge Mining. \textbf{FP}: Fact Probing. \textbf{TC}: Text Classification. \textbf{MR}: Mathematical Reasoning. \textbf{SR}: Symbolic Reasoning. \textbf{AR}: Analogical Reasoning. \textbf{Theory}: Theoretical Analysis. \textbf{IE}: Information Extraction. \textbf{D2T}: Data-to-text. \textbf{TAG}: Sequence Tagging. \textbf{SEMP}: Semantic Parsing. \textbf{EVALG}: Evaluation of Text Generation. \textbf{VQA}: Visual Question Answering. \textbf{VFP}: Visual Fact Probing. \textbf{MG}: Multimodal Grounding. \textbf{CodeGen}: Code generation. The \textbf{PLM} column lists all the pre-trained LMs that have been used in corresponding papers for downstream tasks. \textbf{GPT-like} is an autoregressive language model which makes small modifications to the original GPT-2 architecture. For other pre-trained LMs, please refer to \S\ref{sec:lm} for more information. \textbf{Setting} column lists the settings for prompt-based learning, can be zero-shot learning (\textbf{Zero}), few-shot learning (\textbf{Few}), fully supervised learning (\textbf{Full}). Under \textbf{Prompt Engineering}, \textbf{Shape} denotes the shape of the template (\textbf{Clo} for cloze and \textbf{Pre} for prefix), \textbf{Man} denotes whether human effort is needed, \textbf{Auto} denotes data-driven search methods (\textbf{Disc} for discrete search, \textbf{Cont} for continuous search). Under \textbf{Answer Engineering}, \textbf{Shape} indicates the shape of the answer (\textbf{Tok} for token-level, \textbf{Sp} for span-level, \textbf{Sen} for sentence- or document-level), and \textbf{Man} and \textbf{Auto} are the same as above. The \textbf{Tuning} column lists tuning strategies (\S\ref{sec:tuning}). \textbf{TFP}: Tuning-free Prompting. \textbf{LMT}: Fixed-prompt LM Tuning. \textbf{PT}: Fixed-LM Prompt Tuning. \textbf{LMPT}: LM+Prompt Tuning. The \textbf{Mul-Pr} column lists multi-prompt learning methods. \textbf{PA}: Prompt Augmentation.  \textbf{PE}: Prompt Ensembling. \textbf{PC}: Prompt Composition. \textbf{PD}: Prompt Decomposition.} 
\end{table*}

\clearpage

\subsection{Information Extraction}
Unlike classification tasks where cloze questions can often be intuitively constructed, for \term{information extraction} tasks constructing prompts often requires more finesse.

\paragraph{Relation Extraction}
\term{Relation extraction} is a task of predicting the relation between two entities in a sentence.
\citet{DBLP:journals/corr/abs-2104-07650} first explored the application of \term{fixed-prompt LM Tuning} in relation extraction and discuss two major challenges that hinder the direct inheritance of prompting methodology from classification tasks: (1) The larger label space (e.g.~80 in relation extraction v.s 2 in binary sentiment classification) results in more difficulty in answer engineering. (2) In relation extraction, different tokens in the input sentence may be more or less important (e.g.~entity mentions are more likely to participate in a relation), which, however, can not be easily reflected in the prompt templates for classification since the original prompt template regards each word equally. To address the above problems, \citet{DBLP:journals/corr/abs-2104-07650} propose an adaptive answer selection method to address the issue (1) and task-oriented prompt template construction for the issue (2), where they use special markers (e.g.~\texttt{[E]}) to
highlight the entity mentions in the template. Similarly, \citet{han2021ptr} incorporate entity type information via multiple prompt composition techniques (illustrated in Fig.~\ref{fig:multi-prompt-example}). 

\paragraph{Semantic Parsing} 
\term{Semantic parsing} is a task of generating a structured meaning representation given a natural language input.
\citet{Shin2021ConstrainedLM} explore the task of few-shot semantic parsing using LMs by (1) framing the semantic parsing task as a paraphrasing task \citep{berant-liang-2014-semantic} and (2) constraining the decoding process by only allowing output valid according to a grammar. They experiment with the \term{in-context learning} setting described in \S\ref{sec:tuning-free-prompting}, choosing answered prompts that are semantically close to a given test example (determined by the conditional generation probability of generating a test sample given another training example). The results demonstrate the effectiveness of the paraphrasing reformulation for semantic parsing tasks using pre-trained LMs.

\paragraph{Named Entity Recognition}
\term{Named entity recognition} (NER) is a task of identifying named entities (e.g., person name, location) in a given sentence.
The difficulty of prompt-based learning's application to tagging tasks, exemplified as NER, is that, unlike classification, (1) each unit to be predicted is a token or span instead of the whole input text, (2) there is a latent relationship between the token labels in the sample context.
Overall, the application of prompt-based learning in tagging task has not been fully explored.
\citet{cui2021templatebased} recently propose a template-based NER model using BART, which enumerates text spans and considers the generation probability of each type within manually crafted templates. For example, given an input ``Mike went to New York yesterday", to determine what type of entity ``Mike" is, they use the template ``Mike is a \texttt{[Z]} entity", and the answer space $\mathcal{Z}$ consists of values such as ``person'' or ``organization''.

\subsection{``Reasoning'' in NLP}
There is still a debate\footnote{e.g.~\url{https://medium.com/reconstruct-inc/the-golden-age-of-computer-vision-338da3e471d1}} about if deep neural networks are capable of performing ``reasoning'' or just memorizing patterns based on large training data~\cite{arpit2017closer,niven-kao-2019-probing}.
As such, there have been a number of attempts to probe models' reasoning ability by defining benchmark tasks that span different scenarios.
We detail below how prompting methods have been used in these tasks.

\paragraph{Commonsense Reasoning}
There are a number of benchmark datasets testing commonsense reasoning in NLP systems~\cite{huang-etal-2019-cosmos,rajani-etal-2019-explain,lin-etal-2020-commongen,ponti-etal-2020-xcopa}.
Some commonly attempted tasks involve solving Winograd Schemas \cite{levesque2012winograd}, which require the model to identify the antecedent of an ambiguous pronoun within context, or involve completing a sentence given multiple choices. For the former, an example could be ``The trophy doesn’t fit into the brown suitcase because it is too large." And the task for the model is to infer whether ``it" refers to the trophy or the ``suitcase". By replacing ``it" with its potential candidates in the original sentences and calculating the probability of the different choices, pre-trained LMs can perform quite well by choosing the choice that achieves the highest probability \citep{DBLP:journals/corr/abs-1806-02847}. For the latter, an example could be ``Eleanor offered to fix her visitor some coffee. Then she realized she didn’t have a clean \texttt{[Z]}.". The candidate choices are ``cup", ``bowl" and ``spoon". The task for the pre-trained LM is to choose the one from the three candidates that most conforms to common sense. For these kinds of tasks, we can also score the generation probability of each candidate and choose the one with the highest probability \citep{DBLP:journals/tacl/Ettinger20}.

\paragraph{Mathematical Reasoning}
Mathematical reasoning is the ability to solve mathematical problems, e.g. arithmetic addition, function evaluation. Within the context of pre-trained LMs, researchers have found that pre-trained embeddings and LMs can perform simple operations such as addition and subtraction when the number of digits is small, but fail when the numbers are larger \citep{naik-etal-2019-exploring,wallace-etal-2019-nlp,brown2020language}. \citet{10.1145/3411763.3451760} explore more complex mathematical (e.g.~$f(\bm{x}) = \bm{x} * \bm{x}$, what is $f(f(3))$?) reasoning problems and improve LM performance through serializing reasoning for the question.

\subsection{Question Answering}
Question answering (QA) aims to answer a given input question, often based on a context document.
QA can take a variety of formats, such as extractive QA (which identifies content from the context document containing the answer; e.g. SQuAD \citep{rajpurkar-etal-2016-squad}), multiple-choice QA (where the model has to pick from several choices; e.g. RACE \citep{lai-etal-2017-race}), and free-form QA (where the model can return an arbitrary textual string as a response; e.g. NarrativeQA~\citep{kocisky-etal-2018-narrativeqa}).
Generally, these different formats have been handled using different modeling frameworks. 
One benefit of solving QA problems with LMs, potentially using prompting methods, is that different formats of QA tasks can be solved within the same framework.
For example, \citet{khashabi-etal-2020-unifiedqa} reformulate many QA tasks as a text generation problem by fine-tuning seq2seq-based pre-trained models (e.g.~T5) and appropriate prompts from the context and questions. \citet{jiang2020when} take a closer look at such prompt-based QA systems using sequence to sequence pre-trained models (T5, BART, GPT2) and observe that probabilities from these pre-trained models on QA tasks are not very predictive of whether the model is correct or not.

\subsection{Text Generation}
Text generation is a family of tasks that involve generating text, usually conditioned on some other piece of information.
Prompting methods can be easily applied to these tasks by using \term{prefix prompts} together with autoregressive pre-trained LMs.
\citet{Radford2019LanguageMA} demonstrated impressive ability of such models to perform generation tasks such as text summarization and machine translation using prompts such as ``{translate to
french}, \texttt{[X]}, \texttt{[Z]}''. \citet{brown2020language} perform \term{in-context learning} (\S\ref{sec:tuning-free-prompting}) for text generation, creating a prompt with manual templates and augmenting the input with multiple \term{answered prompts}. \citet{schick2020fewshot} explore \term{fixed-prompt LM tuning} (\S\ref{sec:prompt-fix-fine-tuning}) for few-shot text summarization with manually crafted templates.
\cite{li2021prefix} investigate \term{fixed-LM prompt tuning} (\S\ref{sec:prompt-only-tuning}) for text summarization and data-to-text generation in few-shot settings, where learnable prefix tokens are prepended to the input while parameters in pre-trained models are kept frozen.
\citet{dou-etal-2021-gsum} explored the \term{prompt+LM tuning}  strategy (\S\ref{sec:prompt-fine-tuning}) on text summarization task, where learnable prefix prompts are used and initialized by different types of guidance signals, which can then be updated together with parameters of pre-trained LMs.

\subsection{Automatic Evaluation of Text Generation}

\citet{yuan2021bartscore} have demonstrated that prompt learning can be used for automated evaluation of generated texts.
Specifically, they conceptualize the evaluation of generated text as a text generation problem, modeled using a pre-trained sequence-to-sequence, and then use \term{prefix prompts} that bring
the evaluation task closer to the pre-training task.
They experimentally find that simply adding the phrase ``such as'' to the translated text when using pre-trained models can lead to a significant improvement in correlation on German-English
machine translation (MT) evaluation.

\subsection{Multi-modal Learning}
 \citet{DBLP:journals/corr/abs-2106-13884} shift the application of prompt learning from text-based NLP to the \term{multi-modal} setting (vision and language).
 Generally, they adopt the \term{fixed-LM prompt tuning} strategy together with \term{prompt augmentation} techniques.
 They specifically represent each image
as a sequence of continuous embeddings, and a pre-trained LM whose parameters are frozen is prompted with this prefix to generate texts such as image captions. 
Empirical results show few-shot learning ability: with the help of a few demonstrations (answered prompts), system can rapidly learn words for new objects and novel visual categories.

\subsection{Meta-Applications}

There are also a number of applications of prompting techniques that are not NLP tasks in and of themselves, but are useful elements of training strong models for any application.

\paragraph{Domain Adaptation}
Domain adaptation is the practice of adapting a model from one domain (e.g.~news text) to another (e.g.~social media text).
\citet{bendavid2021pada} use self-generated \term{domain related features} (DRFs) to augment the original text input and perform sequence tagging as a sequence-to-sequence problem using a seq2seq pre-trained model.

\paragraph{Debiasing} \citet{schick2021selfdiagnosis} found that LMs can perform self-diagnosis and self-debiasing based on biased or debiased instructions. For example, to self-diagnosis whether the generated text contains violent information, we can use the following template ``The following text contains violence. \texttt{[X]}\texttt{[Z]}". Then we fill \texttt{[X]} with the input text and look at the generation probability at \texttt{[Z]}, if the probability of ``Yes" is greater than ``No", then we would assume the given text contains violence, and vice versa. To perform debiasing when generating text, we first compute the probability of the next word $P(x_t|\bm{x}_{<t}; \theta)$ given the original input. Then we compute the probability of next word $P(x_t | [\bm{x}_{<t};\bm{x}_{\text{diagnosis}}];\theta)$ by appending self-diagnosis textual input to the original input as mentioned above. These two probability distributions for the next token can be combined to suppress the undesired attribute.

\paragraph{Dataset Construction}
\citet{schick2021generating} propose to use pre-trained LMs to generate datasets given certain instructions. As an example, suppose we have an unlabeled dataset in which each sample is a sentence. If we want to construct a dataset containing pairs of semantically similar sentences, then we can use the following template for each input sentence: ``Write two sentences that mean the same thing. \texttt{[X]}\texttt{[Z]}" and attempt to generate a sentence that shares the same meaning as the input sentence.

\subsection{Resources}
We also collect some useful resources for different prompt-based applications. 
\paragraph{Dataset} Some datasets specifically designed for few-shot and zero-shot learning are shown in Tab.~\ref{tab:resources-dataset}.

\begin{table*}[!ht]
\centering
\setlength\tabcolsep{3.5pt}
\renewcommand\arraystretch{1.1}
\footnotesize
\begin{tabular}{llll}
\toprule
\textbf{Task} & \textbf{Dataset} & \textbf{Setting} & \textbf{URL} \\
\midrule
    & Pronoun Disambiguation Problems \citenumber{DBLP:conf/aaaiss/Levesque11}   & Zero      & \href{https://cs.nyu.edu/~davise/papers/WinogradSchemas/PDPChallenge2016.xml}{https://cs.nyu.edu/~davise/papers/...} \\
    & Winograd Schema Challenge \citenumber{DBLP:conf/aaaiss/Levesque11}     & Zero      &  \href{https://cs.nyu.edu/~davise/papers/WinogradSchemas/WSCollection.xml}{https://cs.nyu.edu/~davise/papers/...}     \\
\multirow{-3}{*}{Commonsense Reasoning}          & CPRAG-102 \citenumber{DBLP:journals/tacl/Ettinger20}                                  & Zero      & \href{https://github.com/aetting/lm-diagnostics}{https://github.com/aetting/lm-diagnostics}      
\\
\midrule
& WNLaMPro \citenumber{DBLP:conf/aaai/SchickS20}  & Zero & \href{https://github.com/timoschick/am-for-bert}{https://github.com/timoschick/...}                                    \\
 & ROLE-88 \citenumber{DBLP:journals/tacl/Ettinger20}   & Zero  & \href{https://github.com/aetting/lm-diagnostics}{https://github.com/aetting/lm-diagnostics}                                   \\
\multirow{-3}{*}{Linguistic Capacity Probing}    & NEG-136 \citenumber{DBLP:journals/tacl/Ettinger20}                                   & Zero      & \href{https://github.com/aetting/lm-diagnostics}{https://github.com/aetting/lm-diagnostics}                                    \\
\midrule
& LAMA \citenumber{petroni-etal-2019-language}    & Zero      & \href{https://dl.fbaipublicfiles.com/LAMA/data.zip}{https://dl.fbaipublicfiles.com/LAMA/...}                                \\
 & Negated LAMA \citenumber{DBLP:conf/acl/KassnerS20}  & Zero      & \href{https://github.com/norakassner/LAMA\_primed\_negated}{https://github.com/norakassner/LAMA...}           \\
 & Misprimed LAMA \citenumber{DBLP:conf/acl/KassnerS20}   & Zero      & \href{https://github.com/norakassner/LAMA\_primed\_negated}{https://github.com/norakassner/LAMA...}                        \\
\multirow{-3}{*}{Fact Probing}                   & X-FACTR \citenumber{jiang-etal-2020-x}                                    & Zero      & \href{https://x-factr.github.io/}{https://x-factr.github.io/}                                                  \\
&  LAMA-TREx-easy-hard \citenumber{zhong2021optiprompt} & Zero      &  \href{https://github.com/princeton-nlp/OptiPrompt}{https://github.com/princeton-nlp/...}                                 \\
\midrule
& FLEX \citenumber{DBLP:journals/corr/abs-2107-07170} & Zero,Few & \href{https://github.com/allenai/flex}{https://github.com/allenai/flex} \\
\multirow{-2}{*}{Text Classification }
                             & FewGLUE \citenumber{schick2021its}  & Few       & \href{https://github.com/timoschick/fewglue}{https://github.com/timoschick/fewglue}                                       \\
\midrule
 & REALTOXICITYPROMPTS \citenumber{gehman2020realtoxicityprompts}  & Zero      &  \href{https://allenai.org/data/real-toxicity-prompts}{https://allenai.org/data/...}                              \\
\multirow{-2}{*}{General Conditional Gen.} & Natural-Instructions \citenumber{DBLP:journals/corr/abs-2104-08773}                       & Few,Full & \href{https://instructions.apps.allenai.org/}{https://instructions.apps.allenai.org/}\\
\bottomrule
\end{tabular}
\caption{\label{tab:resources-dataset}Few-shot and zero-shot datasets for prompt-based learning.}
\end{table*}

\paragraph{Prompts}
As shown in Tab.~\ref{tab:task-prompt}, we collect existing commonly-used prompts designed manually, which can be regarded as off-the-shelf resource for future research and applications.

\begin{table*}[pt]
\centering
\footnotesize
\renewcommand\arraystretch{1.1}
\setlength\tabcolsep{4.5pt}
\begin{tabular}{lll}
\toprule
\textbf{Task}                                          & \textbf{Example Prompt-Answer}                                                        & \textbf{Resource}                                                                              \\
\midrule
\multirow{6}{*}{Fact Probing}                 & \textbf{Prompt} Adolphe Adam died in \texttt{[Z]}.                                         &    \href{https://dl.fbaipublicfiles.com/LAMA/data.zip}{LAMA dataset}             \\
    & \textbf{Answer}
    $\mathcal{V}$                                         & \href{https://github.com/jzbjyb/LPAQA}{LPAQA dataset}                      \\
    & \textbf{Prompt} iPod Touch is produced by \texttt{[Z]}.                                    &   \href{https://x-factr.github.io/}{X-FACTR dataset}                                                                                     \\
    & \textbf{Answer} $\mathcal{V}$                                        &                                                  \\
 & \textbf{Prompt} The official language of Mauritius is \texttt{[Z]}.                        &                                                                                       \\
   & \textbf{Answer} $\mathcal{V}$                                         &                                                                                       \\
\midrule
\multirow{13}{*}{Text Classificatin}          & \textbf{Prompt} Which of these choices best describes the following                   & Meta \citenumber{zhong2021meta}                                            \\
    & document? "\texttt{[Class A]}", "\texttt{[Class B]}", "\texttt{[Class C]}".   &                                                                                       \\
    & \texttt{[X]}\texttt{[Z]} &
    \\
    & \textbf{Answer} \texttt{[Class A]}, \texttt{[Class B]}, \texttt{[Class C]}                          &                                                                                       \\
    & \textbf{Prompt} How is the text best described? : "\texttt{[Class A]}",                   &                                                    \\
   & “\texttt{[Class B]}” , or “\texttt{[Class C]}”. \texttt{[X]}\texttt{[Z]}                         &                                                                                       \\
  & \textbf{Answer} \texttt{[Class A]}, \texttt{[Class B]}, \texttt{[Class C]}                           &                                                                                       \\
 & \textbf{Prompt} This passage is about \texttt{[Z]}: \texttt{[X]}  &    \\
 & \textbf{Answer} \texttt{[Class A]}, \texttt{[Class B]}, \texttt{[Class C]}                           &                                                                                       \\
  & \textbf{Prompt} \texttt{[X]}. Is this review positive? \texttt{[Z]}                            &                                              \\
     & \textbf{Answer} Yes, No                                                               &                                                                                       \\
     & \textbf{Prompt} \texttt{[X]} It was \texttt{[Z]}.                                               &                                                                                       \\
 & \textbf{Answer} great, terrible                                                       &                                                                                       \\
\midrule                                              
\multirow{4}{*}{Natural Language Inference}   & \textbf{Prompt} \texttt{[X1]}? \texttt{[Z]}, \texttt{[X2]}                                         &                                       \\
  & \textbf{Answer} Yes, No, Maybe                                                       &                                    \\
 & \textbf{Prompt} \texttt{[X1]} \texttt{[Z]}, \texttt{[X2]}  &                                     \\
  & \textbf{Answer} Yes, No, Maybe                                                       &                                                                                       \\
\midrule
\multirow{6}{*}{Commonsense Reasoning}        & \textbf{Prompt} The trophy doesn’t ﬁt into the brown suitcase                         & \href{https://cs.nyu.edu/~davise/papers/WinogradSchemas/PDPChallenge2016.xml}{PDP dataset}  \\
 & because \texttt{[Z]} is too large.                                                & \href{https://cs.nyu.edu/~davise/papers/WinogradSchemas/WSCollection.xml}{WSC dataset}                                     \\  & \textbf{Answer} trophy, suitcase                                                      &     \href{https://github.com/aetting/lm-diagnostics}{CPRAG-102 dataset}                                                                                  \\ & \textbf{Prompt} Ann asked Mary what time the library closes,                          &   \\
 & because \texttt{[Z]} had forgotten.                                               &         \\
  & \textbf{Answer} Ann, Mary                                                             &            \\
\midrule                                              
\multirow{6}{*}{Linguistic Knowledge Probing} & \textbf{Prompt} A robin is a \texttt{[Z]}.                                                  & \href{https://github.com/timoschick/am-for-bert}{WNLaMPro dataset} \\
 & \textbf{Answer} bird, tree                                                            & \href{https://github.com/aetting/lm-diagnostics}{ROLE-88 dataset}  \\
  & \textbf{Prompt} A robin is not a \texttt{[Z]}.                                              & \href{https://github.com/aetting/lm-diagnostics}{NEG-136 dataset}  \\
 & \textbf{Answer} bird, tree     &      \\
 & \textbf{Prompt} New is the opposite of \texttt{[Z]}.                                       &    \\
 & \textbf{Answer} old, young, current     &      \\
\midrule                                              
\multirow{6}{*}{Named Entity Recognition}     & \textbf{Prompt-Pos} \texttt{[X]} \texttt{[Span]} is a \texttt{[Z]} entity.                            & TemplateNER \citenumber{cui2021templatebased}   \\
  & \textbf{Prompt-Neg} \texttt{[X]} \texttt{[Span]} is not a named entity.                          &                                           \\
   & \textbf{Answer} person, location, organization, miscellaneous                         &                                     \\       
   & \textbf{Prompt-Pos} The entity type of \texttt{Span} is \texttt{[Z]}.                         &                                         \\
   & \textbf{Prompt-Neg} \texttt{[X]} The entity type of \texttt{[Span]} is none entity.              &                                                                                       \\
   & Answer person, location, organization, miscellaneous                         &                                                                                       \\
\midrule                                              
\multirow{5}{*}{Question Answering}           & \textbf{Prompt} \texttt{[Question]} \texttt{[Passage]} \texttt{[Z]}   &     \\
 & \textbf{Prompt} \texttt{[Passage]} According to the passage, \texttt{[Question]}       &       \\
 & \texttt{[Z]}  & \\
  & \textbf{Prompt} Based on the following passage, \texttt{[Question]} \texttt{[Z]}.  &                                                                                       \\
  & \texttt{[Passage]}& \\
\midrule                                              
\multirow{3}{*}{Summarization}                & \textbf{Prompt} Text: \texttt{[X]} Summary: \texttt{[Z]}                                        & BARTScore \citenumber{yuan2021bartscore}                                             \\
 & \textbf{Prompt} \texttt{[X]} TL;DR: \texttt{[Z]}                                                &     \\
  & \textbf{Prompt} \texttt{[X]} In summary, \texttt{[Z]}                                             &                              \\
\midrule                                              
\multirow{4}{*}{Machine Translation}          & \textbf{Prompt} French: \texttt{[French sentence]} English:                                &       \\
  & \textbf{Prompt} A French sentence is provided: \texttt{[French sentence]}                  &                                                                     \\     
  & The French translator translates the sentence into English: \texttt{[Z]}          &                                      \\
  & \textbf{Prompt} \texttt{[French sentence]} = \texttt{[Z]}                                       &    \\
\bottomrule
\end{tabular}
\caption{\label{tab:task-prompt} Commonly used prompts and answers for different tasks. \texttt{[X]} and \texttt{[Z]} denote slots for input and answer respectively. $\mathcal{V}$ denotes the vocabulary of the LM. More prompts for each task can be found using the \textbf{Resource} column.}
\end{table*}

\begin{table*}[t]
  \centering
  \footnotesize
    \begin{tabular}{p{3cm}p{2.5cm}p{4cm}p{5cm}}
    \toprule
    \textbf{Prompt Concept} & \textbf{Relevant Topic} & \textbf{Commonality} & \textbf{Peculiarity} \\
    \midrule
    \multirow{1}[1]{*}[-1.5em]{\makecell[{{p{3cm}}}]{Prompt Ensembling \citenumber{jiang-etal-2020-know,schick2021exploiting}}} & \multirow{1}[1]{*}[-1.5em]{\makecell[{{p{2.5cm}}}]{Ensemble Learning \citenumber{DBLP:conf/ijcai/TingG97,zhou2002ensembling}}} & \multirow{1}[1]{*}[-1.5em]{\makecell[{{p{4cm}}}]{Combine results of multiple systems to get better performance}} & In prompt ensembling, multiple predictions result from different prompt variants. This contrasts with architecture or feature variations, each of which requires separate training.\\
    \midrule
    \multirow{2}[2]{*}[-2em]{\makecell[{{p{3cm}}}]{Prompt Augmentation \citenumber{brown2020language,gao2021making}}} & Few-shot Learning \citenumber{DBLP:conf/nips/SnellSZ17,DBLP:conf/icml/FinnAL17} & Use few examples to learn generalized rules & Prompt augmentation is a specific subset of few-shot learning.\\ 
    \cmidrule{2-4}
          & Larger-context Learning~\citenumber{cao-etal-2018-retrieve,guu2020realm} & Introduce larger context to aid the learning process & Additional information introduced in larger-context learning is not necessarily the labeled data. \\
    \midrule
    Discrete Prompt Search \citenumber{jiang-etal-2020-know,autoprompt:emnlp20} & Query reformulation \citenumber{nogueira2017task,nogueira2017task} & Reformulate the input into a query form & Query reformulation commonly focuses on information extraction and question answering tasks, while prompt learning can be applied to a variety of NLP tasks \\
    \midrule
    Discrete Prompt Fine-tuning \citenumber{gao2021making} & QA-based multi-task learning \citenumber{mccann2018natural,li-etal-2020-unified} & Reformulate many tasks into an QA form & QA-based formulations aim to solve different tasks through question answering, while prompting additionally targets full use of pre-trained models. \\
    \midrule
    Continuous Prompt Fine-tuning \citenumber{liu2021ptuning,dou-etal-2021-gsum} & Controlled Text Generation \citenumber{yu2020survey,kikuchi-etal-2016-controlling,sennrich-etal-2016-controlling} & Input is augmented with additional inputs to control the generation process & Controlled generation targets generation of a particular type of text while prompt learning uses prompts to specify the task itself. \\
    \midrule    
    \multirow{3}[2]{*}[-4em]{\makecell[{{p{3cm}}}]{Prompt-based downstream task learning \citenumber{schick2021exploiting,yuan2021bartscore}}} & \multirow{1}[1]{2.5cm}[-1.5em]{Supervised Attention~\citenumber{liu-etal-2016-neural,sugano2016seeing}}  & \multirow{1}[1]{4cm}[-1em]{Require external hint to remind the model of which part information should be focused on} & Research works on supervised attention usually target at salient information from an image or text, while prompt learning aims to utilize relevant knowledge from the pre-trained model. \\
      \cmidrule{2-4}
      & Data augmentation \citenumber{fadaee-etal-2017-data,DBLP:conf/nips/RatnerEHDR17}  & Improving downstream tasks' performance by introducing additional samples & Data augmentation introduce additional training samples in an explicit way while prompts can be regarded as highly-condensed training samples \citenumber{le-scao-rush-2021-many}. \\

    
    \bottomrule
    \end{tabular}%
    \caption{Other research topics relevant to prompting methods.}
  \label{tab:addlabel}%
\end{table*}%

\section{Prompt-relevant Topics}
\label{sec:related}
What is the essence of prompt-based learning and how does it relate to other learning methods?
In this section, we connect prompt learning with other similar learning methods.


\paragraph{Ensemble Learning}
\term{Ensemble learning} \cite{DBLP:conf/ijcai/TingG97,zhou2002ensembling} is a technique that aims to improve the performance of a task by taking advantage of the complementarity of multiple systems.
Generally, the different systems used in an ensemble result from different choices of architectures, training strategies, data ordering, and/or random initialization.
In prompt ensembling (\S\ref{sec:prompt-ensembling}), the choice of prompt templates becomes another way to generate multiple results to be combined.
This has the clear advantage that this does not necessarily require training the model multiple times.
For example, when using discrete prompts, these prompts can simply be changed during the inference stage \citep{jiang-etal-2020-know}.

\paragraph{Few-shot Learning}
\term{Few-shot learning} aims to learn a machine learning system in the data-scarce scenarios with few training samples. 
There are a wide variety of methods to achieve few-shot learning including model agnostic meta-learning~\cite{finn2017model} (learning features rapidly adaptable to new tasks), embedding learning~\cite{bertinetto2016learning} (embedding each sample in a lower-dimensional space where similar samples are close together), memory-based learning~\cite{kaiser2017learning} (representing each sample by a weighted average of contents from the memory) etc.~\cite{wang2020generalizing}.
Prompt augmentation can be regarded as another way to achieve few-shot learning (a.k.a.~priming-based few-shot learning \cite{kumar2021reordering}).
Compared to previous methods, prompt augmentation directly prepends several labeled samples to the currently-processed sample elicit knowledge from pre-trained LMs even without any parameter tuning.

\paragraph{Larger-context Learning}
\term{Larger-context learning} aims to improve the system's performance by augmenting the input with additional contextual information, e.g.~retrieved from the training set \cite{cao-etal-2018-retrieve} or external data sources \cite{guu2020realm}.
Prompt augmentation can be regarded as adding relevant labeled samples into the input, but a minor difference is in larger-context learning, the introduced context is not necessarily labeled data.

\paragraph{Query Reformulation}
\term{Query reformulation} \cite{mathieu-sabatier-1986-interfacile,daume-iii-brill-2004-web} is commonly used in information retrieval \cite{nogueira2017task} and question answering tasks \cite{buck2017ask,vakulenko-etal-2020-wrong}, which aim to elicit more relevant texts (documents or answers) by expanding the input query with related query terms \cite{hassan-2013-identifying} or generating paraphrases.
There are several commonalities between prompt-based learning and query reformulation, for example (1) both aim to make better use of some existing knowledge bases by asking a right questions
(2) the knowledge bases are usually a black-box, not available to the users, so researchers must learn how to probe it optimally based on solely questions.

There are also differences:
the knowledge base in traditional query reformulation problems is usually a search engine \cite{nogueira2017task}, or QA system \cite{buck2017ask}. By contrast,  for prompt-based learning, we usually define this knowledge base as an LM, and need to find the appropriate query to elicit an appropriate answer from it.
 The input reformulation in prompt learning has changed the form of tasks. For example, an original text classification task has been converted into a cloze question problem, therefore bringing  additional complexity regarding how to (1) make an appropriate task formulation, and (2) change the modeling framework accordingly.
 These steps are not required in traditional query formulation.
Despite these discrepancies, some methodologies from query reformulation research still can be borrowed for prompt learning, such as decomposing input query into multiple sub-queries \cite{nogueira2019multi}, similar to prompt decomposition.

\paragraph{QA-based Task Formulation}
\term{QA-based task formulation} aims to conceptualize different NLP tasks as a question-answering problem.
\cite{kumar2016ask,mccann2018natural} are earlier works that attempt to unify multiple NLP tasks into a QA framework.
Later, this idea has been further explored in information extraction \cite{li-etal-2020-unified,wu-etal-2020-corefqa} and text classification \cite{chai2020description}.
These methods are very similar to the prompting methods introduced here in that they use textual questions to specify which task is to be performed.
However, one of the key points of prompting methods is how to better use the knowledge in pre-trained LMs, and these were not covered extensively on previous works advocating for QA formulations.

\paragraph{Controlled Generation}
\term{Controlled generation} aims to incorporate various types of guidance beyond the input text into the generation model \cite{yu2020survey}.
Specifically, the guidance signals could be \emph{style tokens} \cite{sennrich-etal-2016-improving,fan-etal-2018-controllable}, \emph{length specifications} \cite{kikuchi-etal-2016-controlling}, \emph{domain tags} \cite{chu-etal-2017-empirical}, or any variety of other pieces of information used to control of the generated text.
It could also be \textit{keywords} \cite{saito2020abstractive}, \textit{relation triples} \cite{zhu2020enhancing} or even \textit{highlighted phrases or sentences} \cite{grangier-auli-2018-quickedit,liu-etal-2021-refsum} to plan the content of generated texts.
In a way, many of the prompting methods described here are a type of controllable generation, where the prompt is usually used to specify the \emph{task itself}.
Thus, it is relatively easy to find commonalities between the two genres:
(1) both add extra information to the input text for better generation, and these additional signals are (often) learnable parameters.
(2) If ``controlled generation'' is equipped with seq2seq-based pre-trained models (e.g., BART), then it is can be regarded as prompt learning with input-dependent prompts and the \term{prompt+LM fine-tuning} strategy (\S\ref{sec:prompt-fine-tuning}), e.g. \textit{GSum} \cite{dou-etal-2021-gsum}, where both the prompt's and pre-trained LM's parameters can be tuned.

Also, some clear discrepancies between controlled generation and prompt-based text generation are:
(1) In controlled generation work, the control is generally performed over the style or content of the generations \cite{fan-etal-2018-controllable,dou-etal-2021-gsum} while the underlying task remains the same. They don't necessarily require a pre-trained model.
In contrast, the main motivation for using prompts for text generation is to specify the task itself and better utilize the pre-trained model.
(2) Moreover, most of the current work on prompt learning in text generation shares a dataset- or task-level prompt \cite{li2021prefix}. Only very few works have explored input-dependent ones \cite{DBLP:journals/corr/abs-2106-13884}. However, this is a common setting and effective in the controlled text generation, which may provide valuable direction for the future work on prompt learning.


\paragraph{Supervised Attention}
Knowing to pay attention to the important information is a key step when extracting useful information from objects such as long text sequences \cite{liu-etal-2016-neural,sood2020improving}, images \cite{sugano2016seeing,zhang2020human}, or knowledge bases \cite{yu2020survey,dou-etal-2021-gsum}).
\term{Supervised attention} \cite{liu2017exploiting}  aims to provide explicit supervision over the attention of models based on the fact that completely data-driven attention can overfit to some artifacts \cite{liu2017attention}.
In this respect, prompt learning and supervised attention share ideas that both aim to extract salient information with some clues, which need to be provided separately.
To solve this problem, supervised attention methods tried to use additional loss functions to learn to predict gold attention on a manually labeled corpus \cite{jiang2015salicon,qiao2018exploring,gan2017vqs}.
Research on prompt learning may also borrow ideas from this literature.

\paragraph{Data Augmentation}
Data augmentation is a technique that targets increasing the amount of data that can be used for training by making modifications to existing data \cite{fadaee-etal-2017-data,DBLP:conf/nips/RatnerEHDR17}.
As recently observed by \cite{scao2021many}, adding prompts can achieve a similar accuracy improvement to the addition of 100s of data points on average across classification tasks, which suggests that using prompts for a downstream task is similar to conducting data augmentation implicitly.



\section{Challenges} \label{sec:challenges}
Although prompt-based learning has shown significant potential among different tasks and scenarios, several challenges remain, some of which we detail below.

\subsection{Prompt Design}

\paragraph{Tasks beyond  Classification and Generation}
Most existing works about prompt-based learning revolve around either text classification or generation-based tasks.
Applications to information extraction and text analysis tasks have been discussed less, largely because the design of prompts is less straightforward.
We expect that applying prompting methods to these tasks in the future it will require either reformulating these tasks so that they can be solved using classification or text generation-based methods, or performing effective answer engineering that expresses structured outputs in an appropriate textual format.

\paragraph{Prompting with Structured Information}
In many NLP tasks, the inputs are imbued with some variety of structure, such as tree, graph, table, or relational structures.
How to best express these structures in prompt or answer engineering is a major challenge.
Existing works \cite{DBLP:journals/corr/abs-2104-07650} make a step by making prompts with additional marks to encode lexical information, such as entity markings.
\citet{aghajanyan2021htlm} present structured prompts based on hyper text markup language for more fine-grained web text generation. However, moving beyond this to more complicated varieties of structure is largely unexplored, and a potentially interesting research area.

\paragraph{Entanglement of Template and Answer}
The performance of a model will depend on \emph{both} the templates being used and the answer being considered.
How to simultaneously search or learn for the best combination of template and answer remains a challenging question.
Current works typically select answers before select template \citep{gao2021making, autoprompt:emnlp20}, but \citet{Hambardzumyan2021WARPWA} have demonstrated the initial potential of simultaneously learning both.

\subsection{Answer Engineering}

\paragraph{Many-class and Long-answer Classification Tasks}
For classification-based tasks, there are two main challenges for answer engineering: (a) When there are too many classes, how to select an appropriate answer space becomes a difficult combinatorial optimization problem. (b) When using multi-token answers, how to best decode multiple tokens using LMs remains unknown, although some multi-token decoding methods have been proposed \citep{jiang-etal-2020-x}.

\paragraph{Multiple Answers for Generation Tasks}
For text generation tasks, qualified answers can be semantically equivalent but syntactically diverse.
So far, almost all works use prompt learning for text generation relying solely on a single answer, with only a few exceptions \citep{jiang-etal-2020-know}.
How to better guide the learning process with multiple references remains a largely open research problem.

\subsection{Selection of Tuning Strategy}
As discussed in \S\ref{sec:tuning}, there are a fairly wide variety of methods for tuning parameters of prompts, LMs, or both.
However, given the nascent stage of this research field, we still lack a systematic understanding of the tradeoffs between these methods.
The field could benefit from systematic explorations such as those performed in the pre-train and fine-tune paradigm regarding the tradeoffs between these different strategies \citep{peters-etal-2019-tune}.

\subsection{Multiple Prompt Learning}

\begin{figure}
    \centering
 
    \includegraphics[height=0.15\linewidth]{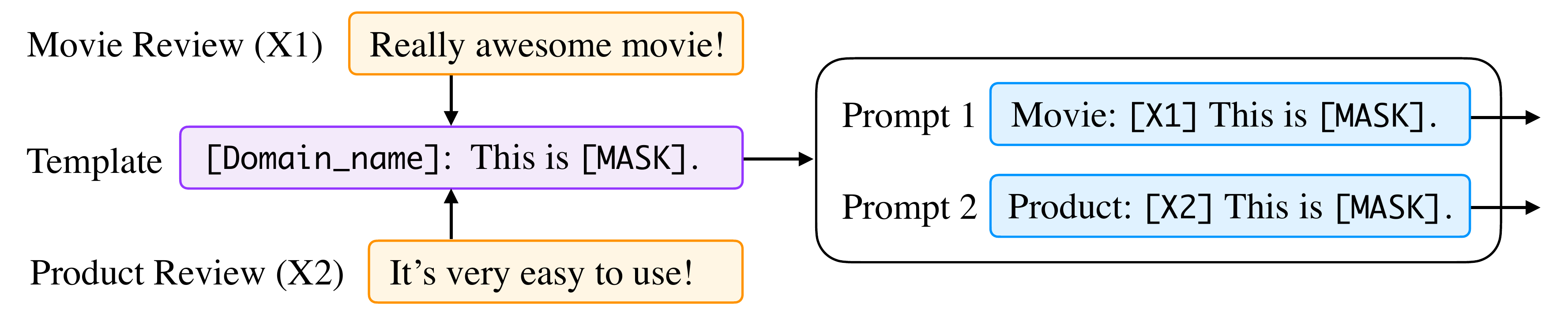}
 
    \caption{Multi-prompt learning for multi-task, multi-domain or multi-lingual learning. We use different colors to differentiate different components as follows. `` \includegraphics[scale=0.25]{fig/multi-prompt/input.pdf} " for input text, 
    `` \includegraphics[scale=0.25]{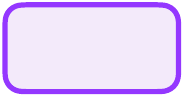} " for template, 
    `` \includegraphics[scale=0.25]{fig/multi-prompt/prompt.pdf} " for prompt.}
    \label{fig:prompt-sharing}
\end{figure}

\paragraph{Prompt Ensembling} 
In prompt ensembling methods, the space and time complexity increase as we consider more prompts.
How to distill the knowledge from different prompts remains underexplored.
\citet{schick2020fewshot, schick2021exploiting, schick2021its} use an ensemble model to annotate a large dataset to distill the knowledge from multiple prompts.

In addition, how to select ensemble-worthy prompts is also under-explored.
For text generation tasks, the study of prompt ensemble learning has not been performed so far, probably because ensemble learning in text generation itself is relatively complicated. To remedy this problem, some recently proposed neural ensembling methods such as \term{Refactor} \cite{liu-etal-2021-refsum} could be considered as a method for prompt ensembling in text generation tasks.

\paragraph{Prompt Composition and Decomposition}
Both prompt composition and decomposition aim to break down the difficulty of a complicated task input by introducing multiple sub-prompts.
In practice, how to make a good choice between them is a crucial step.
Empirically, for those token~\cite{ma-hovy-2016-end} or span~\cite{fu2021spanner} prediction tasks (e.g., NER), prompt decomposition can be considered, while for those span relation prediction~\cite{lee-etal-2017-end} tasks (e.g., entity coreference), prompts composition would be a better choice.
In the future, the general idea of de-/composing can be explored in more scenarios.

\paragraph{Prompt Augmentation}
Existing prompt augmentation methods are 
limited by the input length, i.e., feeding too many demonstrations to input is infeasible.
Therefore, how to select informative demonstrations, and order them in an appropriate is an interesting but challenging problem \cite{kumar2021reordering}.

\paragraph{Prompt Sharing}
\label{sec:prompt-sharing}
All the above considerations refer to the application of prompt in a single task, domain or language.
We may also consider \term{prompt sharing}, where prompt learning is applied to multiple tasks, domains, or languages.
Some key issues that may arise include how to design individual prompts for different tasks, and how to modulate their interaction with each other.
So far this field has not been explored.
Fig.\ref{fig:prompt-sharing} illustrates a simple multiple prompt learning strategy for multiple tasks, where prompt templates are partially shared.

\subsection{Selection of Pre-trained Models}
With plenty of pre-trained LMs to select from (see \S\ref{sec:lm}), how to choose them to better leverage prompt-based learning is an interesting and difficult problem.
Although we have conceptually introduced (\S\ref{sec:typical-architecture}) how different paradigms of pre-trained models are selected for diverse NLP tasks, there are few to no systematic comparisons of the benefits brought by prompt-based learning for different pre-trained LMs.

\subsection{Theoretical and Empirical Analysis of Prompting}
Despite their success in many scenarios, theoretical analysis and guarantees for prompt-based learning are scarce. \citet{wei2021pretrained} showed that soft-prompt tuning can relax the non-degeneracy assumptions (the generation probability of each token is linearly independent) needed for downstream recovery (i.e. recover the ground-truth labels of the downstream task.), making it easier to extract task-specific information. \citet{DBLP:conf/iclr/SaunshiMA21} verified that text classification tasks can be reformulated as sentence completion tasks, thus making language modeling a meaningful pre-training task. \citet{scao2021many} empirically show that prompting is often worth 100s of data points on average across classification tasks.

\subsection{Transferability of Prompts}
Understanding the extent to which prompts are specific to the model and improving the transferability of prompts are also important topics. \citep{perez2021true} show that prompts selected under tuned few-shot learning scenario (where one has a larger validation set to choose prompts) generalize well across models of similar sizes while prompts selected under true few-shot learning scenario (where one only has a few training samples) do not generalize as effectively as the former setting among models with similar sizes.
The transferability is poor when the model sizes are quite different in both scenarios.

\subsection{Combination of Different Paradigms}
Notably, much of the success of the prompting paradigm is built on top of pre-trained models that were developed for the pre-train and fine-tune paradigm, such as BERT. However, are the pre-training methods that are effective for the latter applicable as-is to the former, or can we entirely re-think our pre-training methods to further improve accuracy or ease of applicability to prompting-based learning?
This is an important research question that has not been covered extensively by the literature.

\subsection{Calibration of Prompting Methods}
Calibration \cite{gleser1996measurement}  refers to the ability of a model to make good probabilistic predictions.
When using the generation probability of the pre-trained LMs (e.g., BART) to predict the answer, we need to be careful since the probability distribution is typically not well calibrated. 
\citet{jiang2020when} observed the probabilities of pre-trained models (e.g., BART, T5, GPT-2) on QA tasks are well calibrated.
\citet{zhao2021calibrate} identify three pitfalls (majority label bias, recency bias and common token bias) that lead the pre-trained LMs to be biased toward certain answers when provided answered prompts. For example, if the final answered prompt has a positive label, then this will bias the model towards predicting positive words. To overcome those pitfalls, \citet{zhao2021calibrate} first use context-free input (e.g. the prompt would be ``Input: Subpar acting. Sentiment: Negative\textbackslash n Input: Beautiful film. Sentiment: Positive\textbackslash n Input: N/A. Sentiment:") to get the initial probability distribution $P_0$, then they use the real input (e.g. the prompt would be ``Input: Subpar acting. Sentiment: Negative\textbackslash n Input: Beautiful film. Sentiment: Positive\textbackslash n Input: Amazing. Sentiment:") to get the probability distribution $P_1$. Finally, these two distributions can be used to get a calibrated generation probability distribution. However, this method has two drawbacks: (1) it comes with the overhead of finding proper context-free input (e.g.~whether to use ``N/A" or ``None") and (2) the probability distribution of the underlying pre-trained LM is still not calibrated.

Even though we have a calibrated probability distribution, we also need to be careful when we assume a single gold answer for an input. This is because that all surface forms of a same object will compete for finite probability mass~\citep{holtzman2021surface}. For example, if we consider the gold answer to be ``Whirlpool bath", the generation probability of it will typically be low since the word ``Bathtub" shares the same meaning and it will take over a large probability mass. To address this issue, we could either (i) perform answer engineering to construct a comprehensive gold answer set using paraphrasing methods (\S\ref{sec:discrete-answer-search}) or (ii) calibrate the probability of a word based on its prior likelihood within the context~\citep{holtzman2021surface}.

\section{Meta Analysis}
\label{sec:meta-analysis}

\begin{table*}
\scriptsize
\renewcommand\arraystretch{2}\arrayrulecolor{black}
\captionsetup{singlelinecheck=false, font=black, labelfont=sc, labelsep=quad}
\caption{Timeline of prompt-based learning. The time for each paper is based on its first arXiv version (if exists) or estimated submission time. A web-version can refer to \href{http://pretrain.nlpedia.ai/timeline.html}{NLPedia-Pretrain}. Works in \textcolor{red}{red} consider natural language understanding (NLU) tasks; works in \textcolor{blue}{blue} consider natural language generation (NLG) tasks; works in \textcolor{tlgreen}{green} consider both NLU tasks and NLG tasks.}\vskip -1.5ex
\begin{tabular}{@{\,}r <{\hskip 2pt} !{\foo}
>{\raggedright\arraybackslash}p{7cm}@{\,}r <{\hskip 2pt} !{\foo} >{\raggedright\arraybackslash}p{5.5cm}}
\toprule
\addlinespace[1.5ex]
2018.06.07 & \textcolor{red}{LMComm} \citep{DBLP:journals/corr/abs-1806-02847} & 2021.04.14 & \textcolor{red}{Soft} \citep{qin-eisner-2021-learning} \\

2019.02.14 & \textcolor{tlgreen}{GPT-2} \citep{Radford2019LanguageMA} & 2021.04.15 & \textcolor{blue}{DINO} \citep{schick2021generating} \\

2019.04.14 & \textcolor{red}{WNLaMPro} \citep{DBLP:conf/aaai/SchickS20} & 2021.04.15 & \textcolor{red}{AdaPrompt} \citep{DBLP:journals/corr/abs-2104-07650} \\ 

2019.07.31 & \textcolor{red}{LMDiagnose} \citep{DBLP:journals/tacl/Ettinger20} & 2021.04.16 & \textcolor{tlgreen}{PMI$_\text{DC}$} \citep{holtzman2021surface} \\

2019.08.20 & \textcolor{blue}{AdvTrigger} \citep{DBLP:conf/emnlp/WallaceFKGS19} & 2021.04.18 & \textcolor{red}{Prompt-Tuning} \citep{lester2021power} \\

2019.09.02 & \textcolor{red}{CohRank} \citep{DBLP:conf/emnlp/DavisonFR19} & 2021.04.18 & \textcolor{blue}{Natural-Instr} \citep{DBLP:journals/corr/abs-2104-08773} \\

2019.09.03 & \textcolor{red}{LAMA} \citep{petroni-etal-2019-language} & 2021.04.18 & \textcolor{red}{OrderEntropy} \citep{Lu2021FantasticallyOP} \\

2019.09.11 & \textcolor{blue}{CTRL} \citep{DBLP:journals/corr/abs-1909-05858} & 2021.04.18 & \textcolor{red}{FewshotSemp} \citep{Shin2021ConstrainedLM} \\

2019.10.23 & \textcolor{tlgreen}{T5} \citep{JMLR:v21:20-074} & 2021.04.26 & \textcolor{tlgreen}{PanGu-$\alpha$} \citep{zeng2021pangualpha}\\

2019.11.08 & \textcolor{red}{Neg \& Misprim} \citep{DBLP:conf/acl/KassnerS20} & 2021.05.24 & \textcolor{red}{TrueFewshot} \citep{perez2021true} \\

2019.11.28 & \textcolor{red}{LPAQA} \citep{jiang-etal-2020-know} & 2021.05.24 & \textcolor{red}{PTR} \citep{han2021ptr} \\
 
2019.12.10 & \textcolor{red}{ZSC} \citep{DBLP:journals/corr/abs-1912-10165} & 2021.06.03 & \textcolor{red}{TemplateNER} \citep{cui2021templatebased} \\

2020.01.21 & \textcolor{red}{PET-TC} \citep{schick2021exploiting} & 2021.06.03 & \textcolor{red}{PERO} \citep{kumar2021reordering} \\

2020.03.10 & \textcolor{red}{ContxFP} \citep{Petroni2020HowCA} & 2021.06.16 & \textcolor{red}{PromptAnalysis} \citep{wei2021pretrained} \\

2020.05.02 & \textcolor{red}{UnifiedQA} \citep{khashabi-etal-2020-unifiedqa} & 2021.06.20 & \textcolor{tlgreen}{CPM-2} \citep{DBLP:journals/corr/abs-2106-10715} \\

2020.05.22 & \textcolor{tlgreen}{RAG} \citep{DBLP:conf/nips/LewisPPPKGKLYR020} & 2021.06.21 & \textcolor{red}{BARTScore} \citep{yuan2021bartscore} \\

2020.05.28 & \textcolor{tlgreen}{GPT-3} \citep{brown2020language}  & 2021.06.24 & \textcolor{red}{NullPrompt} \citep{logan2021cutting}\\

2020.09.08  & \textcolor{red}{CommS2S}\citep{yang-etal-2020-designing} & 2021.06.25 &  \textcolor{blue}{Frozen} \citep{DBLP:journals/corr/abs-2106-13884} \\

2020.09.15 &  \textcolor{red}{PET-SGLUE} \citep{schick2021its} & 2021.07.05 & \textcolor{red}{ERNIE-B3} \citep{DBLP:journals/corr/abs-2107-02137} \\

2020.09.24 & \textcolor{blue}{ToxicityPrompts} \citep{gehman2020realtoxicityprompts} & 2021.07.07 & \textcolor{blue}{Codex} \citep{chen2021evaluating} \\

2020.10.07 & \textcolor{red}{WhyLM} \citep{DBLP:conf/iclr/SaunshiMA21} & 2021.07.14 & \textcolor{tlgreen}{HTLM} \citep{aghajanyan2021htlm} \\

2020.10.13 & \textcolor{red}{X-FACTR} \citep{jiang-etal-2020-x} & 2021.07.15 & \textcolor{red}{FLEX} \citep{DBLP:journals/corr/abs-2107-07170} \\

2020.10.26 & \textcolor{red}{Petal} \citep{DBLP:conf/coling/SchickSS20} \\

2020.10.29 & \textcolor{red}{AutoPrompt} \citep{autoprompt:emnlp20} \\

2020.12.08 & \textcolor{blue}{CTRLsum} \citep{DBLP:journals/corr/abs-2012-04281} \\

2020.12.22 & \textcolor{blue}{PET-Gen} \citep{schick2020fewshot} \\

2020.12.31 & \textcolor{red}{LM-BFF} \citep{gao2021making} \\

2021.01.01 & \textcolor{red}{WARP} \citep{Hambardzumyan2021WARPWA} \\

2021.01.01 & \textcolor{blue}{Prefix-Tuning} \citep{li2021prefix} \\

2021.01.17 & \textcolor{tlgreen}{KATE} \citep{liu2021makes} \\

2021.02.15 & \textcolor{tlgreen}{PromptProg} \citep{10.1145/3411763.3451760} \\

2021.02.19 & \textcolor{red}{ContxCalibrate} \citep{zhao2021calibrate} \\

2021.02.24 & \textcolor{red}{PADA} \citep{bendavid2021pada} \\

2021.02.27 & \textcolor{blue}{SD} \citep{schick2021selfdiagnosis} \\

2021.03.09 & \textcolor{red}{BERTese} \citep{haviv-etal-2021-bertese} \\

2021.03.15 & \textcolor{red}{Prompt2Data} \citep{scao2021many} \\

2021.03.18 & \textcolor{red}{P-Tuning} \citep{liu2021ptuning} \\

2021.03.18 & \textcolor{red}{GLM} \citep{du2021nlp} \\

2021.03.22 & \textcolor{red}{ADAPET} \citep{tam2021improving} \\

2021.04.10 & \textcolor{red}{Meta} \citep{zhong2021meta} \\

2021.04.12 & \textcolor{red}{OptiPrompt} \citep{zhong2021optiprompt} \\

\end{tabular} \label{tab:timeline}
\end{table*}

In this section, we aim to give a quantitative birds-eye view of existing research on prompting methods by performing a meta analysis over existing research works along different dimensions.

\subsection{Timeline}
We first summarize a number of existing research papers in a chronological order with in the form of a \term{timeline}, which hopefully, help researchers who are new to this topic understand the evolution of the field.

\subsection{Trend Analysis}
We also calculate the number of prompt-based papers with respect to different dimensions.

\begin{figure*}[!ht]
    \centering
    \subfloat[Year.]{
    \includegraphics[height=0.185\linewidth]{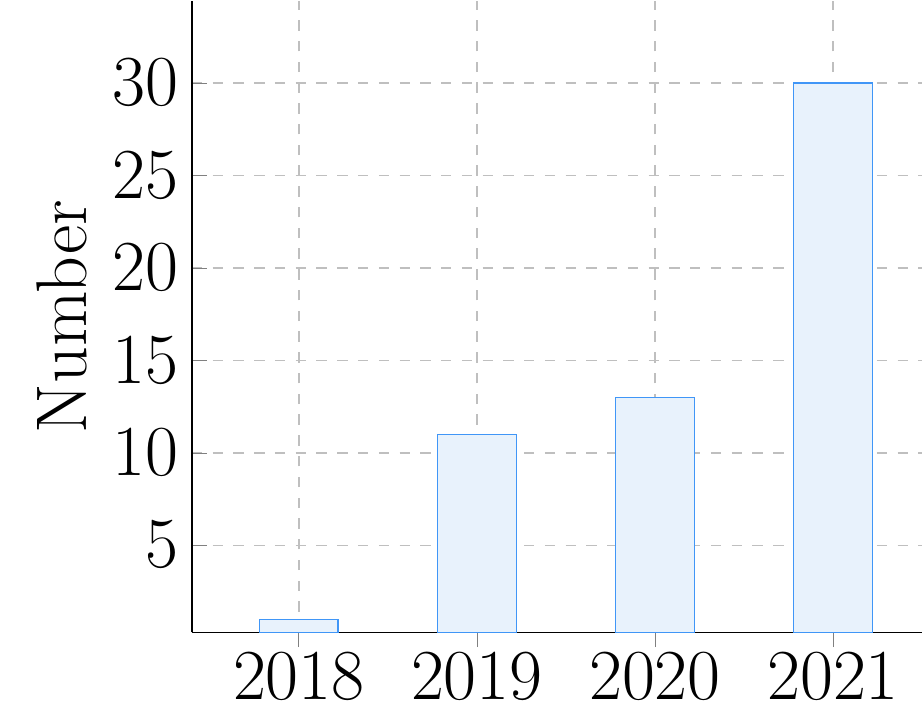}
    }        
    \subfloat[Task.]{  
    \includegraphics[height=0.185\linewidth]{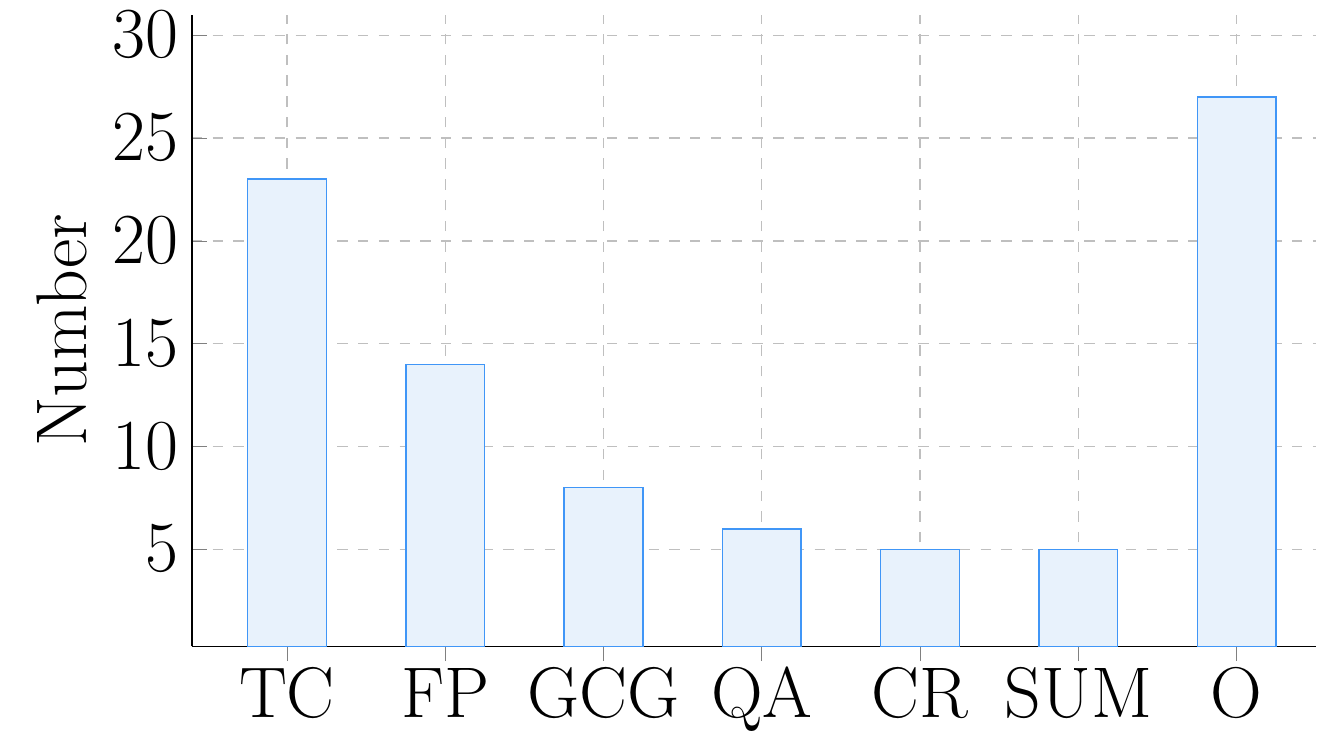}
    }        
    \subfloat[Automatic Search.]{  
    \includegraphics[height=0.185\linewidth]{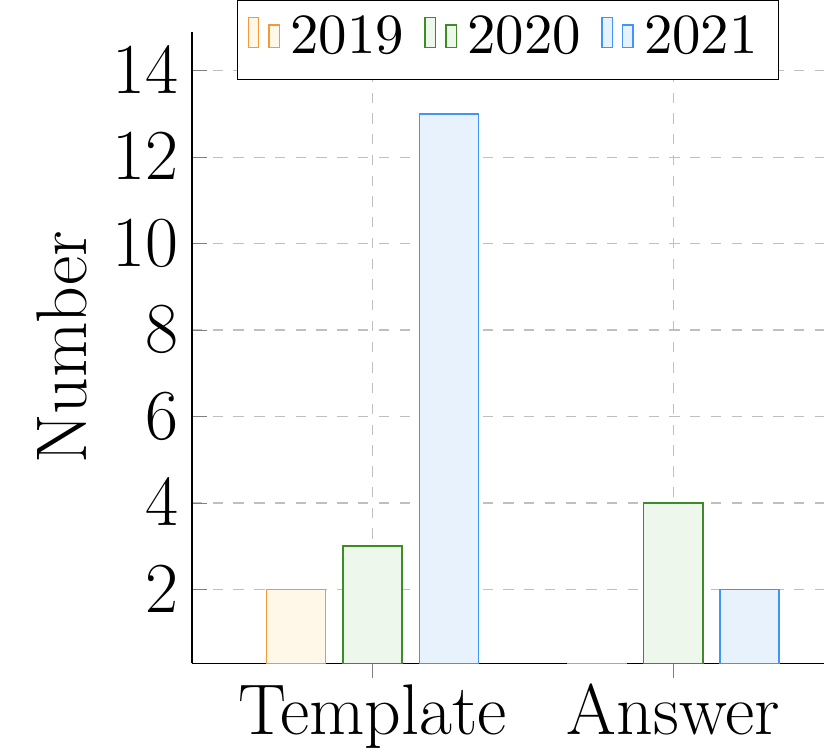}
    }        
    \subfloat[Search Space.]{ 
    \includegraphics[height=0.185\linewidth]{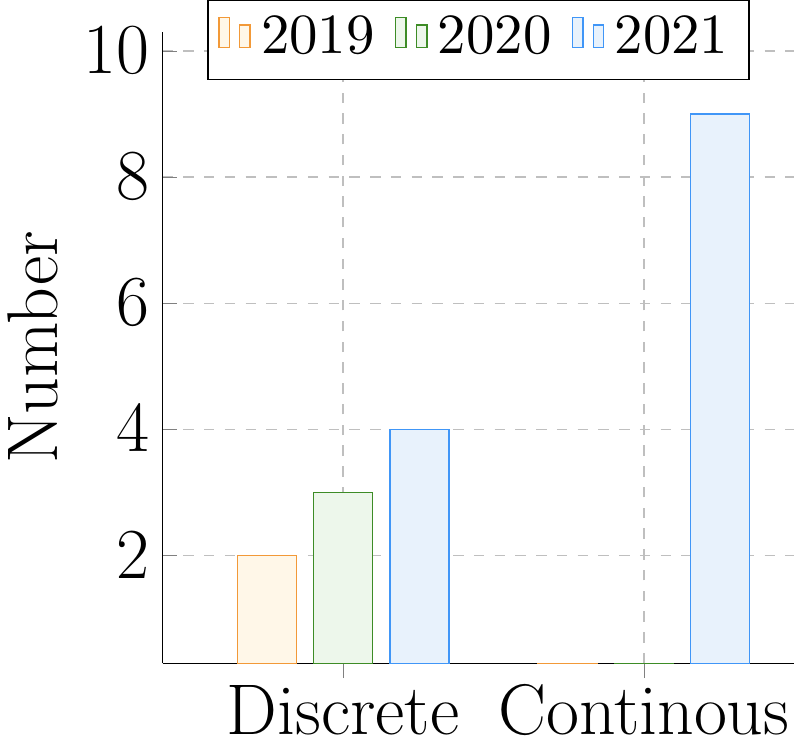}
        } 
    \caption{Meta-analyses over different dimensions. The statistics are based on the works in Tab.~\ref{tab:papers-part1} and Tab.~\ref{tab:papers-part2}. In (d), we use the following abbreviations. TC: text classification, FP: factual probing, GCG: general conditional generation, QA: question answering, CR: commonsense reasoning, SUM: summarization, O: others.}
    \label{fig:meta-analysis}
\end{figure*}

\paragraph{Year} 
With the emergence of different kinds of pre-trained LMs, prompt-based learning has become a more and more active research field, as can be seen in Fig.~\ref{fig:meta-analysis}-(a). We can see a huge surge in 2021, which is perhaps due to the prevalence of GPT-3 \citep{brown2020language}, which greatly increased the popularity of prompting in the few-shot multi-task setting.

\paragraph{Tasks}
We plot the number of works that investigate various tasks in Fig.~\ref{fig:meta-analysis}-(b). For a task that has fewer than 5 relevant works, we group it into ``Others". As the bar chart indicates, most tasks regarding prompt-based learning revolve around text classification and factual probing.
We conjecture that this is because that for these tasks, both template engineering and answer engineering are relatively easy to conduct, and experiments are relatively computationally inexpensive.

\paragraph{Prompt vs. Answer Search}
As noted in previous sections, both prompt and answer search are important tools to take advantage of pre-trained language models for many tasks.
Current research mainly focuses on template search instead of answer search, as shown in Fig.~\ref{fig:meta-analysis}-(c).  

Likely reasons are: (1) For conditional generation tasks (e.g.~summarization or translation), the gold references can be directly used as answer. Although there are many sequences that may share the same semantics, how to effectively conduct multi-reference learning in conditional text generation problems is non-trivial. (2) For classification tasks, most of the time, label words are relative easy to select using domain knowledge.

\paragraph{Discrete Search vs. Continuous Search}
Since there are only a few works focus on automatic answer search, we analyze the automatic template search. As time goes by, there has been a shift from discrete search to continuous search for prompt engineering, as shown in Fig.~\ref{fig:meta-analysis}-(d). Likely reasons are: (1) discrete search is harder to optimize compared to continuous search, (2) soft prompts have greater representation ability.

\section{Conclusion}
In this paper, we have summarized and analyzed several paradigms in the development of statistical natural language processing techniques, and have argued that \term{prompt-based learning} is a promising new paradigm that may represent another major change in the way we look at NLP.
First and foremost, we hope this survey will help researchers more effectively and comprehensively understand the paradigm of prompt-based learning, and grasp its core challenges so that more scientifically meaningful advances can be made in this field.
In addition, looking all the way back to the summary of the four paradigms of NLP research presented in \S\ref{sec:four-paradigms}, we hope to highlight the commonalities and differences between them, making research on any of these paradigms more full-fledged, and potentially providing a catalyst to inspire work towards the next paradigm shift as well.





\section*{Acknowledgements}

We would like to thank Chunting Zhou for her constructive comments on this work.

\bibliography{prompt,misc,sec2,related}
\bibliographystyle{acl_natbib}

\appendix

\clearpage

\section{Appendix on Pre-trained LMs}

In this appendix we present some auxiliary information on pre-trained LMs that may be useful to the readers to better understand the current lay of the land with respect to this dynamic research area.

\subsection{Evolution of Pre-trained LM Parameters}

\begin{figure}[h]
    \centering
    \includegraphics[width=0.6\textwidth]{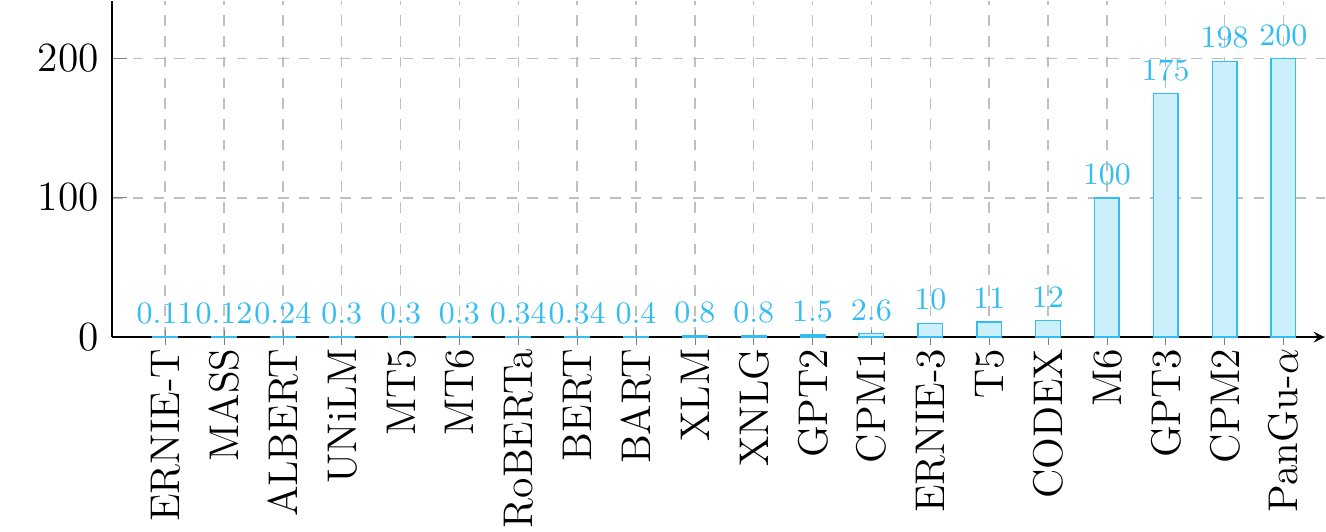}
    \caption{Comparison of the size of existing popular pre-trained language models.}
    \label{fig:param}
\end{figure}

Fig.~\ref{fig:param} lists several popular pre-trained models' statistics of parameters, ranging from 0 to 200 billion. GPT3, CPM2, and PanGu-$\alpha$ are the top three largest models with parameters greater than 150 billion.

\subsection{Auxiliary Objective} \label{appendix:sec:auxiliary}
In this subsection, more auxiliary objectives for pre-training language models have been listed.
\begin{itemize*}
    \item \textbf{Next Sentence Prediction (NSP)} \cite{devlin-etal-2019-bert}: A binary classification loss predicting whether two segments appear consecutively within a larger document, or are random unrelated sentences.
    \item \textbf{Sentence Order Prediction (SOP)} \cite{DBLP:conf/iclr/LanCGGSS20}: A binary classification loss for predicting whether two sentences are in a natural or swapped order.
    \item \textbf{Capital Word Prediction (CWP)} \cite{DBLP:conf/ijcai/0001HH0Z20}: A binary classification objective calculated over each word, predicting whether whether
    each word is capitalized or not. 
    \item \textbf{Sentence Deshuffling (SDS)} \cite{DBLP:conf/ijcai/0001HH0Z20}: A multi-class classification task to reorganize permuted segments.
    \item \textbf{Sentence distance prediction (SDP)} \cite{DBLP:conf/ijcai/0001HH0Z20} : A three-class classification task, predicting the positional relationship between two sentences (adjacent in the same document, not adjacent but in the same document, in different documents).
    \item \textbf{Masked Column Prediction (MCP)} \cite{DBLP:conf/acl/YinNYR20}: Given a table, recover the names and data types of masked columns.
    \item \textbf{Linguistic-Visual Alignment (LVA)} \cite{DBLP:conf/nips/LuBPL19}: A binary classification to Predict whether the text content can be aligned to visual content.
    \item \textbf{Image Region prediction (IRP)} \cite{DBLP:conf/iclr/SuZCLLWD20}: Given an image whose partial features are masked (zeroed out), predict the masked regions.
    \item \textbf{Replaced Token Detection (RTD)} \cite{DBLP:conf/naacl/XiaoLZSTWW21}: A binary classification loss predicting whether each token in corrupted input was replaced by a generative sample or not.
    \item \textbf{Discourse Relation Prediction (DRP)} \cite{DBLP:conf/aaai/SunWLFTWW20}: Predict the semantic or rhetorical relation between two sentences.
    \item \textbf{Translation Language Modeling (TLM)} \cite{lample2019cross}: Consider parallel sentences and mask words randomly in both source and target sentences.  
    \item \textbf{Information Retrieval Relevance (IRR)} \cite{DBLP:conf/aaai/SunWLFTWW20}: Predict the information retrieval relevance of two sentences.
    \item \textbf{Token-Passage Prediction (TPP)} \cite{DBLP:conf/ijcai/0001HH0Z20}: Identify the keywords of a passage appearing in the segment.
    \item \textbf{Universal Knowledge-Text Prediction (UKTP)} \cite{DBLP:journals/corr/abs-2107-02137}: Incorporate knowledge into one pre-trained language model.
    
    \item \textbf{Machine Translation (MT)} \cite{DBLP:journals/corr/abs-2104-08692} : Translate a sentence from the source language into the target language.
    \item \textbf{Translation Pair Span Corruption (TPSC)} \cite{DBLP:journals/corr/abs-2104-08692} : Predict the masked spans from a translation pair.
     \item \textbf{Translation Span Corruption (TSC)} \cite{DBLP:journals/corr/abs-2104-08692} : Unlike TPSC, TSC only masks and predicts the spans in one language.
     \item \textbf{Multilingual Replaced Token Detection (MRTD)} \cite{DBLP:journals/corr/abs-2106-16138}: Distinguish real input tokens from corrupted multilingual sentences by a Generative Adversarial Network, where both the generator and the discriminator are shared across languages.
     \item \textbf{Translation Replaced Token Detection (TRTD)} \cite{DBLP:journals/corr/abs-2106-16138}: Distinguish the real tokens and masked tokens in the translation pair by the Generative Adversarial Network.
     \item \textbf{Knowledge Embedding (KE)} \cite{DBLP:journals/tacl/WangGZZLLT21}: Encode entities and relations in knowledge graphs (KGs) as distributed representations
    \item \textbf{Image-to-text transfer (ITT)} \cite{DBLP:journals/tacl/WangGZZLLT21}: Is similar to the image caption that generates a corresponding description for the input image.
    \item \textbf{Multimodality-to-text transfer (MTT)} \cite{DBLP:journals/tacl/WangGZZLLT21}: Generate the target text based on both the visual information and the noised linguistic information.
\end{itemize*}


\subsection{\includegraphics[scale=0.01]{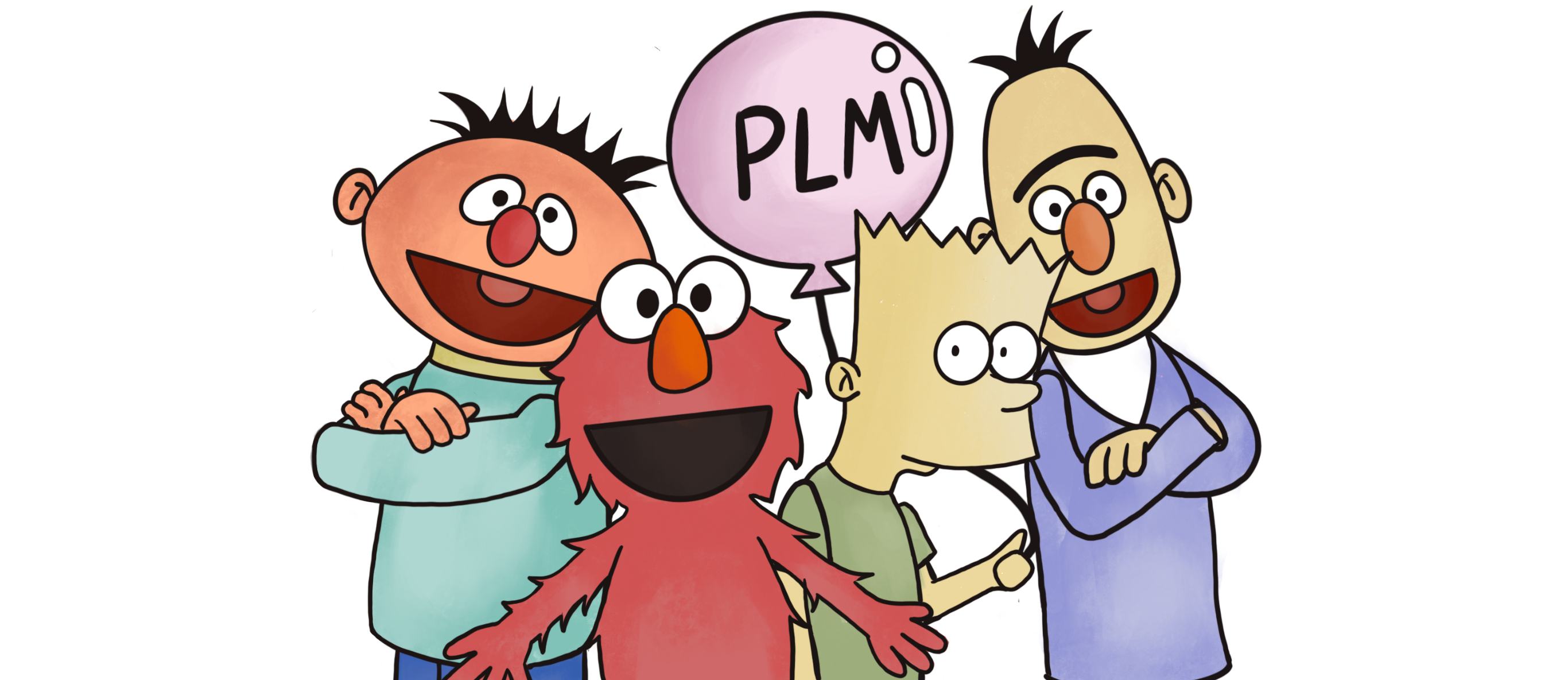} Pre-trained Language Model Families}

The increasing number of models makes it difficult for people to clearly grasp the differences between them. Based on this, we cluster the current mainstream pre-training models and characterize them from diverse dimensions.

\begin{figure}[h]
\centering
\includegraphics[width=15cm]{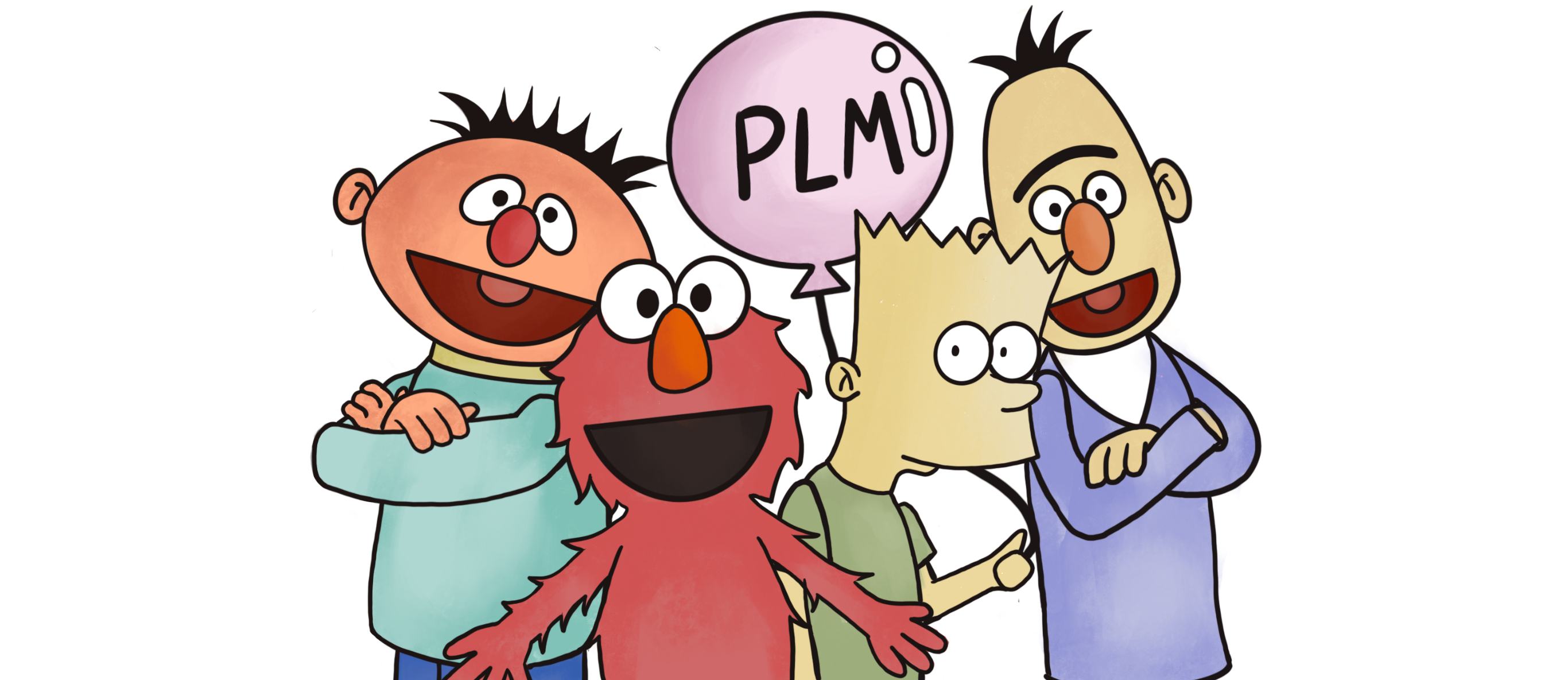}
\end{figure}


\begin{table*}[!htb]
  \renewcommand\tabcolsep{4.2pt}
\renewcommand{\arraystretch}{1.3}
  \centering \scriptsize

    \begin{tabular}{lllllclllll}
    \toprule%
    \multirow{2}[4]{*}{\textbf{Family}} & \multirow{2}[4]{*}{\textbf{Models}} & \multirow{2}[4]{*}{\textbf{LM}} & \multicolumn{3}{c}{\textbf{Pre-training Tasks}} & \multicolumn{4}{c}{\textbf{Corruption}} & \multirow{2}[4]{*}{\textbf{Application}} \\
\cmidrule(lr){4-6} \cmidrule(lr){7-10}          &       &       & \textbf{Main} & \textbf{Auxiliary} & \textbf{Parallel} & \textbf{Mask}  &\textbf{Replace} & \textbf{Delete} & \textbf{Permute}  &  \\
    \midrule
    \multirow{3}[2]{*}{\textbf{GPT}} & GPT \citenumber{Radford2018ImprovingLU}  & L2R    & SLM   & -     & \XSolidBrush  & -     & -     & -     & -     & NLG \\
          & GPT-2 \citenumber{Radford2019LanguageMA}  & L2R    & SLM   & -     & \XSolidBrush   & -    & -     & -     & -     & NLG \\
          & GPT-3 \citenumber{brown2020language}  & L2R    & SLM   & -     & \XSolidBrush  & -  & -     & -     & -     & NLG \\
          & Codex \citenumber{chen2021evaluating}  & L2R    & SLM   & -     & \XSolidBrush  & -  & -     & -     & -     & NLG \\

    \midrule
    \multirow{1}[1]{*}{\textbf{ELMo} \includegraphics[scale=0.01]{fig/appendix/elmo.pdf}}  & ELMo \citenumber{peters-etal-2018-deep}  & L2R    & SLM   & -     & \XSolidBrush  & -   & -     & -     & -     & NLU, NLG \\
    \midrule
    \multirow{12}[2]{*}{\textbf{BERT} \includegraphics[scale=0.01]{fig/appendix/bert.pdf}} & BERT \citenumber{devlin-etal-2019-bert}  & Mask  & CTR   & NSP   & \XSolidBrush  & Tok  & -     & -     & -     & NLU \\
          & RoBERTa \citenumber{Liu2019RoBERTaAR} & Mask  & CTR   & -    & \XSolidBrush   & Tok   & -     & -     & -     & NLU \\
          & SpanBERT \citenumber{joshi-etal-2020-spanbert} & Mask  & CTR   & -    & \XSolidBrush   & Span & -     & -     & -     & NLU \\
          & DeBERTa \citenumber{he2020deberta} & Mask  & CTR   & -    & \XSolidBrush   & Tok & -     & -     & -     & NLU \\          
          & SciBERT \citenumber{beltagy-etal-2019-scibert} & Mask  & CTR   & NSP   & \XSolidBrush  & Tok   & -     & -     & -     & Sci-NLU \\
          & BioBERT \citenumber{10.1093/bioinformatics/btz682} & Mask  & CTR   & NSP  & \XSolidBrush   & Tok  & -     & -     & -     & Bio-NLU \\
          & ALBERT \citenumber{DBLP:conf/iclr/LanCGGSS20} & Mask  & CTR   & SOP  & \XSolidBrush   & Tok   & -     & -     & -     & mSent \\
          & FinBERT \citenumber{DBLP:conf/ijcai/0001HH0Z20} & Mask  & CTR   & \makecell[l]{CWP, SDS, \\ SDP, TPP} & \XSolidBrush  & Span & -     & -     & Sent  & Fin-NLU \\
          & VLBERT \citenumber{DBLP:conf/iclr/SuZCLLWD20} & Mask  & CTR   & IRP   & \Checkmark  & Tok, Region & -     & -     & -     & VLU \\
          & ViLBERT \citenumber{DBLP:conf/nips/LuBPL19} & Mask  & CTR   & IRP, LVA & \Checkmark  & Tok, Region  & -     & -     & -     & VLU \\
          & BEIT \citenumber{bao2021beit} & Mask  & CTR,FTR   & - & \XSolidBrush  & Visual ``Tok''\footnote{Visual tokens refer to image patches.}  & -     & -     & -     & VLU \\          
          & VideoBERT \citenumber{DBLP:conf/iccv/SunMV0S19} & Mask  & CTR   & LVA   & \Checkmark  & Tok, Frame   & -     & -     & -     & VLU \\
          & TaBERT \citenumber{DBLP:conf/acl/YinNYR20} & Mask  & CTR   & MCP & \Checkmark  & Tok, Column   & -    & -     & -     & Tab2Text \\
          & mBERT \citenumber{devlin-etal-2019-bert} & Mask  & CTR   & NSP   & \XSolidBrush  & Tok   & -     & -     & -     & XLU \\
          & TinyBERT \citenumber{DBLP:conf/emnlp/JiaoYSJCL0L20} & Mask  & CTR   & NSP  & \XSolidBrush   & Tok   & -     & -     & -     & XLU \\
    \midrule
    \multirow{5}[2]{*}{\textbf{ERNIE} \includegraphics[scale=0.01]{fig/appendix/ernie.pdf}} & ERNIE-T \citenumber{DBLP:conf/acl/ZhangHLJSL19} & Mask  & CTR   & NSP  & \XSolidBrush   & Tok, Entity & -     & -     & -     & NLU \\
          & ERNIE-B \citenumber{sun2019ernie} & Mask  & CTR   & -   & \XSolidBrush  & Tok,Entity, Phrase   & -     & -     & -     & NLU \\
          & ERNIE-NG \citenumber{DBLP:conf/naacl/XiaoLZSTWW21} & Mask  & CTR   & RTD  & \XSolidBrush   & N-gram    & Tok     & -     & -     & NLU \\
          & ERNIE-B2 \citenumber{DBLP:conf/aaai/SunWLFTWW20} & Mask  & CTR   & \makecell[l]{CWP,SDS,SOP,\\SDP,DRP,IRR}  & \XSolidBrush  & Entity, Phrase  & -     & -     & Sent  & NLU \\
          & ERNIE-M \citenumber{DBLP:journals/corr/abs-2012-15674} & LPM   & CTR   & -    & \Checkmark   & Tok  & -     & -     &       & XLU, XLG \\
          &ERNIE-B3 \citenumber{DBLP:journals/corr/abs-2107-02137} & Mask     & CTR     & SOP,SDP,UKTP  & \XSolidBrush     & Entity, Phrase    & -     & -     & -     &  NLU    \\ 
    \midrule 
    \multirow{2}[2]{*}{\textbf{BART} \includegraphics[scale=0.012]{fig/appendix/bart.pdf}} & BART \citenumber{lewis-etal-2020-bart}  & En-De & FTR   & -    & \XSolidBrush   & Tok    & Span  & Tok   & Sent,Doc     & NLU, NLG \\
          & mBART \citenumber{DBLP:journals/tacl/LiuGGLEGLZ20} & En-De & FTR   & -    & \XSolidBrush   & Span   & -     & -     & Sent  & NLG \\
    \midrule
    \multirow{2}[2]{*}{\textbf{UniLM}} & UniLM1 \citenumber{DBLP:conf/nips/00040WWLWGZH19} & LPM   & SLM,CTR & NSP   & \XSolidBrush  & Tok    & -     & -     & -     & NLU, NLG \\
          & UniLM2 \citenumber{DBLP:conf/icml/Bao0WW0L0GP0H20} & LPM   & SLM,CTR & -  & \XSolidBrush     & Tok    & -     & -     & Tok   & NLU, NLG \\
    \midrule
    \multirow{3}[2]{*}{\textbf{T5}} & T5 \citenumber{JMLR:v21:20-074}    & En-De & CTR   & -   & \XSolidBrush    & -   & Span  & -     & -     & NLU, NLG \\
          & mT5 \citenumber{DBLP:conf/naacl/XueCRKASBR21}   & En-De & CTR   & -  & \XSolidBrush     & -    & Span  & -     & -     & XLU, XLG \\
          & mT6 \citenumber{DBLP:journals/corr/abs-2104-08692}   & En-De & CTR    & MT,TPSC,TSC  & \Checkmark     & -     & Span & -     & -     & XLU, XLG \\
          & ByT5 \citenumber{DBLP:journals/corr/abs-2105-13626}   & En-De & CTR   & -   & \XSolidBrush    & -     & byte-span & -     & -     & XLU, XLG \\
    \midrule
    \multirow{2}[2]{*}{\textbf{XLM}} & XLM \citenumber{lample2019cross}  & LPM   & CTR   & TLM  & \Checkmark     & Tok   & -     & -     & -     & XLU, XLG \\
          & XLM-R \citenumber{DBLP:conf/acl/ConneauKGCWGGOZ20} & Mask  & CTR   & -    & \XSolidBrush    & Tok    & -     & -     & -     & XLU \\
          & XLM-E \citenumber{DBLP:journals/corr/abs-2106-16138} & Mask  & CTR   &  MRTD,TRTD  & \XSolidBrush     & -    & Tok     & -     & -     & XLU, XLG \\
    \midrule
    \multirow{2}[2]{*}{\textbf{CPM}} & CPM \citenumber{DBLP:journals/corr/abs-2012-00413}  & L2R    & SLM   & -   & \XSolidBrush    & -   & -     & -     & -     & NLG \\
          & CPM-2  \citenumber{DBLP:journals/corr/abs-2106-10715}  & En-De    & CTR   & -  & \XSolidBrush     & Span   & -     & -     & -     & NLU,NLG \\    
     \midrule   
    \multirow{7}[2]{*}{\textbf{Other}} & XLNet \citenumber{DBLP:conf/nips/YangDYCSL19} & L2R    & SLM   & -   & \XSolidBrush    & -    & -      &  -     & Tok   & NLU \\
          & PanGu-$\alpha$ \citenumber{zeng2021pangualpha} & L2R    & SLM   & -   & \XSolidBrush    & -     & -     & -     & -     & NLG \\
          & ELECTRA \citenumber{DBLP:conf/iclr/ClarkLLM20} & Mask & CTR   & RTD & \XSolidBrush        & Tok   & Tok     & -     & -     & NLU,NLG \\
          & MASS \citenumber{DBLP:conf/icml/SongTQLL19}  & En-De & CTR   & -   & \XSolidBrush    & Span     & -     & -     & -     & NLG \\
          & PEGASUS \citenumber{DBLP:conf/icml/ZhangZSL20} & En-De & CTR   & - & \XSolidBrush         & Tok, Sent   & -     & -     & -     & Summarization \\
          & M6 \citenumber{DBLP:journals/tacl/WangGZZLLT21} & En-De & CTR   & ITT,MTT  & \XSolidBrush     & Span   & -     & -     & -     & NLG \\
          
    \bottomrule
    \end{tabular}%
  \caption{
  A detailed illustration of different pre-trained models characterized by the four aspects.  
  ``\textbf{Parallel}'' represents if parallel data have been used for pre-training.
  \texttt{Sci}, \texttt{Bio}, \texttt{Fin}, \texttt{K} represent scientific, biomedical, financial, and knowledge, respectively.
  \texttt{Tok}, \texttt{Sent}, \texttt{Doc} denote token, sentence and document, respectively.
  \texttt{Region}, \texttt{Frame} denote basic units of images and video respectively.
  }
  \label{tab:pretrained-aspect}%
\end{table*}%

\clearpage

\end{document}